\documentclass{article}

\PassOptionsToPackage{numbers, compress}{natbib}
\usepackage[final]{neurips_2020}

\usepackage[utf8]{inputenc} %
\usepackage[T1]{fontenc}    %
\usepackage[hypertexnames=false]{hyperref}      %
\usepackage{xcolor}
\hypersetup{
    colorlinks,
    linkcolor={red!50!black},
    citecolor={blue!50!black},
    urlcolor={blue!80!black}
}

\renewcommand{\paragraph}[1]{\vspace{-0.1em}\noindent \textbf{#1}}

\usepackage{inconsolata}
\usepackage{url}            %
\usepackage{booktabs}       %
\usepackage{amsfonts}       %
\usepackage{nicefrac}       %
\usepackage{microtype}      %
\usepackage{marginalia}
\usepackage{xr-hyper}
\usepackage{amsthm}
\usepackage{array,amssymb,latexsym,amssymb,epsfig,epsf,amsmath,cite,accents,float,graphicx,natbib,mathrsfs,verbatim,bm,color,enumitem}
\graphicspath{{figures/}}
\usepackage[linesnumbered,noend]{algorithm2e}
\SetKwInput{KwEn}{Encoding}
\SetKwInput{KwDe}{Decoding}
\usepackage{wrapfig}
\usepackage{transparent}

\usepackage[capitalise]{cleveref}
\usepackage[font=small,labelfont=bf]{caption}
\usepackage{subcaption}

\newcommand{\threecolfigwidth}{0.32\textwidth}
\newcommand{\twocolfigwidth}{0.49\textwidth}
\newcommand{\capshift}{-6mm}

\crefname{ineq}{}{}
\creflabelformat{ineq}{#2{\upshape(#1)}#3} 
\crefname{prob}{}{}
\creflabelformat{prob}{#2{\upshape(#1)}#3} 

\newtheorem{lemma}{Lemma}
\newtheorem{theorem}{Theorem}
\newtheorem{corollary}{Corollary}

\newtheorem{definition}{Definition}

\newtheorem{proposition}{Proposition}

\newcommand{\ud}{\,\mathrm{d}}
\newcommand{\ENCODE}{\mathrm{ENCODE}}
\newcommand{\DECODE}{\mathrm{DECODE}}

\newcommand{\erf}{\mathrm{erf}}

\newcolumntype{C}[1]{>{\centering\arraybackslash}m{#1}}

\def\reals{\mathbb{R}}

\def\integers{\mathbb{Z}}

\def\hbf{{\bf h}}

\def\sbf{{\bf s}}

\def\vbf{{\bf v}}
\def\wbf{{\bf w}}

\def\ybf{{\bf y}}

\def\ybf{{\bf y}}

\def\Ibf{{\bf I}}

\def\Pbf{{\bf P}}

\def\Ic{{\cal I}}

\def\Lc{{\cal L}}

\def\Nc{{\cal N}}

\def\Sc{{\cal S}}
\def\Tc{{\cal T}}
\def\Uc{{\cal U}}

\def\nn{\nonumber}
\def\beq{\begin{equation}}
\def\eeq{\end{equation}}
\def\beqa{\begin{eqnarray}}
\def\eeqa{\end{eqnarray}}
\def\balign{\begin{align}}
\def\ealign{\end{align}}
\def\bpr{\begin{proof}}
\def\epr{\end{proof}}
\def\bth{\begin{theorem}}
\def\eth{\end{theorem}}
\def\blm{\begin{lemma}}
\def\elm{\end{lemma}}
\def\bprop{\begin{proposition}}
\def\eprop{\end{proposition}}
\def\bcr{\begin{corollary}}
\def\ecr{\end{corollary}}

\def\ie{{\it i.e.,\ \/}}
\def\eg{{\it e.g.,\ \/}}
\def\defeq{\triangleq}

\def\E{\mathbb{E}}
\def\gb{\succeq}

\def\and {{\rm and}}
\def\where {{\rm where}}

\newcommand{\fun}{f}
\newcommand{\pdf}{p}
\newcommand{\cdf}{F}
\newcommand{\ql}{\ell}
\newcommand{\bql}{{\boldsymbol \ell}}
\newcommand{\level}[1]{{\tau(#1)}}
\newcommand{\levelt}[1]{{\tilde \ql(#1)}}
\newcommand{\qcoeff}{{\rho}}
\newcommand{\feasl}{{\Lc}}
\newcommand{\rsym}{{\theta}}  %
\newcommand{\onesol}{{\beta}}
\DeclareMathOperator{\sign}{sign}
\newcommand{\signvec}{{\sbf}}
\newcommand{\dxdy}[2]{{\frac{\partial {#1}}{\partial {#2}}}} %
\newcommand{\dpsidl}{{\dxdy{\Psi(\bql(t))}{\ql_{j}}}}

\newcommand{\sgd}{{SGD}\xspace}
\newcommand{\supersgd}{{SuperSGD}\xspace}
\newcommand{\qinf}{{QSGDinf}\xspace}
\newcommand{\terngrad}{{TRN}\xspace}
\newcommand{\nuq}{{NUQSGD}\xspace}
\newcommand{\alqgnbased}{{ALQG}\xspace}
\newcommand{\alqgnless}{{ALQG-N}\xspace}
\newcommand{\alqnbased}{{ALQ}\xspace}
\newcommand{\amqnbased}{{AMQ}\xspace}
\newcommand{\alqnless}{{ALQ-N}\xspace}
\newcommand{\amqnless}{{AMQ-N}\xspace}

\title{Adaptive Gradient Quantization\\
for Data-Parallel SGD}

\newcommand*\samethanks[1][\value{footnote}]{\footnotemark[#1]}
\author{
    Fartash Faghri$^{1, 2}$\thanks{Equal contributions.}
    \And
    Iman Tabrizian$^{1, 2}$\samethanks{}
    \And
    Ilia Markov$^3$
    \And
    Dan Alistarh$^{3, 4}$
    \AND
    \vspace*{-.8cm}
    Daniel M. Roy$^{1, 2}$
    \And
    Ali Ramezani-Kebrya$^2$
    \And
    \vspace*{-.7cm}
    \\
    $^1$University of Toronto\quad
    $^2$Vector Institute\quad
    $^3$IST Austria\quad
    $^4$NeuralMagic\\[.3cm]
    \texttt{faghri@cs.toronto.edu}\quad
    \texttt{iman.tabrizian@mail.utoronto.ca}\quad
    \texttt{alir@vectorinstitute.ai}
}

\begin{document}

\maketitle

\begin{abstract}
    Many communication-efficient variants of SGD use gradient quantization 
    schemes.  These schemes are often heuristic and fixed over the course of 
    training. We empirically observe that the statistics of gradients of 
    deep models change during the training. Motivated by this observation, we 
    introduce two adaptive quantization schemes, ALQ and AMQ\@.  In both 
    schemes, processors update their compression schemes in parallel by 
    efficiently computing sufficient statistics of a parametric distribution.  
    We improve the validation accuracy by almost $2\%$ on CIFAR-10 and $1\%$ on 
    ImageNet in challenging low-cost communication setups.  Our adaptive 
    methods are also significantly more robust to the choice of hyperparameters.
\end{abstract}

\section{Introduction}\label{sec:intro}
    \begin{wrapfigure}{R}{0.4\textwidth}
        \vspace*{-0.7cm}
            \includegraphics[width=0.39\textwidth]{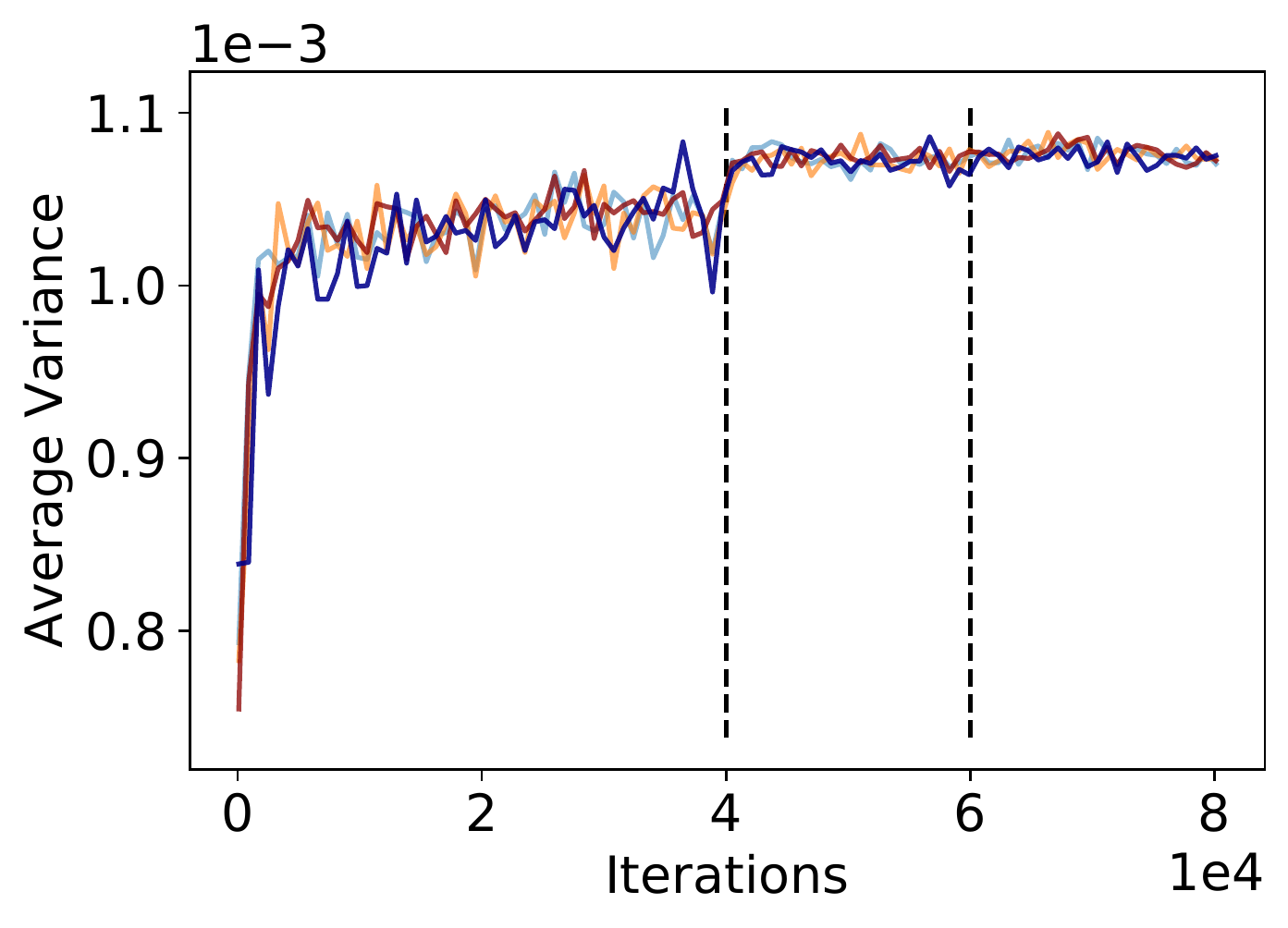}
            \caption{Changes in the average variance of normalized gradient 
            coordinates in a ResNet-$32$ model trained on CIFAR-10. Colors 
            distinguish different runs with different seeds.  Learning rate is 
            decayed by a factor of $10$ twice at 40K and 60K iterations. The 
            variance changes rapidly during the first epoch. The next 
            noticeable change happens after the first learning rate drop and 
            another one appears after the second drop.}
            \label{fig:variance-change}
        \vspace*{-0.4cm}
    \end{wrapfigure}
Stochastic gradient descent (SGD) and its 
variants are currently the method of choice for training deep models. 
Yet, large datasets cannot always be trained on
a single computational node due to memory and scalability limitations. 
Data-parallel SGD is a remarkably scalable 
variant, in particular on multi-GPU systems~\citep{Zinkevich, Scaleup, Recht11, 
Dean12, Coates, Projectadam, Li14, Duchi15, Petuum, Zhang15}. 
However, despite its many advantages, distribution  
introduces new challenges for optimization algorithms. In particular, 
data-parallel SGD has large communication cost due to the need to transmit potentially huge gradient vectors.  
Ideally, we want distributed optimization methods that match the performance of 
SGD on a single hypothetical super machine, while paying a negligible 
communication cost.

A common approach to reducing the communication cost in data-parallel SGD is 
gradient compression and quantization~\citep{Dean12, Seide14, Gupta, Abadi, 
Dorefa, TernGrad, signSGD}. In full-precision data-parallel SGD, each processor 
broadcasts its locally computed stochastic gradient vector at every iteration, 
whereas in quantized data-parallel SGD, each processor compresses its 
stochastic gradient before broadcasting. Current quantization methods are 
either designed heuristically or fixed prior to training. Convergence rates in a stochastic optimization problem are controlled by the trace of the gradient covariance matrix, which is referred as the gradient variance in this paper \citep{Bubeck}.  As 
\cref{fig:variance-change} shows, no fixed method can be optimal  
throughout the entire training because the distribution of gradients changes.  
A quantization method that is optimal at the first iteration will not be optimal 
after only a single epoch.

In this paper, we propose two adaptive methods for quantizing the gradients in 
data-parallel SGD\@. We study methods that are defined by a norm and a set of quantization levels. In Adaptive 
Level Quantization (ALQ), we minimize the excess variance of quantization given 
an estimate of the distribution of the gradients. In Adaptive Multiplier 
Quantization (AMQ), we minimize the same objective as ALQ by modelling 
quantization levels as exponentially spaced levels. AMQ solves for the optimal 
value of a single multiplier parametrizing the exponentially spaced levels.

\subsection{Summary of contributions}
\begin{itemize}[leftmargin=*,itemsep=0ex]
    \item We propose two adaptive gradient quantization methods, ALQ and AMQ, 
        in which processors update their compression methods in parallel.
      \item We establish an upper bound on the excess variance for any arbitrary 
        sequence of quantization levels under general normalization that is tight in dimension, an upper 
        bound on the expected number of communication bits per iteration, and 
        strong convergence guarantees on a number of problems under standard 
        assumptions. Our bounds hold for any adaptive method, including ALQ and 
        AMQ.
    \item
        We improve the validation accuracy by almost $2\%$ on CIFAR-10 and $1\%$ on 
        ImageNet in challenging low-cost communication setups.  Our adaptive 
        methods are significantly more robust to the choice of hyperparameters.\footnote{Open source code: 
        \url{http://github.com/tabrizian/learning-to-quantize}}
\end{itemize}

\subsection{Related work}  
Adaptive quantization has been used for speech communication and storage \citep{Cummiskey}. In machine learning, several biased and unbiased  schemes have been proposed to compress networks and gradients. Recently, lattice-based quantization has been studied for distributed mean estimation and variance reduction \citep{lattice}. In this work, we focus on unbiased and coordinate-wise schemes to compress gradients. 

\citet{QSGD} proposed Quantized SGD (QSGD) focusing on the uniform quantization of stochastic 
gradients normalized to have unit Euclidean norm. Their experiments illustrate a similar quantization method, where gradients are normalized to have unit $L^\infty$ norm, achieves better performance. We refer to this method as QSGDinf or Qinf in short. \citet{TernGrad} proposed TernGrad, which can be viewed as a special case of QSGDinf with three quantization levels.

\citet{NUQSGD} proposed nonuniform quantization levels (NUQSGD) and demonstrated 
superior empirical results compared to QSGDinf. \citet{Samuel} proposed natural compression and dithering schemes, where the latter is a special case of logarithmic quantization.%

There have been prior attempts at adaptive quantization methods.  
\citet{ZipML} proposed ZipML, which is an optimal quantization method if all points to be quantized
are known a priori. To find the optimal sequence of 
quantization levels, a dynamic program is solved whose computational and 
memory cost is quadratic in the number of points to be quantized, which in the case of gradients would correspond to their dimension. 
For this reason, ZipML is 
impractical for quantizing on the fly, and is in fact used for (offline) dataset compression. 
They also proposed an 
approximation where a subsampled set of points is used and proposed to scan 
the data once to find the subset. However, as we show in this paper, this 
one-time scan is not enough as the distribution of stochastic gradients changes 
during the training.

\citet{LQ-Net} proposed LQ-Net, where weights and activations are 
quantized such that the inner products can be computed efficiently with bitwise 
operations. Compared to LQ-Net, our methods do not need additional memory for 
encoding vectors. Concurrent with our work, \citet{TINYSCRIPT} proposed to 
quantize activations and gradients by modelling them with Weibull 
distributions. In comparison, our proposed methods accommodate general 
distributions. Further, our approach does not require any assumptions on the 
upper bound of the gradients. 

\section{Preliminaries: data-parallel SGD}\label{sec:conv}

Consider the problem of training a model parametrized by a high-dimensional vector  
$\wbf\in\reals^d$. Let $\Omega\subseteq\reals^d$ denote a closed and compact set. 
Our goal is to minimize  $\fun:\Omega\rightarrow \reals$. Assume we have access to 
unbiased stochastic gradients of $\fun$, which is $g$, such that
$\E[g(\wbf)]=\nabla \fun(\wbf)$ for all $\wbf\in\Omega$.

The update rule for full-precision SGD is given by
$\wbf_{t+1}=\Pbf_{\Omega}\big(\wbf_t-\alpha g(\wbf_t))$
where $\wbf_t$ is the current parameter vector, $\alpha$ is the learning rate, and $\Pbf_{\Omega}$ is the Euclidean projection onto $\Omega$.  
We consider data-parallel SGD, which is a synchronous and distributed framework 
consisting of $M$ processors. Each processor receives gradients from all other 
processors and aggregates them.
In data-parallel SGD with compression, gradients are compressed by each 
processor before transmission and decompressed before 
aggregation~\citep{QSGD,NUQSGD,Samuel,ZipML}. A stochastic compression method 
is unbiased if the vector after decompression is in expectation the same as the 
original vector. %

\section{Adaptive quantization}\label{sec:aqsgd}
\begin{algorithm}[t]
\small
\SetAlgoLined
    \KwIn{Local data, parameter vector (local copy) $\wbf_t$, learning rate 
    $\alpha$, and set of update steps $\Uc$}
\For{$t=1$ {\bfseries to} $T$}{
    \If{$t\in\Uc$}{
        \For{$i=1$ {\bfseries to} $M$}{
            Compute sufficient statistics and update quantization levels 
            $\bql$\;
        }
    }
    \For{$i=1$ {\bfseries to} $M$}{
        Compute $g_i(\wbf_t)$, encode $c_{i,t}\leftarrow 
        \ENCODE_\bql\big(g_i(\wbf_t)\big)$, and broadcast $c_{i,t}$\;
    }
    \For{$j=1$ {\bfseries to} $M$ }{
        Receive $c_{i,t}$ from each processor $i$ and decode $\hat 
        g_i(\wbf_t)\leftarrow \DECODE_\bql\big(c_{i,t}\big)$\;
        Aggregate
        $\wbf_{t+1}\leftarrow \Pbf_{\Omega}\big(\wbf_t-\frac{\alpha}{M}\sum_{i=1}^M\hat g_i 
        (\wbf_t)\big)$\;
        \label{step:agg}
    }
}
\vspace{0.1cm}
\caption{Adaptive data-parallel SGD. Loops are executed in parallel on each 
    machine. At certain steps, each processor computes sufficient statistics of a parametric 
distribution to estimate distribution of normalized coordinates.
    }
\label{AQSGDalg}
\end{algorithm}

In this section, we introduce novel adaptive compression methods that adapt during the training 
(\cref{AQSGDalg}).
Let ${\vbf\in\reals^d}$ be a vector we seek to quantize and 
${r_i=|v_i|/\|\vbf\|}$ be its normalized coordinates for ${i=1,\ldots,d}$.\footnote{In this section, we use $\|\cdot\|$ to denote a general $L^q$ norm with $q\geq 1$ for simplicity.} Let
${q_\bql(r):[0,1]\rightarrow[0,1]}$
denote a random quantization function applied to the normalized coordinate 
${r}$ using adaptable quantization levels,
$\bql= [\ql_0,\ldots,\ql_{s+1}]^\top$,
where $0=\ql_0<\ql_1<\cdots<\ql_s<\ql_{s+1}=1$.
For $r\in[0,1]$, let ${\level{r}}$ denote the index of a level such that 
${\ql_{\level{r}}\leq r<\ql_{\level{r}+1}}$.
Let $\qcoeff(r)=(r-\ql_{\level{r}})/(\ql_{\level{r}+1}-\ql_{\level{r}})$ be the 
relative distance of $r$ to level $\level{r}+1$. We define the random variable
$h(r)$ such that $h(r)=\ql_{\level{r}}$ with probability $1-\qcoeff(r)$ and 
$h(r)=\ql_{\level{r}+1}$ with probability $\qcoeff(r)$. %

\begin{wrapfigure}[10]{r}{0.4\textwidth}
    \vspace*{-0.2in}
    \includegraphics[width=0.4\textwidth]{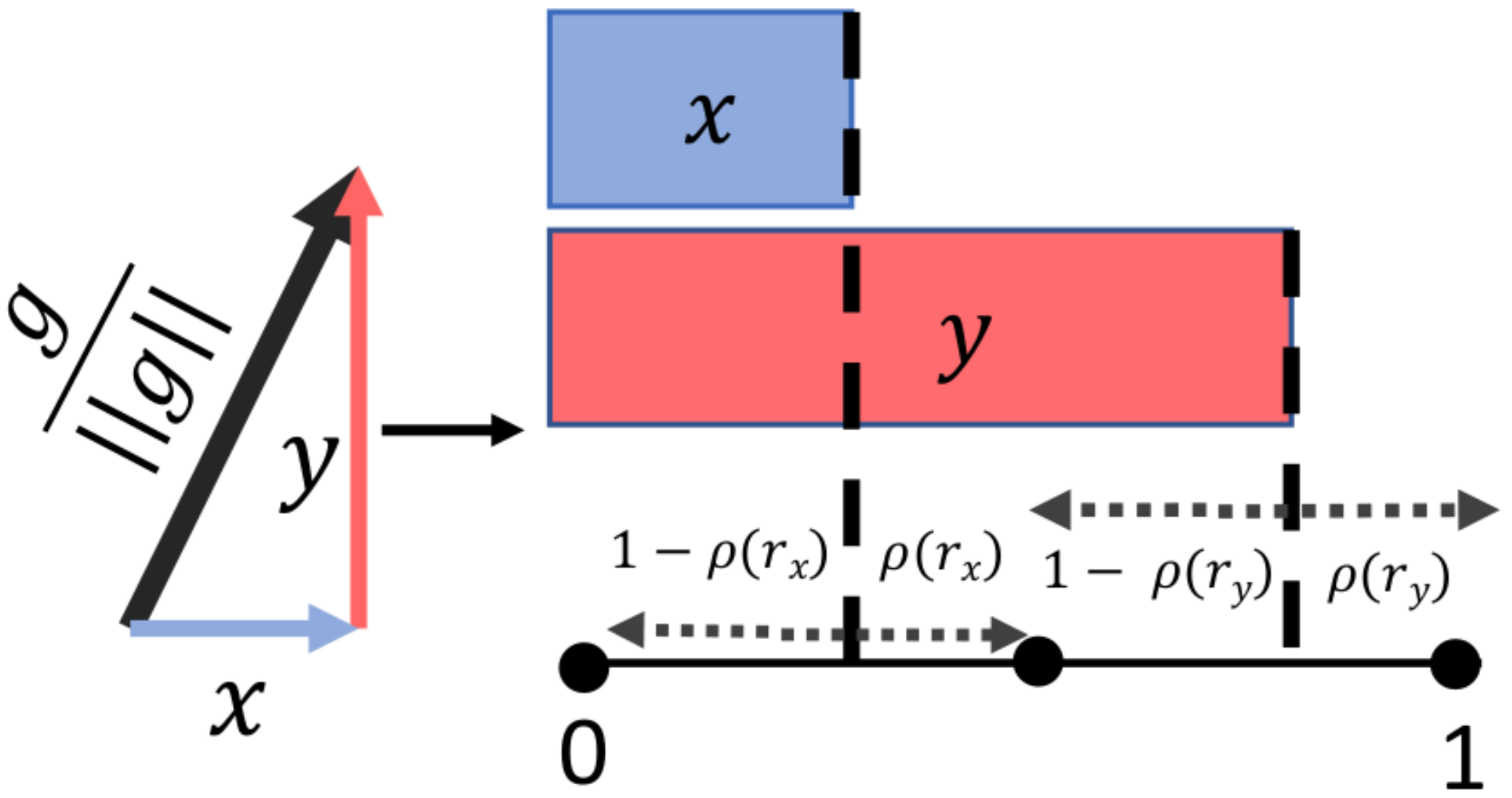}
    \caption{Random quantization of normalized gradient.}
\end{wrapfigure}

We define the quantization of $\vbf$  as
$Q_\bql(\vbf)\defeq [q_\bql(v_1),\ldots,q_\bql(v_d)]^\top$
where $q_\bql(v_i)={\|\vbf\|\cdot \sign(v_i)\cdot h(r_i)}$ and 
$\hbf=\{h(r_i)\}_{i=1,\ldots,d}$ are independent random variables.
The encoding, $\ENCODE(\vbf)$, of a stochastic gradient is the combined 
encoding of $\|\vbf\|$ using a standard floating point encoding along with an optimal encoding of $h(r_i)$ and binary encoding of $\sign(v_i)$ for each coordinate $i$.
The decoding, $\DECODE$, recovers the norm, $h(r_i)$, and the sign.
Additional details of the encoding method are described in \cref{app:coding}.

We define the variance of vector quantization to be the trace of the covariance matrix, 
\begin{align}\label{var}
    \E_\hbf[\|Q_\bql(\vbf)-\vbf\|_2^2]=\|\vbf\|^2 \sum_{i=1}^d\sigma^2(r_i),
\end{align}
where  $\sigma^2(r)=\E[(q_\bql(r)-r)^2]$ is the variance of quantization for a single 
coordinate that is given by%
\begin{align}
    \sigma^2(r) &=(\ql_{\level{r}+1}-r)(r-\ql_{\level{r}}).
\end{align}

Let $\vbf$ be a random vector corresponding to a stochastic gradient and $\hbf$ 
capture the randomness of quantization for this random vector as defined above.  
We define two minimization problems, expected variance and expected normalized 
variance minimization:
\begin{align}\nn%
    \min_{\bql\in\feasl}\, \E_{\vbf,\hbf}\left[\|Q_\bql(\vbf)-\vbf\|_2^2\right]
        \qquad\text{and}\qquad
    \min_{\bql\in\feasl}\, 
        \E_{\vbf,\hbf}\left[\|Q_\bql(\vbf)-\vbf\|_2^2/\|\vbf\|^2\right],
\end{align}
where $\feasl=\{\bql:\ql_j\leq \ql_{j+1},~\forall~j,~\ql_0=0,~\ql_{s+1}=1 \}$ denotes the set of 
feasible solutions. We first focus on the problem of minimizing the expected 
normalized variance and then extend our methods to minimize the expected 
variance in \cref{body:norm}.  Let $\cdf(r)$ denote the marginal cumulative 
distribution function (CDF) of a normalized coordinate $r$.
Assuming normalized coordinates $r_i$ are i.i.d.\  given $\|\vbf\|$, the 
expected normalized variance minimization can be written as 
\begin{align}\label[prob]{exp_var_min_cond}
    \min_{\bql\in\feasl}\,
        \Psi(\bql)
        &,~\where~\Psi(\bql)\defeq\sum_{j=0}^s\int_{\ql_j}^{\ql_{j+1}}\sigma^2(r)\ud \cdf(r).
\end{align}

The following theorem suggests that solving \cref{exp_var_min_cond} is 
challenging in general; however, the sub-problem of optimizing a single level 
given other levels can be solved efficiently in closed form. Proofs are 
provided in \cref{app:varmin}.

\bth[Expected normalized variance minimization]\label{thm:exp_var_min}
Problem \cref{exp_var_min_cond} is nonconvex in general. However, the optimal 
solution to minimize \emph{one} level given other levels, $\min_{\ql_i} 
\Psi(\bql)$,  is given by $\ql_i^*=\onesol(\ql_{i-1}, \ql_{i+1})$, where
\begin{align}\label{eq:onesol}
    \onesol(a, c)
    =\cdf^{-1}\left(\cdf(c) - \int_a^c\frac{r-a}{c-a}\ud \cdf(r)\right).
\end{align}
\eth

\subsection{ALQ: Adapting individual levels using coordinate descent}\label{sec:alq}

Using the single level update rule in \cref{eq:onesol} we iteratively adapt 
individual levels to minimize the expected normalized variance in 
\cref{exp_var_min_cond}.
We denote quantization levels at iteration $t$ by $\bql(t)$ starting from 
$t=0$.  The update rule is
\begin{align}\label{update_coor_gen}
\begin{split}
    \ql_{j}(t+1) &= \onesol(\ql_{j-1}(t), \ql_{j+1}(t))
    \qquad\forall j=1,\ldots,s\,.
\end{split}
\end{align}

Performing the update rule above sequentially over coordinates $j$ is a form of 
coordinate descent (CD) that is guaranteed to converge to a local minima. CD is 
particularly interesting because it does not involve any projection step to the 
feasible set $\feasl$. In practice, we initialize the levels with either 
uniform levels~\citep{QSGD} or exponentially spaced levels proposed in 
\citep{NUQSGD}. We observe that starting from either initialization CD 
converges in small number of steps (less than $10$).

\subsection {Gradient descent}\label{body:pgd}

Computing $\nabla \Psi$ using Leibniz's rule \citep{Calculus}, the gradient 
descent (GD) algorithm to solve \cref{exp_var_min_cond} is based on the 
following update rule: %
\begin{align}\label{update_grad_gen}
\begin{split}
    \ql_j(t+1) &= \Pbf_{\feasl}\left(\ql_j(t) - \eta(t) \dpsidl\right)\\
    \dpsidl
    &=\int_{\ql_{j-1}(t)}^{\ql_j(t)}(r-\ql_{j-1}(t))\ud \cdf(r)
    -\int_{\ql_j(t)}^{\ql_{j+1}(t)}(\ql_{j+1}(t)-r)\ud \cdf(r)
\end{split}
\end{align}
for $t=0,1,\ldots$ and $j=1,\ldots,s$. 
Note that the projection step in \cref{update_grad_gen} is itself a convex 
optimization problem. We propose a projection-free modification of GD update rule to 
systematically ensure $\bql\in\feasl$. Let $\delta_j(t)=\min 
\{\ql_{j}(t)-\ql_{j-1}(t),\ql_{j+1}(t)-\ql_j(t)\}$ denote the minimum distance 
between two neighbouring levels at iteration $t$ for $j=1,\ldots,s$. If the change in level $j$ is bounded by $\delta_j(t)/2$, it is guaranteed that 
$\bql\in\feasl$. We propose to replace \cref{update_grad_gen} with the 
following update rule:
\begin{align}\label{auto_tune_gd_gen}
    \ql_j(t+1) = \ql_j(t) - \sign\left(\dpsidl\right)
    \min\left\{\eta(t)\left|\dpsidl\right|, \frac{\delta_j(t)}{2}\right\}.
\end{align}

\subsection{AMQ: Exponentially spaced levels}\label{sec:amq}

We now focus on $\bql=[-1,-p,\ldots,-p^s,p^s,\ldots,p,1]^\top$, \ie exponentially 
spaced levels with symmetry. We can update $p$ efficiently by gradient descent 
using the first order derivative
\begin{align}
    \frac{1}{2}\frac{\ud \Psi(p)}{\ud p}
    &= \int_{0}^{p^s} 2sp^{2s-1} \ud \cdf(r)
    +\sum_{j=0}^{s-1}\int_{p^{j+1}}^{p^j}
    \left((jp^{j-1} + (j+1)p^j)r - (2j+1)p^{2j}\right)\ud \cdf(r).
\end{align}

\subsection{Expected variance minimization}\label{body:norm}
In this section, we consider the problem of minimizing the expected variance of quantization:
\begin{align}\label[prob]{exp_var_min_uncond}
    \min_{\bql\in\feasl}\, \E_{\vbf,\hbf}\left[\|Q_\bql(\vbf)-\vbf\|_2^2\right].
\end{align}

To solve the expected variance minimization problem, suppose that we observe 
$N$ stochastic gradients $\{\vbf_1,\ldots,\vbf_N\}$. Let $\cdf_n(r)$ and 
$\pdf_n(r)$ denote the CDF and PDF of normalized coordinate conditioned on 
observing $\|\vbf_n\|$, respectively. By taking into account randomness in 
$\|\vbf\|$ and using the law of total expectation, an approximation of the 
expected variance in \cref{exp_var_min_uncond} is given by 
\begin{align}\label{expvarapprox}
    \E[\|Q_s(\vbf)-\vbf\|_2^2]
    \approx  \frac{1}{N}\sum_{n=1}^N
        \|\vbf_n\|^2
            \sum_{j=0}^s
                    \int_{\ql_j}^{\ql_{j+1}}
                        \sigma^2(r)\ud \cdf_n(r).
\end{align}

The optimal levels to minimize \cref{expvarapprox} 
are a solution to the following problem: 
\begin{align}%
    \bql^*
    =\arg\min_{\bql\in\feasl}
    \sum_{n=1}^N
        \|\vbf_n\|^2
            \sum_{j=0}^s
                \int_{\ql_j}^{\ql_{j+1}}
                    \sigma^2(r)\ud \cdf_n(r)
    =\arg\min_{\bql\in\feasl}
        \sum_{j=0}^s
            \int_{\ql_j}^{\ql_{j+1}}
                \sigma^2(r)\ud \overline \cdf(r),\nn
\end{align} where $\bql^*= [\ql_1^*,\ldots,\ql_s^*]^\top$ and $\overline \cdf(r)= \sum_{n=1}^N\gamma_n\cdf_n(r)$ is the weighted sum of the conditional CDFs with $\gamma_n=\|\vbf_n\|^2/\sum_{n=1}^N\|\vbf_n\|^2$.
Note that we can accommodate both normal and truncated normal distributions by 
substituting associated expressions into $\pdf_n(r)$ and $\cdf_n(r)$.  Exact 
update rules and analysis of computational complexity of ALQ, GD, and AMQ are discussed in \cref{app:norm}.

\section{Theoretical guarantees}
One can alternatively design quantization levels to minimize the worst-case variance. However, compared to an optimal scheme, this worst-case scheme increases the expected variance by $\Omega(d)$, which is prohibitive in deep networks. We quantify the gap in \cref{app:gap}. Proofs are in appendices.%

A stochastic gradient has a \textit{second-moment upper bound} $B$ when 
$\E[\|g(\wbf)\|_2^2]\leq B$ for all $\wbf\in \Omega$.  Similarly, it has 
a \textit{variance upper bound} $\sigma^2$ when $\E[\|g(\wbf)-\nabla 
\fun(\wbf)\|_2^2]\leq \sigma^2$ for all $\wbf\in \Omega$.

We consider a general adaptively quantized SGD (AQSGD) algorithm, described in \cref{AQSGDalg}, where compression schemes are updated 
over the course of training.\footnote{Our results hold for any adaptive method, including ALQ and AMQ.} Many convergence results in stochastic optimization rely on a variance bound. We establish such a variance bound for our adaptive methods. Further, we verify that these optimization results can be made to rely only on the average variance. In the following, we provide theoretical guarantees for AQSGD algorithm, obtain variance and code-length bounds, and convergence guarantees for convex, nonconvex, and momentum-based variants of AQSGD.

The analysis of nonadaptive methods in \citep{QSGD,NUQSGD,Samuel,ZipML} can  be considered as special cases of our theorems with fixed levels over the course of training. A naive adoption of available convergence guarantees results in having worst-case variance bounds over the course of training. In this paper, we show that an average variance bound can be applied on a number of problems. Under general normalization, we first obtain variance upper bound for arbitrary levels, in particular, for those obtained adaptively. 

\bth[Variance bound]\label{thm:varbound} 
Let $\vbf\in\reals^d$ and $q\geq 1$. The quantization of $\vbf$ under $L^q$ normalization satisfies $\E[Q_\bql(\vbf)]=\vbf$. Furthermore, we have
\begin{align}\label[ineq]{varbound}
\E[\|Q_\bql(\vbf)-\vbf\|_2^2]\leq\epsilon_Q\|\vbf\|_2^2,
\end{align} where $\epsilon_Q=\frac{(\ql_{j^*+1}/\ql_{j^*}-1)^2}{4(\ql_{j^*+1}/\ql_{j^*})}+\inf_{0<p<1}K_p{\ql_1}^{(2-p)}d^{\frac{2-p}{\min\{q,2\}}}$ with $j^*=\arg\max_{1\leq j\leq s} \ql_{j+1}/\ql_j$ and $K_p=\big(\frac{1}{2-p}\big)\big(\frac{1-p}{2-p}\big)^{(1-p)}$. 
\eth

\cref{thm:varbound} implies that if $g(\wbf)$ is a stochastic gradient with a second-moment bound $\eta$, then $Q_\bql(g(\wbf))$ is a stochastic gradient with a variance upper bound $\epsilon_Q\eta$. Note that, as long as the maximum ratio of two consecutive levels does not change, the variance upper bound decreases with the number of quantization levels. In addition, our bound matches the known $\Omega(\sqrt{d})$ lower bound in \citep{anonymous}. %

\bth[Code-length bound]\label{thm:codebound}%
Let $\vbf\in\reals^d$ and $q\geq 1$. The expectation $\E[|\ENCODE(\vbf)|]$
 of the number of communication bits needed to transmit $Q_\bql(\vbf)$ under $L^q$ normalization is bounded  by
\begin{align}\label[ineq]{codebound}\small
\E[|\ENCODE(\vbf)|]\leq b+n_{\ql_1,d}+d(H(L)+1)\leq b+n_{\ql_1,d}+d(\log_2(s+2)+1),  
\end{align} where $b$ is a constant, $n_{\ql_1,d}=\min\{{\ql_1}^{-q}+\frac{ d^{1-1/q}}{\ql_1},d\}$, $H(L)$ is the entropy of $L$ in bits, and $L$ is a random variable with the probability mass function given by 
\begin{align}\nn%
\Pr(\ql_j)=\int_{\ql_j-1}^{\ql_j}\frac{r-\ql_{j-1}}{\ql_{j}-\ql_{j-1}}\ud \cdf(r)+\int_{\ql_j}^{\ql_{j+1}}\frac{\ql_{j+1}-r}{\ql_{j+1}-\ql_{j}}\ud \cdf(r) 
\end{align} for $j=1,\ldots,s$. In addition, we have 
\begin{align}
\Pr(\ql_0=0)=\int_{0}^{\ql_1}\frac{1-r}{\ql_1}\ud \cdf(r)~\and~\Pr(\ql_{s+1}=1)=\int_{\ql_s}^{1}\frac{r-\ql_{s}}{1-\ql_{s}}\ud \cdf(r).\nn
\end{align}
\eth

\cref{thm:codebound} provides a bound on the expected number of communication bits to encode the quantized stochastic gradients. 
As expected, the upper bound in \cref{codebound} increases monotonically with $d$ and $s$. 

We can combine variance and code-length upper bounds and obtain convergence guarantees for AQSGD when applied to various learning problems where we have convergence guarantees for full-precision SGD under standard assumptions.  

Let $\{\bql_1,\ldots,\bql_K\}$ denote the set of quantization levels that AQSGD experiences on the optimization trajectory. Suppose that $\bql_k$ is used for $T_k$ iterations with $\sum_{k=1}^KT_k=T$. For each particular $\bql_k$, we can obtain corresponding variance bound $\epsilon_{Q,k}$ by substituting $\bql_k$ into \cref{varbound}. Then the average variance upper bound is given by $\overline{\epsilon_Q}=\sum_{k=1}^KT_k \epsilon_{Q,k}/T$. For each particular $\bql_k$, we can obtain corresponding expected code-length bound $N_{Q,k}$ by substituting random variable $L_k$ into \cref{codebound}. The average expected code-length bound is given by $\overline{N_Q}= \sum_{k=1}^KT_k N_{Q,k}/T $. 

On convex problems, convergence guarantees can be established along the lines of \citep[Theorems~6.1]{Bubeck}.%
\bth[AQSGD for nonsmooth convex optimization]\label{thm:convnonsmooth}
Let $\fun:\Omega\rightarrow \reals$ denote a convex function and let $R^2 \defeq \sup_{\wbf\in\Omega}\|\wbf-\wbf_0\|_2^2$.
 Let $\hat B=(1+\overline{\epsilon_Q}) B$ and $\fun^*=\inf_{\wbf\in\Omega}\fun(\wbf)$. Suppose that AQSGD is executed for $T$ iterations
with a learning rate $\alpha=RM/(\hat B\sqrt{T})$
on $M$ processors, each with access to independent stochastic gradients of $\fun$ with a second-moment bound $B$, such that quantization levels are updated $K$ times where $\bql_k$ with variance bound $\epsilon_{Q,k}$ and code-length bound $N_{Q,k}$ is used for $T_k$ iterations. Then AQSGD satisfies 
$\E\left[\fun\left(\frac{1}{T}\sum_{t=0}^T\wbf_t\right)\right]-\fun^*\leq R\hat B/(M\sqrt{T})$. 

In addition, AQSGD requires at most $\overline{N_Q}$ communication bits per iteration in expectation. 
\eth 
In \cref{app:nonconvex} and \cref{app:momentum}, we obtain convergence guarantees on nonconvex problems and for momentum-based variants of AQSGD under standard assumptions, respectively. Theoretical guarantees for levels with symmetry are established in \cref{app:theorysym}.

\section{Experimental evaluation}\label{sec:exp}
\begin{table}
    \centering
    \caption{Validation accuracy on CIFAR-10 and
    ImageNet using $3$ bits (except for \supersgd and \terngrad) with $4$ GPUs.}
    \begin{tabular}{C{3cm}|C{2.5cm}C{2.5cm}|C{2.5cm}}
\toprule
       Quantization Method          & ResNet-110 on CIFAR-10      & ResNet-32 on CIFAR-10       & ResNet-18 on ImageNet       \\
\midrule
        Bucket Size                 & $16384$                     & $8192$                      & $8192$                      \\
\midrule
         \supersgd                  & \textbf{93.86\% $\pm$ 0.08} & \textbf{92.26\% $\pm$ 0.04} & \textbf{68.93\% $\pm$ 0.05} \\
\midrule
         \nuq~\citep{NUQSGD,Samuel} & 84.60\% $\pm$ 0.04          & 83.73\% $\pm$ 0.08          & 33.36\% $\pm$ 0.07          \\
         \qinf~\citep{QSGD}         & 91.52\% $\pm$ 0.07          & 89.95\% $\pm$ 0.02          & 66.35\% $\pm$ 0.04          \\
         \terngrad~\citep{TernGrad} & 90.72\% $\pm$ 0.06          & 89.65\% $\pm$ 0.05          & 62.76\% $\pm$ 0.06          \\
\midrule
         \alqnbased                 & \textbf{93.24\% $\pm$ 0.06} & 91.30\% $\pm$ 0.07          & \textbf{67.72\% $\pm$ 0.07} \\
         \alqnless                  & \textbf{93.14\% $\pm$ 0.05} & \textbf{91.96\% $\pm$ 0.04} & 65.64\% $\pm$ 0.07          \\
         \amqnbased                 & \textbf{92.82\% $\pm$ 0.04} & 91.10\% $\pm$ 0.05          & 64.82\% $\pm$ 0.05          \\
         \amqnless                  & \textbf{92.88\% $\pm$ 0.02} & 91.03\% $\pm$ 0.08          & 66.75\% $\pm$ 0.05          \\
\bottomrule
    \end{tabular}
    \label{tab:vacc}
\end{table}

In this section, we showcase the effectiveness of our adaptive quantization 
methods in speeding up training deep models.  We compare our methods to the 
following baselines:
single-GPU SGD (\sgd),
full-precision multi-GPU SGD (\supersgd),
uniform levels under $L^\infty$ normalization (\qinf)~\citep{QSGD},
ternary levels under $L^\infty$ normalization (\terngrad)~\citep{TernGrad},
and exponential levels under $L^2$ normalization with exponential factor 
$p=0.5$ (\nuq)~\citep{NUQSGD,Samuel}.
We present results for the following variations of our proposed methods:
\alqnbased and \amqnbased (with norm adjustments in \cref{body:norm}),
and their normalized variations \alqnless and \amqnless 
(\cref{sec:alq,sec:amq}). We present full training results on ImageNet in 
\cref{app:exp} along with additional experimental details.

We compare methods in terms of the number of training iterations that is 
independent of a particular distributed setup.
In \cref{tab:vacc}, we present results for training ResNet-32 and 
ResNet-110~\citep{Resnet} on CIFAR-10~\citep{CIFAR10},
and ResNet-18 on ImageNet~\citep{ImageNet}.  We simulate training with $4$-GPUs 
on a single GPU by quantizing and dequantizing the gradient from $4$ 
mini-batches in each training iteration.  These simulations allow us to compare 
the performance of quantization methods to the hypothetical full-precision 
{\supersgd}.

All quantization methods studied in this section share two hyper-parameters:
the number of bits ($\log_2$ of number of quantization levels)
and a bucket size.
A common trick used in normalized quantization is to encode and decode 
a high-dimensional vector in buckets such that each coordinate is normalized by 
the norm of its corresponding bucket instead of the norm of the entire 
vector~\citep{QSGD}. The bucket size controls the tradeoff between extra 
communication cost and loss of precision.
With a small bucket size, there are more bucket norms to be communicated, while 
with a large bucket size, we lose numerical precision as a result of dividing 
each coordinate by a large number. In \cref{sec:hparams}, we provide an 
empirical study of the hyperparameters.

{\def\EXPNAME{-g/Vloss}\def\FIGTITLE{Validation loss}

\begin{figure}[t]
    \centering
    \begin{subfigure}[b]{\threecolfigwidth}
        \includegraphics[width=\textwidth]{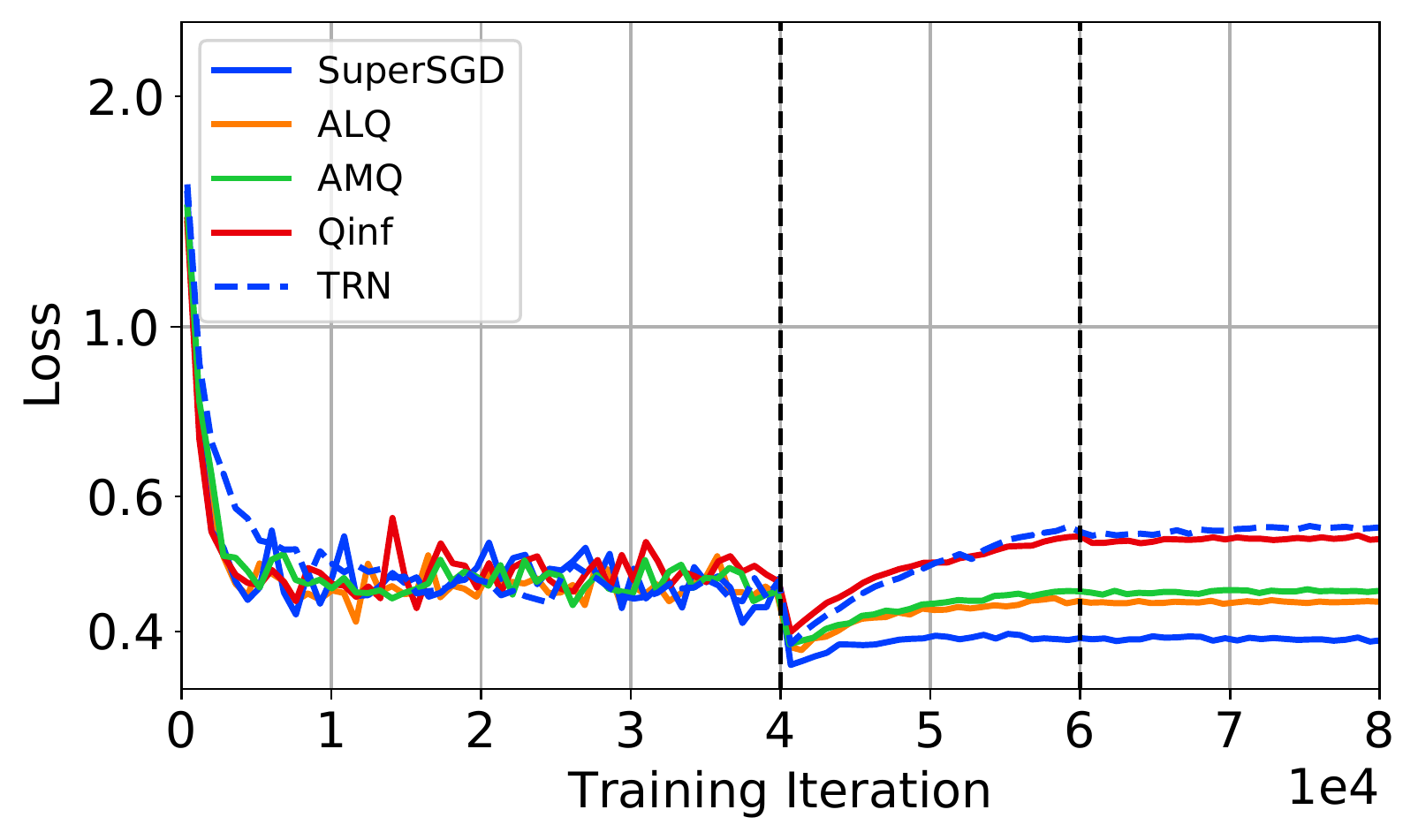}
        \vspace*{\capshift}
        \caption{ResNet-32 on CIFAR-10}
        \label{fig:cifar10_resnet32_\EXPNAME}
    \end{subfigure}
    \begin{subfigure}[b]{\threecolfigwidth}
        \includegraphics[width=\textwidth]{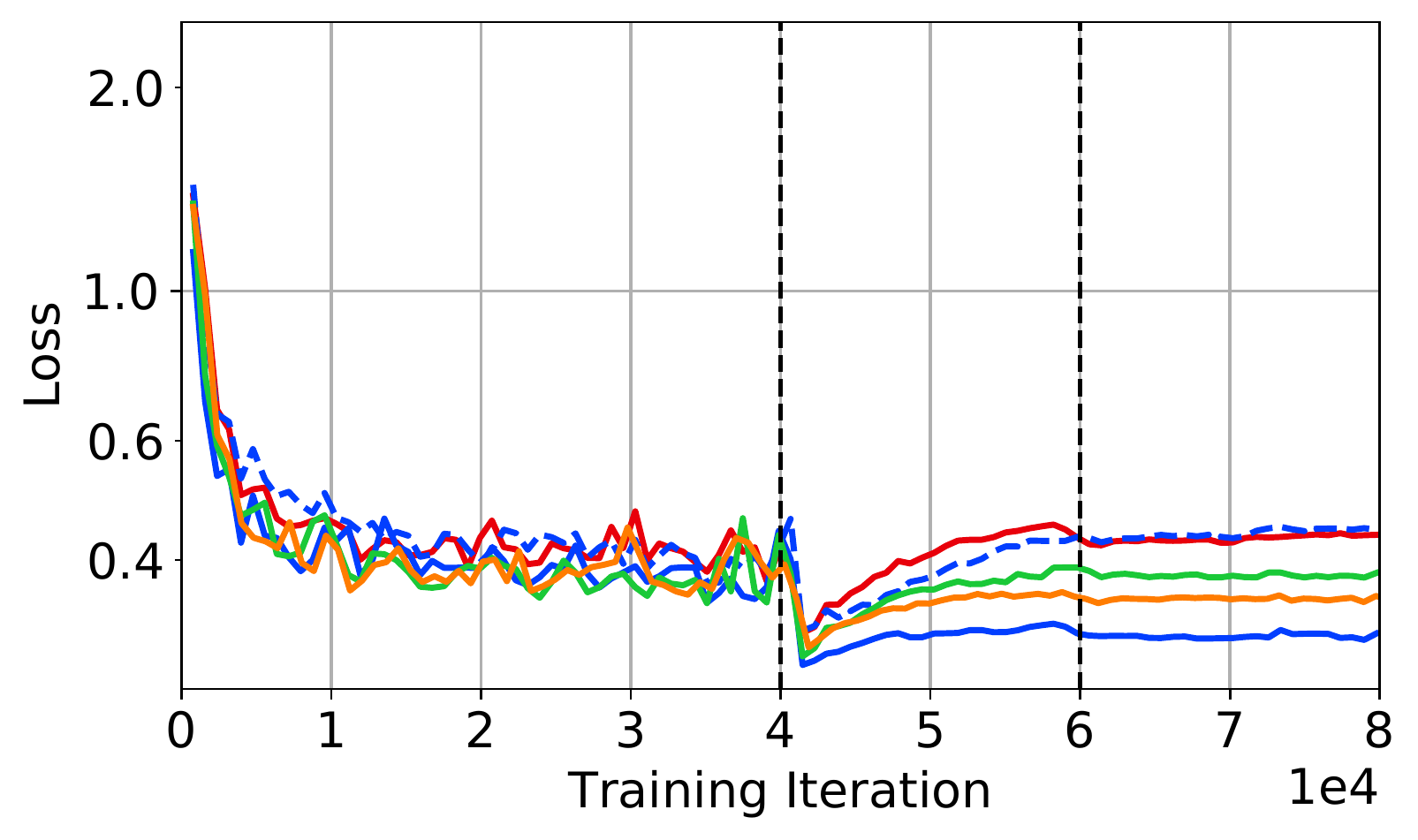}
        \vspace*{\capshift}
        \caption{ResNet-110 on CIFAR-10}
        \label{fig:cifar10_resnet110_\EXPNAME}
    \end{subfigure}
    \begin{subfigure}[b]{\threecolfigwidth}
        \includegraphics[width=\textwidth]{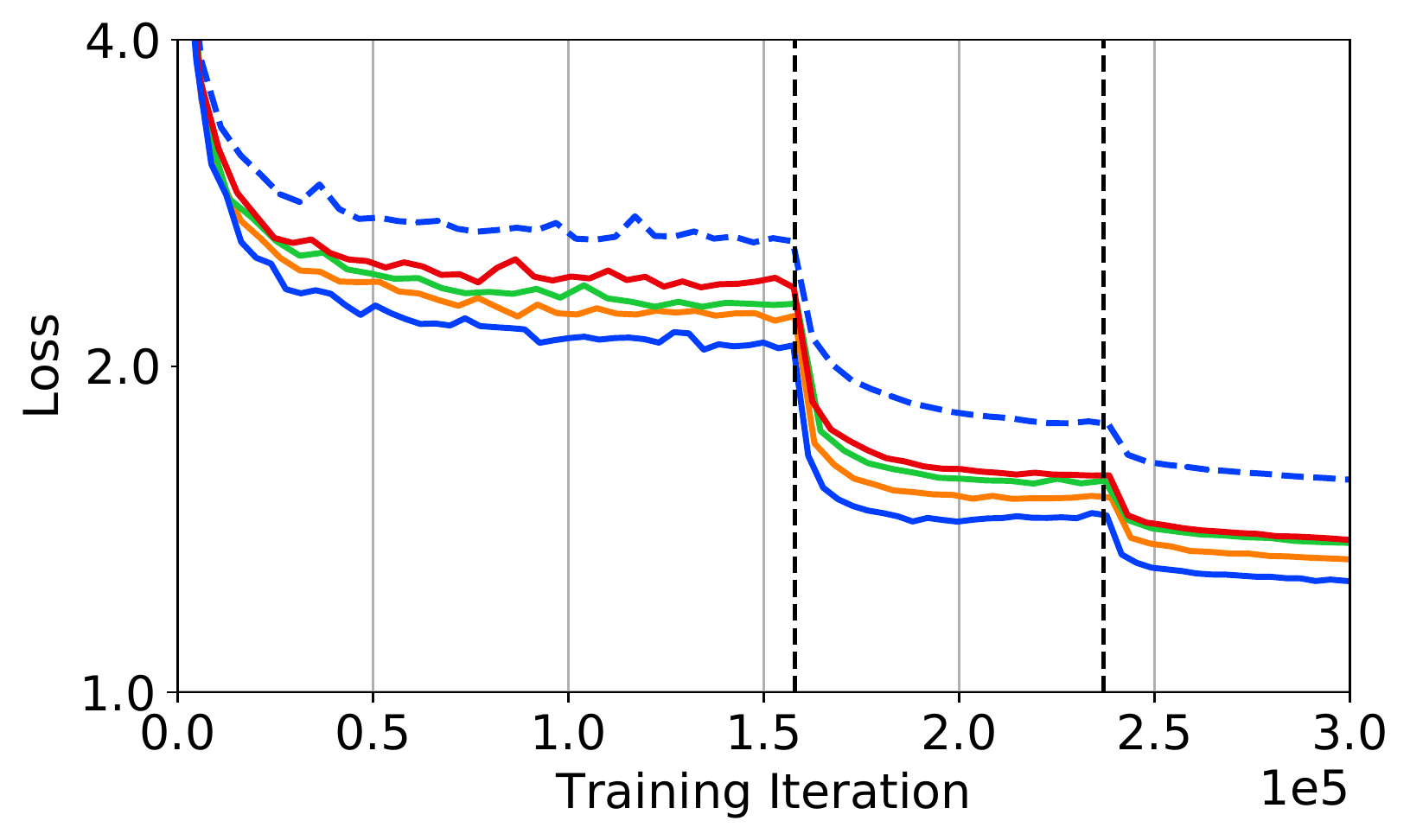}
        \vspace*{\capshift}
        \caption{ResNet-18 on ImageNet}
        \label{fig:imagenet_\EXPNAME}
    \end{subfigure}
    \caption{\textbf{{\FIGTITLE}} on CIFAR-10 and ImageNet. All methods use $3$ 
    bits except for \supersgd and \terngrad. Bucket size for ResNet-110 trained 
    on CIFAR-10 is
    $16384$,  for ResNet-32 is $8192$, and for ResNet-18 on ImageNet is 
    $8192$.}
\label{fig:\EXPNAME}
\end{figure}
}
{\def\EXPNAME{-g/est_var}\def\FIGTITLE{Variance}

\begin{figure}[t]
    \centering
    \begin{subfigure}[b]{\threecolfigwidth}
        \includegraphics[width=\textwidth]{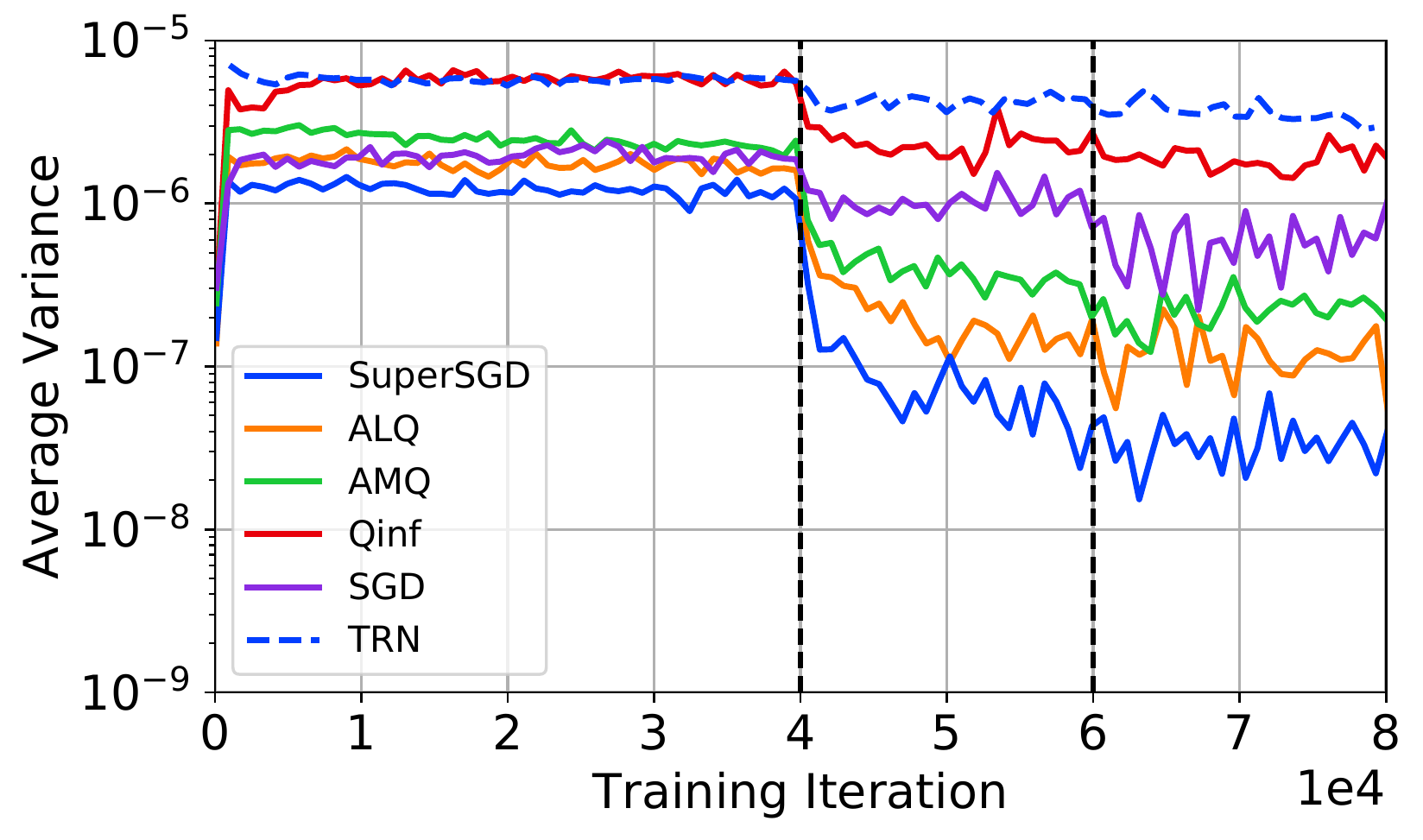}
        \vspace*{\capshift}
        \caption{ResNet-32 on CIFAR-10}
        \label{fig:cifar10_resnet32_\EXPNAME}
    \end{subfigure}
    \begin{subfigure}[b]{\threecolfigwidth}
        \includegraphics[width=\textwidth]{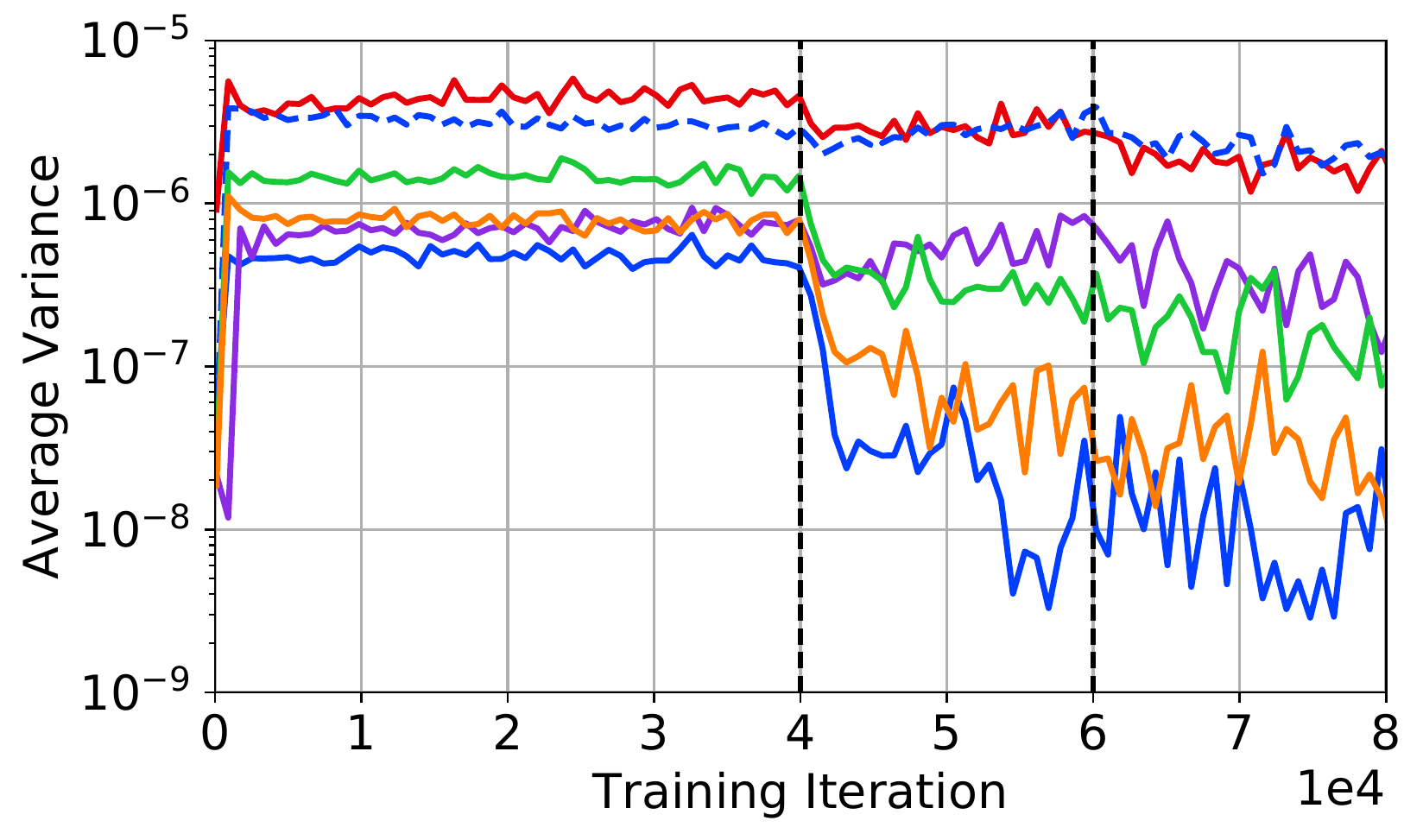}
        \vspace*{\capshift}
        \caption{ResNet-110 on CIFAR-10}
        \label{fig:cifar10_resnet110_\EXPNAME}
    \end{subfigure}
    \begin{subfigure}[b]{\threecolfigwidth}
        \includegraphics[width=\textwidth]{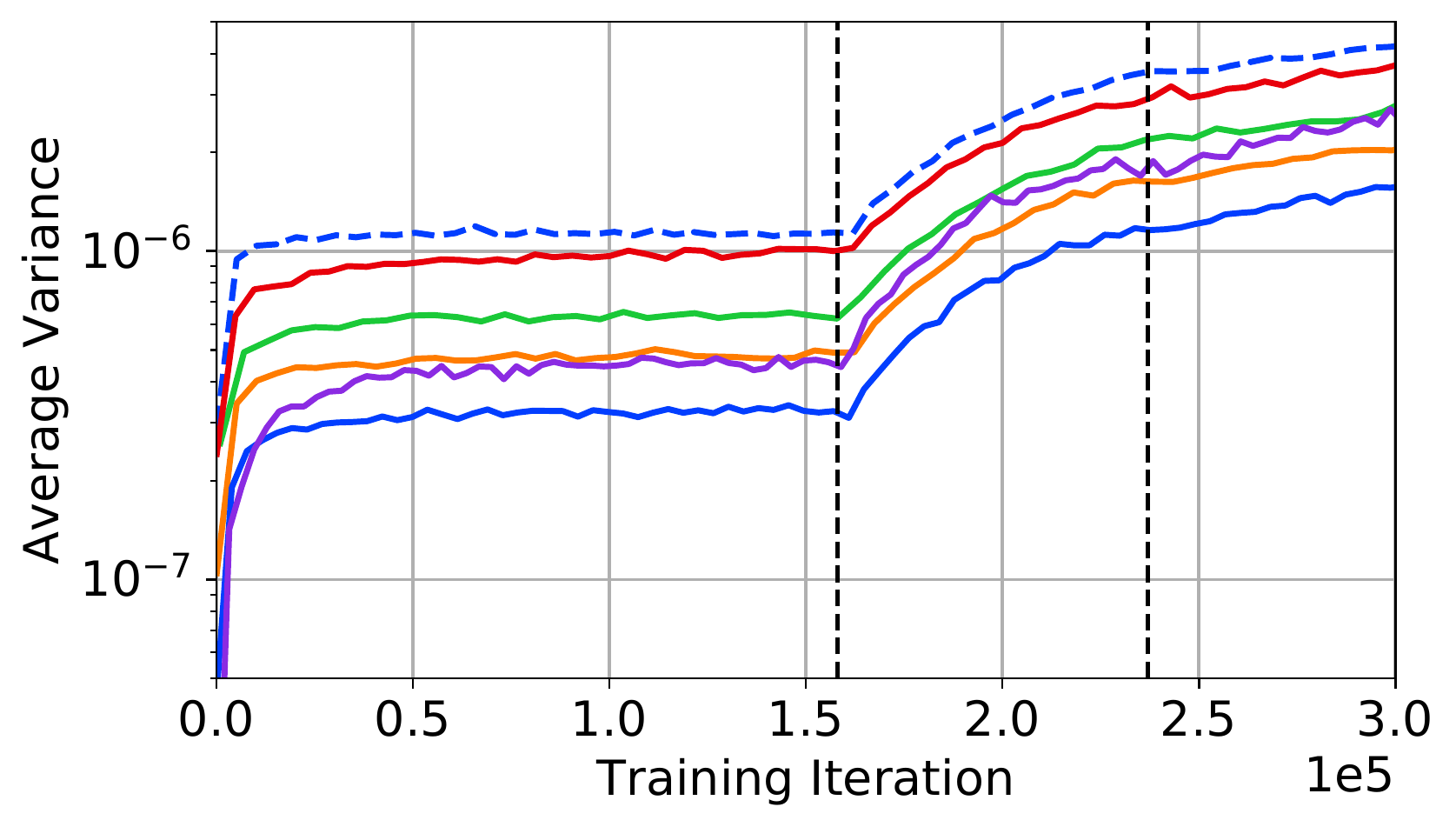}
        \vspace*{\capshift}
        \caption{ResNet-18 on ImageNet}
        \label{fig:imagenet_\EXPNAME}
    \end{subfigure}
    \caption{\textbf{{\FIGTITLE}} on CIFAR-10 and ImageNet. All methods use $3$ 
    bits except for \supersgd and \terngrad. Bucket size for ResNet-110 trained 
    on CIFAR-10 is
    $16384$,  for ResNet-32 is $8192$, and for ResNet-18 on ImageNet is 
    $8192$.}
\label{fig:\EXPNAME}
\end{figure}
}
{\def\EXPNAME{/est_var}\def\FIGTITLE{Variance (no train)}

\begin{figure}[t]
    \centering
    \begin{subfigure}[b]{\threecolfigwidth}
        \includegraphics[width=\textwidth]{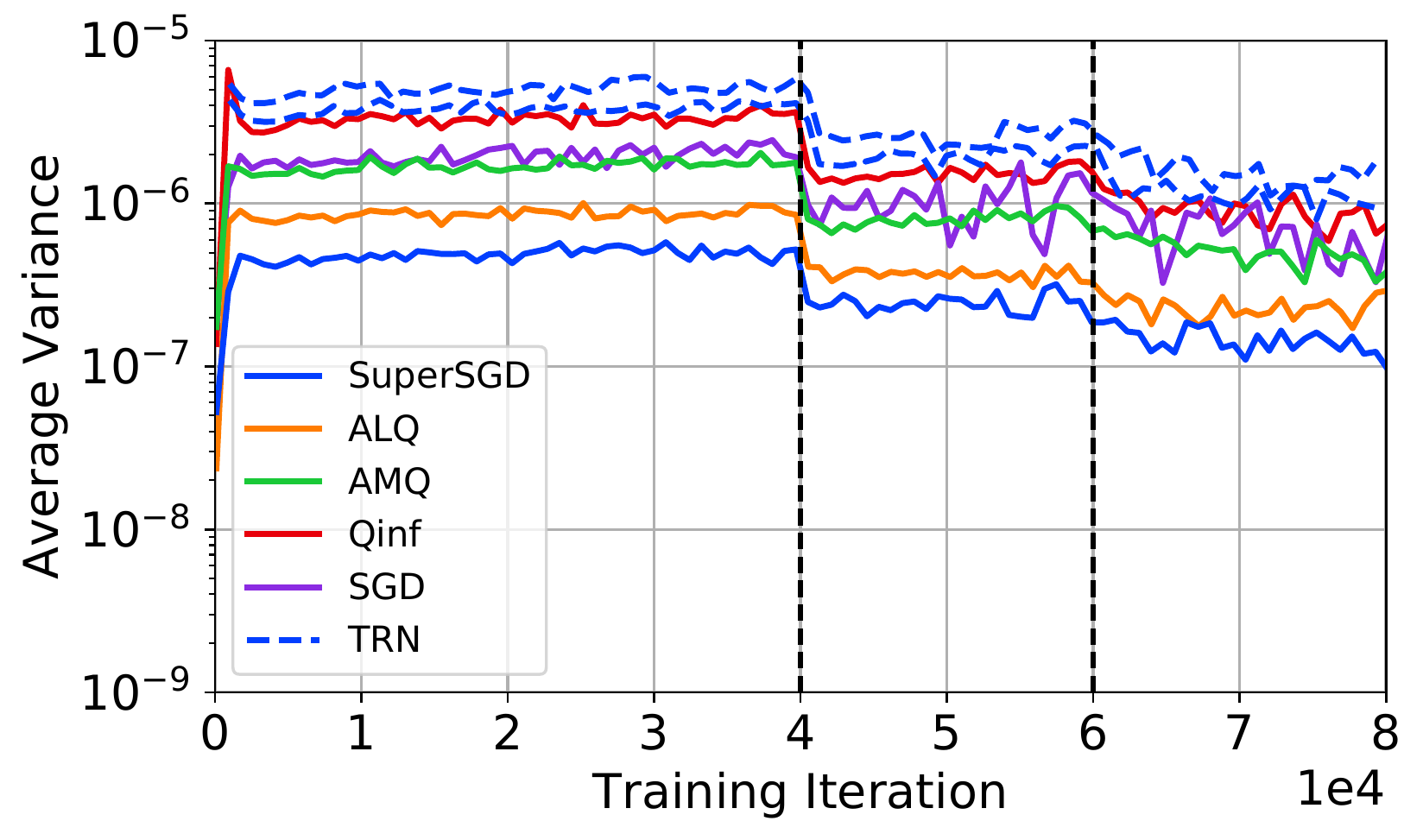}
        \vspace*{\capshift}
        \caption{ResNet-32 on CIFAR-10}
        \label{fig:cifar10_resnet32_\EXPNAME}
    \end{subfigure}
    \begin{subfigure}[b]{\threecolfigwidth}
        \includegraphics[width=\textwidth]{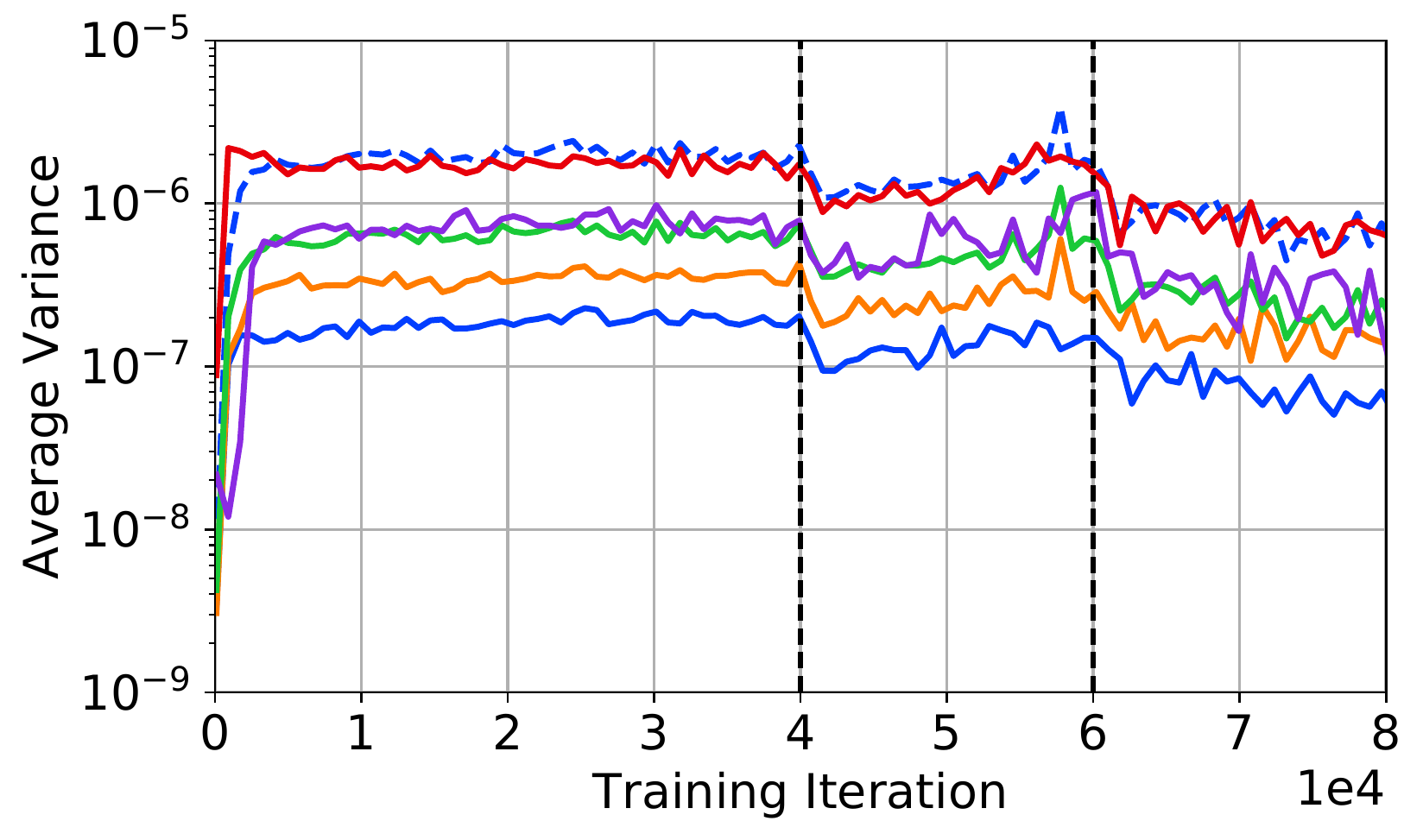}
        \vspace*{\capshift}
        \caption{ResNet-110 on CIFAR-10}
        \label{fig:cifar10_resnet110_\EXPNAME}
    \end{subfigure}
    \begin{subfigure}[b]{\threecolfigwidth}
        \includegraphics[width=\textwidth]{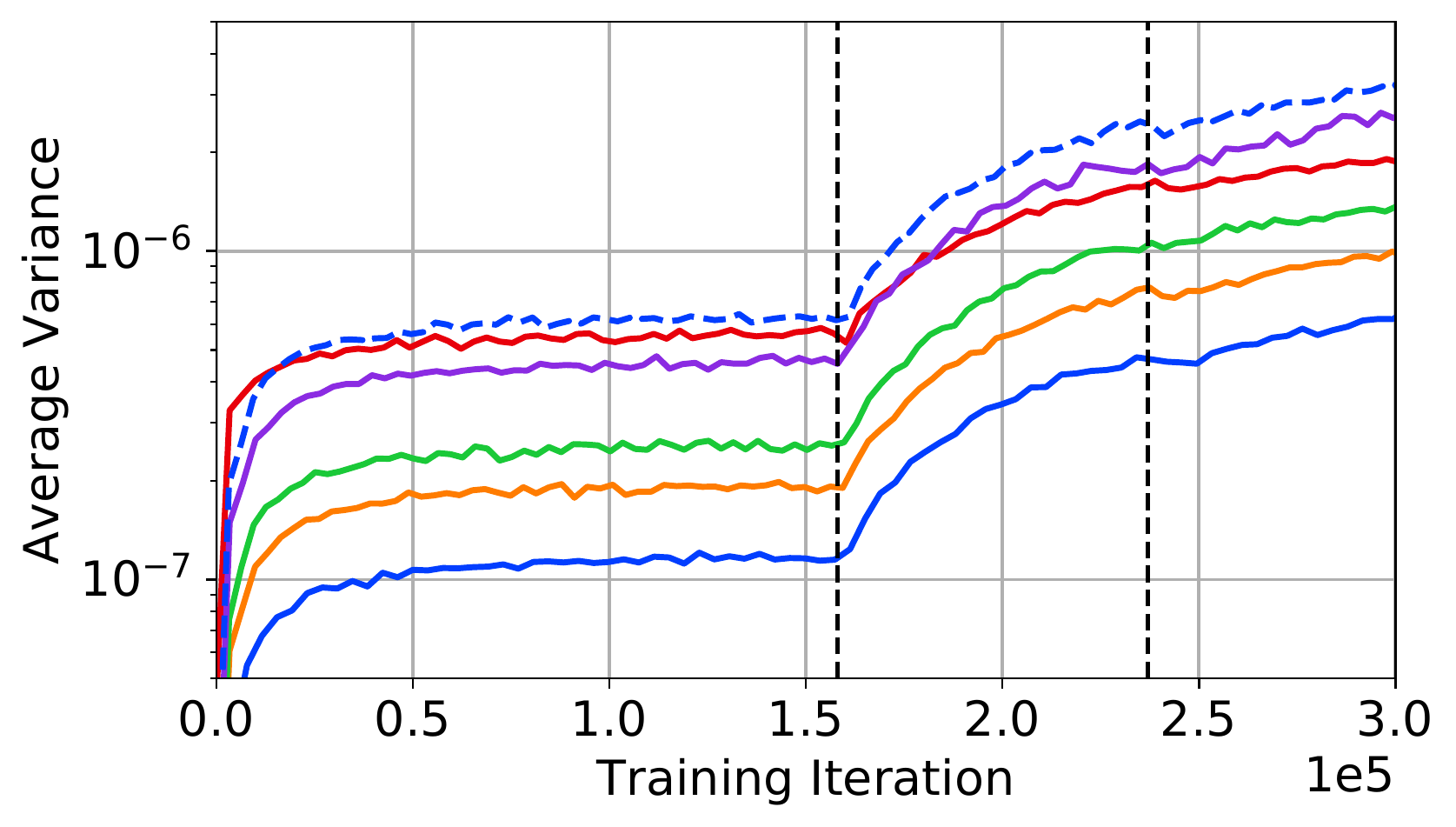}
        \vspace*{\capshift}
        \caption{ResNet-18 on ImageNet}
        \label{fig:imagenet_\EXPNAME}
    \end{subfigure}
    \caption{\textbf{{\FIGTITLE}} on CIFAR-10 and ImageNet. All methods use $3$ 
    bits except for \supersgd and \terngrad. Bucket size for ResNet-110 trained 
    on CIFAR-10 is
    $16384$,  for ResNet-32 is $8192$, and for ResNet-18 on ImageNet is 
    $8192$.}
\label{fig:\EXPNAME}
\end{figure}
}

\textbf{Matching the accuracy of \supersgd.}
Using only $3$ bits ($8$ levels), our adaptive methods match the performance of 
\supersgd on CIFAR-10 and close the gap on ImageNet (bold in \cref{tab:vacc}).
Our most flexible method, \alqnbased, achieves the best overall performance on 
ImageNet and the gap on CIFAR-10 with \alqnless is less than $0.3\%$.
There is at least $1.4\%$ gap between our best performing method and previous 
work in training each model.
To the best of our knowledge, matching the validation loss of \supersgd has not 
been achieved in any previous work using only $3$ bits. \cref{fig:-g/Vloss} 
shows the test loss and \cref{fig:-g/est_var} shows the average gradient 
variance where the average is taken over gradient coordinates. Our adaptive 
methods successfully achieve lower variance during training.

\textbf{Comparison on the trajectory of SGD.}
\cref{fig:/est_var} shows the average variance on the optimization
trajectory of single-GPU without quantization. This graph provides a more fair 
comparison of the quantization error of different methods decoupled from their 
impact on the optimization trajectory.
\alqnbased effectively finds an improved set of levels that reduce the variance 
in quantization. \alqnbased matches the variance of \supersgd on Resnet-110 
(\cref{fig:cifar10_resnet110_/est_var}). In 
\cref{fig:cifar10_resnet110_/est_var,fig:imagenet_/est_var}, the variance of 
\qinf is as high as \terngrad in the first half of training. This shows that 
extra levels ($8$ uniform levels) do not perform better unless designed 
carefully.
As expected, the variance of \supersgd is always smaller than the variance of 
\sgd by a constant factor of the number of GPUs. 

\textbf{Negligible computational overhead.}
Our adaptive methods have similar per-step computation and communication cost 
compared to previous methods.
On ImageNet, we save at least $60$ hours from $95$ hours of training and add 
only an additional cost of at most 10 minutes in total to adapt quantization.  
For bucket sizes $8192$ and $16384$ and $3$--$8$ bits used in our experiments, 
the per-step cost relative to SuperSGD ($32$-bits) is $21$--$25\%$ for 
ResNet-18 on ImageNet and $32$--$36\%$ for ResNet-50.  That is the same as the 
cost of NUQSGD and QSGDinf without additional coding or pruning with the same 
number of bits and bucket sizes.  The cost of the additional update specific to 
ALQ is $0.4$--$0.5\%$ of the total training time.  In \cref{sec:timing}, we 
provide tables with detailed timing results for varying bucket sizes and bits.

\subsection{Hyperparameter studies}\label{sec:hparams}

 \begin{figure}[t]
     \centering
     \begin{minipage}{.32\textwidth}
            \centering
            \begin{subfigure}[b]{\textwidth}
                \includegraphics[width=\textwidth]{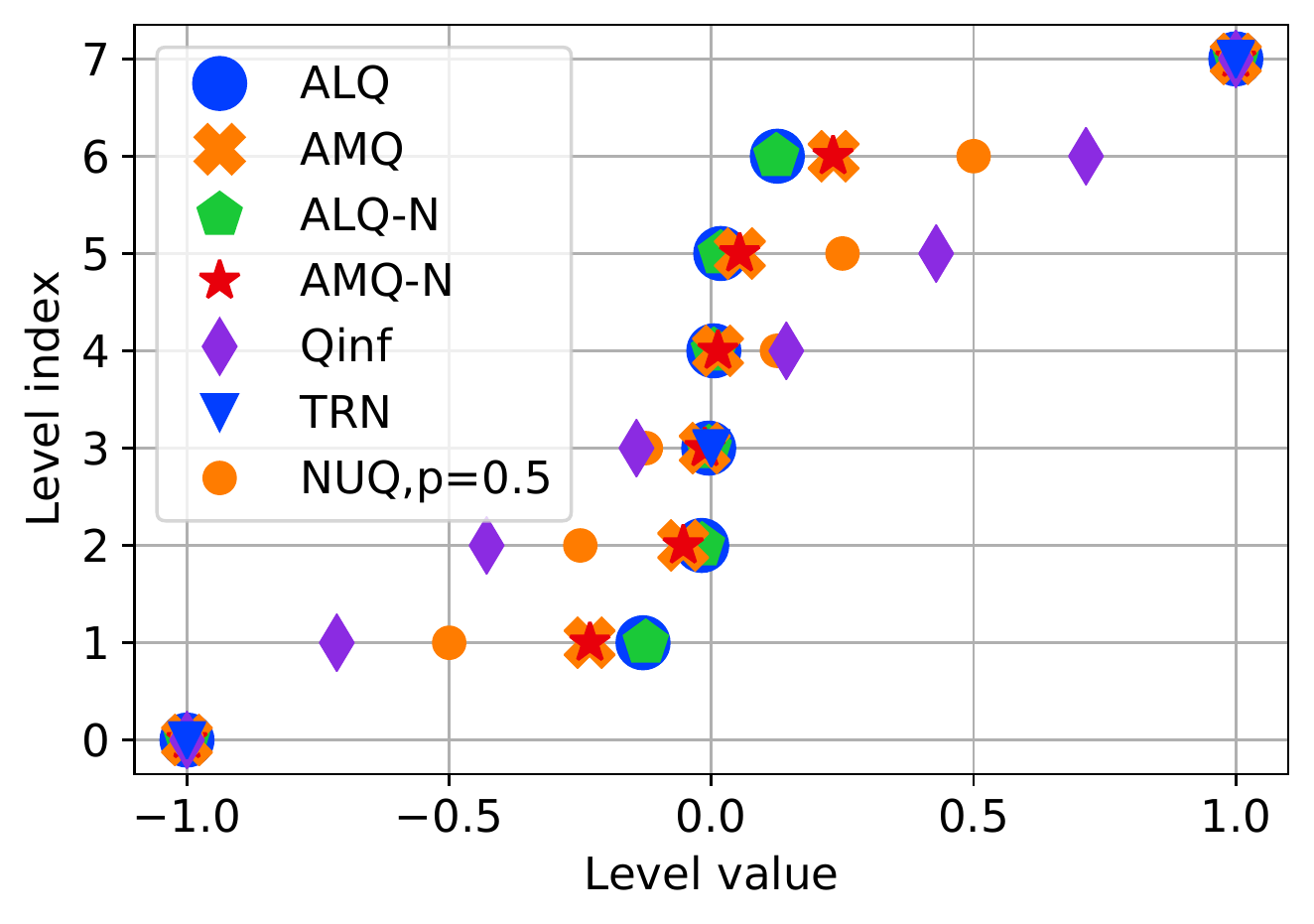}
                \vspace*{\capshift}

            \end{subfigure}
            \caption{Quantization levels at the end of training ResNet-32 on 
                CIFAR-10.}
            \label{fig:levels}
    \end{minipage}
         \hfill
     \begin{minipage}{.64\textwidth}
    \centering
    \begin{subfigure}[b]{0.49\textwidth}
        \includegraphics[width=\textwidth]{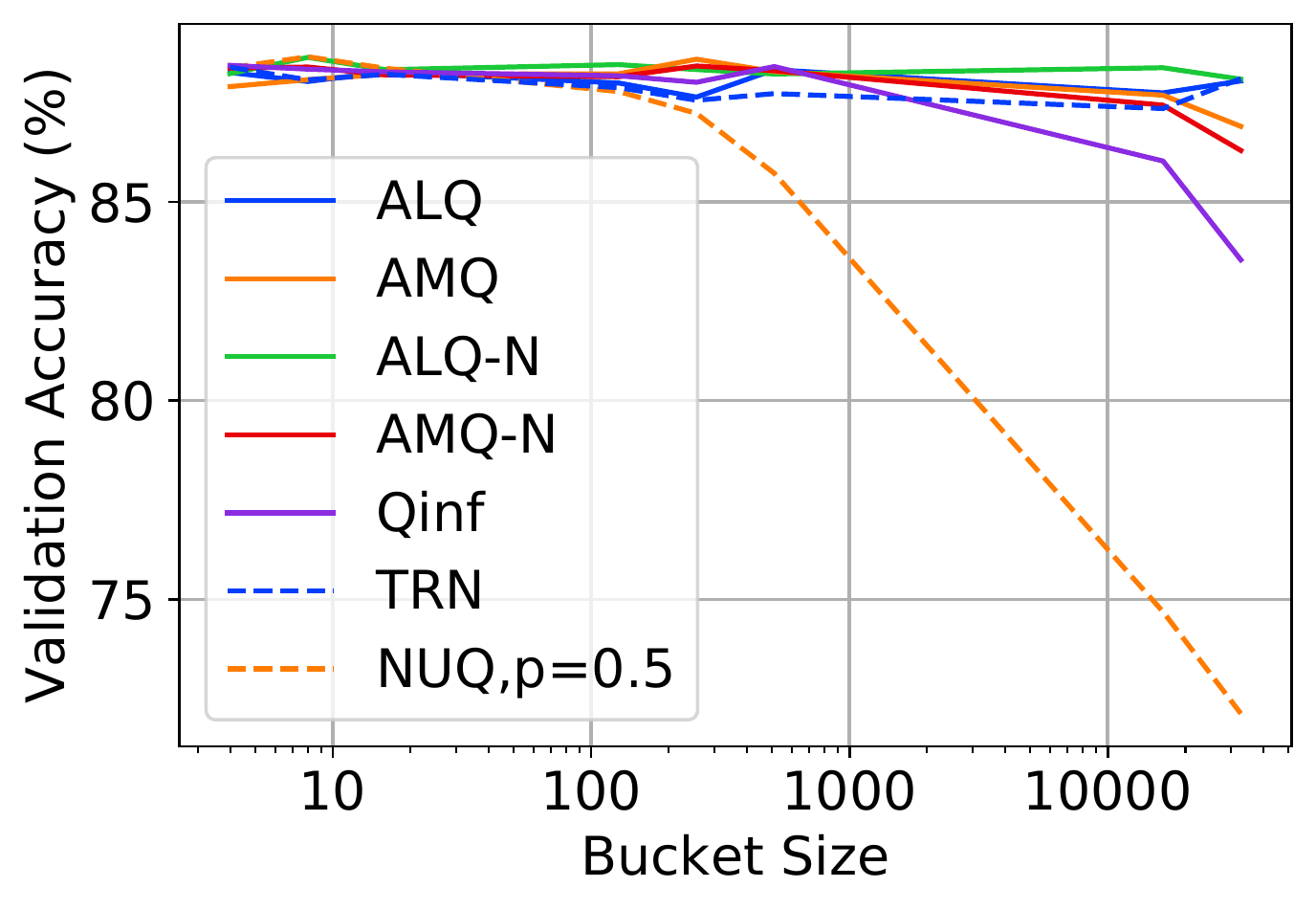}
        \vspace*{\capshift}
        \caption{Bucket Size (bits=$3$)}
        \label{fig:hp_bucket_size_vacc}

    \end{subfigure}
        \hfill
    \begin{subfigure}[b]{0.49\textwidth}
        \includegraphics[width=\textwidth]{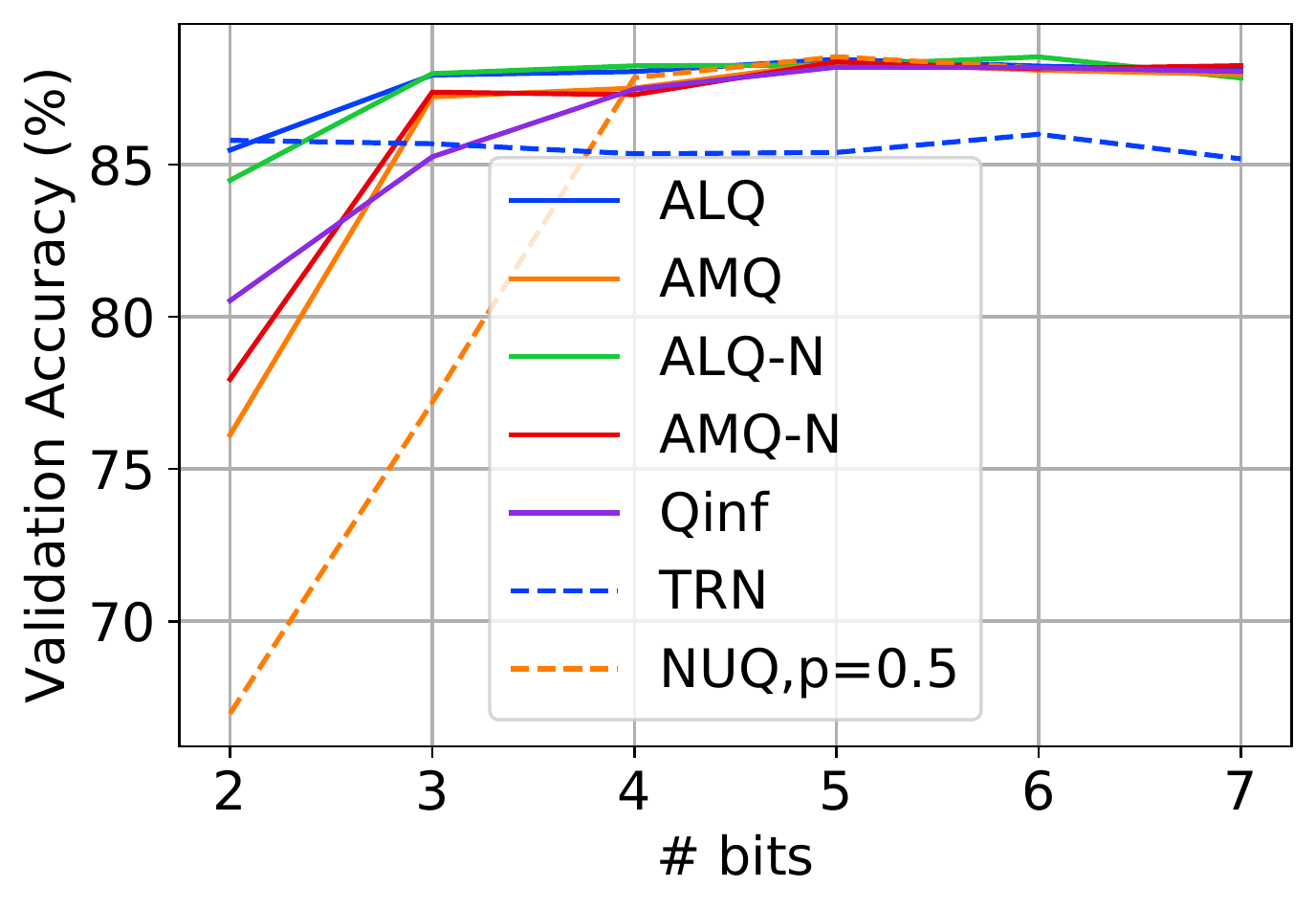}
        \vspace*{\capshift}
        \caption{Bits (bucket size=$16384$)}
        \label{fig:hp_bits_vacc}
    \end{subfigure}
    \caption{Effect of bucket size and number of bits on validation accuracy
    when training ResNet-8 on CIFAR-10}
    \label{fig:hp}
    \end{minipage}
\end{figure}

\cref{fig:levels} shows quantization levels for each method at the end of 
training ResNet-32 on CIFAR-10. The quantization levels for our adaptive 
methods are more concentrated near zero.
In \cref{fig:hp_bucket_size_vacc,fig:hp_bits_vacc}, we study the impact of the 
bucket size and number of bits on the best validation accuracy achieved by 
quantization methods.

\textbf{Adaptive levels are the best quantization methods across all values of 
bucket size and number of bits.}
\alqnbased and \alqnless are the best performing methods across all values of 
bucket size and number of bits. The good performance of \alqnless is unexpected 
as it suggests quantization for vectors with different norms can be shared.  In practice, 
\alqnless is easier to implement and faster to update compared to \alqnbased.  
We observe a similar relation between \amqnbased and \amqnless methods.  
Adaptive multiplier methods show inferior performance to adaptive level methods 
as the bucket size significantly grows (above $10^4$) or shrinks (below $100$) 
as well as for very few bits ($2$). Note that there exists a known 
generalization gap between \sgd and \supersgd in ResNet-110 that can be closed 
by extensive hyperparameter tuning~\citep{choi2019empirical}. Our adaptive 
methods reduce this gap with standard hyperparameters.

\textbf{Bucket size significantly impacts non-adaptive methods.}
For bucket size $100$ and $3$ bits, \nuq performs nearly as good as adaptive 
methods but quickly loses accuracy as the bucket size grows or shrinks. \qinf 
stays competitive for a wider range of bucket sizes but still loses accuracy 
faster than other methods. This shows the impact of bucketing as an 
understudied trick in evaluating quantization methods.

\textbf{Adaptive methods successfully scale to large number of GPUs.} 
\cref{tab:gpus} shows the result of training CIFAR-10 on ResNet-32 using 16 and 
32 GPUs. Note that with 32 GPUs, \terngrad is achieving almost the accuracy of 
   \supersgd with only 3 quantization levels, which is expected because \terngrad 
   is unbiased and the variance of aggregated gradients decreases linearly with 
   the number of GPUs.

\begin{table}
    \centering
    \caption{Validation accuracy of ResNet32 on CIFAR-10 using $3$ quantization 
    bits (except for SuperSGD and TRN) and bucket size $16384$.}
     \begin{tabular}{C{1.5cm}|C{2.8cm}|C{2.8cm}}%
\toprule
Method          & 16 GPUs& 32 GPUs\\
\midrule
\midrule
\supersgd           & \textbf{92.17\% $\pm$ 0.08}  & \textbf{92.19\% $\pm$ 0.04} \\
\midrule
        \nuq        & 85.82\% $\pm$ 0.03           & 86.36\% $\pm$ 0.01 \\
        \qinf       & 89.61\% $\pm$ 0.03           & 89.81\% $\pm$ 0.05 \\
        \terngrad   & 88.68\% $\pm$ 0.10           &  90.22\% $\pm$ 0.05 \\
\midrule
         \alqnbased & \textbf{91.91\% $\pm$ 0.06}  & \textbf{91.89\% $\pm$ 0.07}          \\
         \alqnless  & \textbf{92.07\% $\pm$ 0.04}  & \textbf{91.83\% $\pm$ 0.03}          \\
         \amqnbased & 91.58\% $\pm$ 0.05           & 91.38\% $\pm$ 0.06 \\
         \amqnless  & 91.41\% $\pm$ 0.08           & 91.40\% $\pm$ 0.02 \\
\bottomrule
    \end{tabular}
     \label{tab:gpus}
\end{table}

\section{Conclusions}
To reduce communication costs of data-parallel SGD, we introduce two adaptively quantized methods, ALQ and AMQ, to learn and adapt gradient quantization method on the fly.  In addition to quantization method, in both methods, processors learn and adapt their coding methods in parallel by efficiently computing sufficient statistics of a parametric distribution. We establish tight upper bounds on the excessive variance for any arbitrary sequence of quantization levels under general normalization and on the expected number of communication bits per iteration. Under standard assumptions, we establish a number of convergence guarantees for our adaptive methods. We demonstrate the superiority of ALQ and AMQ over nonadaptive methods empirically on deep models and large datasets.

\section*{Broader impact}

This work provides additional understanding of statistical behaviour of deep machine learning models. We aim to train deep models using popular SGD algorithm as fast as possible without compromising
learning outcome. As the amount of data gathered through web and a plethora
of sensors deployed everywhere (e.g., IoT applications) is
drastically increasing, the design of efficient machine learning
algorithms that are capable of processing large-scale data
in a reasonable time can  improve everyone's quality of life. Our compression schemes can be used in Federated Learning settings, where a deep model is trained on data distributed among multiple owners without exposing that data.  
Developing privacy-preserving learning algorithms is an integral part of 
responsible and ethical AI. However, the long-term impacts of our schemes may depend on how machine learning is used in society.

\section*{Acknowledgement}
The authors would like to thank Blair Bilodeau, David Fleet, Mufan Li, and Jeffrey Negrea for helpful discussions. FF was supported by 
OGS Scholarship. DA and IM were supported the European Research Council (ERC) under the European Union's Horizon 2020 research and innovation programme (grant agreement No 805223 ScaleML). DMR was supported by an NSERC Discovery Grant. ARK was supported by NSERC Postdoctoral Fellowship. Resources used in preparing this research were provided, in 
part, by the Province of Ontario, the Government of Canada through CIFAR, and 
companies sponsoring the Vector Institute.\footnote{\url{www.vectorinstitute.ai/\#partners}}

\bibliographystyle{unsrtnat}
\bibliography{RefL2Q}

\newpage
\appendix

\section{CDF and its inverse}\label{app:cdf}

\subsection{Normal distribution}\label{sec:normal}
The probability density function (PDF) for $X\sim \Nc(\mu,\sigma^2)$ is defined 
as
\begin{align}\label{Normalpdf}
\pdf_{\Nc}(x)
    =\frac{1}{\sqrt{2\pi \sigma^2}}\exp\Big(\frac{-(x-\mu)^2}{2\sigma^2}\Big),
\end{align}
and the cumulative distribution function (CDF) defined as
\begin{align}\label{CDFnormal}
\cdf_{\Nc}(x)
    =\Phi\Big(\frac{x-\mu}{\sigma}\Big)=1-Q\Big(\frac{x-\mu}{\sigma}\Big)
    =\frac{1}{2}\Big(1+\erf\Big(\frac{x-\mu}{\sqrt{2}\sigma}\Big)\Big),
\end{align} where 
\begin{align}
\Phi(x)&=\int_{-\infty}^x \frac{1}{\sqrt{2\pi}}\exp(-u^2/2)\ud u,\nn\\
Q(x)&=\int_x^{\infty} \frac{1}{\sqrt{2\pi}}\exp(-u^2/2)\ud u,\nn\\
\erf(x)&=2\int_0^{x} \frac{1}{\sqrt{\pi}}\exp(-u^2)\ud u.\nn
\end{align}

The inverse of CDF for the normal distribution is given by 
\begin{align}\label{CDFinvnormal}
\cdf_{\Nc}^{-1}(y)=\sigma \Phi^{-1}(y)+\mu=\sqrt{2}\sigma\erf^{-1}(2y-1)+\mu. 
\end{align} %

Various approximations of \cref{CDFnormal} and \cref{CDFinvnormal} are available in the literature. 

\subsection{Truncated normal distribution}\label{sec:Tnormal}

The probability density function (PDF) of a truncated normal distribution that 
lies within the interval $(a,b)$ with $-\infty<a<b<\infty$ is defined as
\begin{align}
    \pdf_{\Tc}(x;a,b)= \frac{\pdf_{\Nc}(x)}{\sigma\big(\cdf_{\Nc}(b)-\cdf_{\Nc}(a)\big)},
\end{align}
where $\pdf_{\Nc}(\cdot)$ is defined in \cref{Normalpdf} and the cumulative 
distribution function (CDF) is defined as
\begin{align}
\begin{split}
\cdf_{\Tc}(x;a,b)&=\frac{\cdf_{\Nc}(x)-\cdf_{\Nc}(a)}{\cdf_{\Nc}(b)-\cdf_{\Nc}(a)}\\
&=\frac{\Phi((x-\mu)/\sigma)-\Phi((a-\mu)/\sigma)}{\Phi((b-\mu)/\sigma)-\Phi((a-\mu)/\sigma)},
\end{split} 
\end{align} where $\cdf_{\Nc}(\cdot)$ and $\Phi(\cdot)$ are defined in \cref{CDFnormal}. Note that the \textit{mean} and \textit{variance} of a random variable with truncated normal distribution are not $\mu$ and $\sigma^2$ based on our notation. The mean and variance depend on the interval $(a,b)$, which is clear in contexts that we use. 

The inverse of CDF for truncated normal distribution is given by 
\begin{align}\label{CDFinvtnormal}
\cdf_{\Tc}^{-1}(y;a,b)=\cdf_{\Nc}^{-1}(\overline y)=\sigma \Phi^{-1}(\overline y)+\mu, 
\end{align} where $\overline y=\big(\cdf_{\Nc}(b)-\cdf_{\Nc}(a)\big)y+\cdf_{\Nc}(a)$ and $\cdf_{\Nc}^{-1}(\cdot),~\Phi^{-1}(\cdot)$ are defined in \cref{CDFinvnormal}.

\section{Expected normalized variance minimization}\label{app:varmin}
\subsection{\cref{thm:exp_var_min}}
We prove \cref{thm:exp_var_min} in two steps in 
\cref{prop:twolevels} and \cref{prop:alq_onelevel}.

Let $R$ denote a random variable with probability density function (PDF) 
$\pdf$ and cumulative distribution function (CDF) $\cdf$. To show that problem \cref{exp_var_min_cond} is nonconvex, we first 
focus on the problem of optimizing two levels $\min_{a,b} Q(a,b)$ where
\begin{align}
    Q(a,b)=\int_0^a(a-r)r\ud \cdf(r)+\int_a^b(b-r)(r-a)\ud 
    \cdf(r)+\int_b^1(1-r)(r-b)\ud \cdf(r)
\end{align}
in the range $0\leq a\leq b\leq 1$.

\bprop\label{prop:twolevels} The function $Q(a,b)$ is nonconvex in general. It becomes convex if for all $0\leq a\leq b\leq 1$, we have 
\begin{align}\label[ineq]{quadcond}
b(1-a)\pdf(a)\pdf(b)>\left(\cdf(b)-\cdf(a)\right)^2.
\end{align}
\eprop
\bpr 
Using Leibniz's rule \citep{Calculus}, we have 
\begin{align}
\nabla^2Q=\begin{bmatrix}b\pdf(a)&\cdf(a)-\cdf(b)\\ \cdf(a)-\cdf(b)&(1-a)\pdf(b)
  \end{bmatrix}.
\end{align}
We can find the eigenvalues of $\nabla^2Q$ by solving 
$|\nabla^2Q-\lambda\Ibf|=0$, which leads to the following quadratic equation:
\begin{align}
\lambda^2-(b\pdf(a)+(1-a)\pdf(b))\lambda+\left(b(1-a)\pdf(a)\pdf(b)-(\cdf(b)-\cdf(a))^2\right)=0.
\end{align}

We note that \cref{quadcond} is sufficient to guarantee $\nabla^2Q\gb \mathbf{0}$. 
\epr

\bcr 
The sufficient condition \cref{quadcond} is satisfied if $R$ is uniformly distributed in the range $[0,1]$. 
\ecr

We now solve the problem of optimizing a single level, \ie  $\min_b Q(b)$ 
where
\begin{align}
    Q(b)= \int_a^b(b-r)(r-a) \ud \cdf(r)+\int_b^c(c-r)(r-b) \ud \cdf(r).
\end{align}

\bprop\label{prop:alq_onelevel}
The optimal solution to minimize $Q(b)$ is given by\footnote{If $\cdf$ is not 
a one-to-one function, $b^*$ can be any solution that satisfies
$(c-a)\int_b^c \ud \cdf(r)=\int_a^c (r-a)\ud \cdf(r)$.}
\begin{align}
b^*=\cdf^{-1}\left(\cdf(c)-\int_a^c\frac{r-a}{c-a}\ud \cdf(r)\right).
\end{align}
\eprop
\bpr
Using Leibniz's rule, we have 
\begin{align}
\frac{\ud Q}{\ud b}&= \int_a^b(r-a)\ud \cdf(r)-\int_b^c(c-r)\ud \cdf(r),\nn\\
\frac{\ud^2 Q}{\ud b^2}&=(c-a)\pdf(b)\nn. 
\end{align}
We note that $Q$ is convex so we can find the closed-form optimal solution through satisfying the first order optimality condition. 
\epr

\bcr\label{cr:onelevel}
In the special case with $a=0$ and $c=1$, the optimal solution to minimize $Q(b)$ is given by
\begin{equation}\nn
b^*=\cdf^{-1}(1-\E[R]).
\end{equation} \ecr

For the special case of a truncated normal, the inner integral is evaluated as
\begin{align}
    \nn
    \int_a^c\frac{r-a}{c-a}\ud \cdf_{\Tc}(r)
    &=
    \frac{\mu-a}{c-a}
    (\cdf_{\Tc}(c) - \cdf_{\Tc}(a))
    -\frac{\sigma^2}{c-a} (\pdf_{\Tc}(c) - \pdf_{\Tc}(a)),
\end{align}
where $\cdf_{\Tc}$ and $\pdf_{\Tc}$, the CDF and PDF of the truncated normal, 
are defined in \cref{app:cdf}.

\subsection{Projected Gradient Descent}
For the special case of a normal or truncated normal distribution, the gradient 
of the expected normalized variance used in \cref{body:pgd} is:
\begin{align}\label{auto_tune_gd_normal}
\begin{split}
    \dpsidl
    &=(\mu-\ql_{j-1}(t))
    \left(\cdf(\ql_j(t)) - \cdf(\ql_{j-1}(t))\right)+\sigma^2\left(\pdf(\ql_{j-1}(t)) - \pdf(\ql_{j+1}(t))\right)\\
    &\quad+(\mu-\ql_{j+1}(t))
    \left(\cdf(\ql_{j+1}(t)) - \cdf(\ql_{j}(t))\right).
\end{split}    
\end{align}

\subsection{Symmetric Levels}\label{app:symmetry}
In this section, we introduce quantization method with symmetrical levels. This 
is particularly useful when  the estimated PDF of normalized coordinates is an 
even function, which is the case for normal distribution with zero 
mean. Let $(-\ql_{s+1},\ldots,-\ql_1,\ql_1,\ldots,\ql_{s+1})$ denote a sequence 
of symmetrical quantization levels w.r.t. $0$ where 
$0<\ql_1<\cdots<\ql_s<\ql_{s+1}=1$. Let rewrite the vector of adaptable quantization 
levels as $\tilde\bql= [\tilde \ql_{1},\ldots,\tilde \ql_{2s+2}]^\top$ where $\tilde \ql_{1}=-1<\tilde 
\ql_{2}<\cdots<\tilde \ql_{2s+1}<\tilde \ql_{2s+2}=1$. For $\rsym\in[-1,1]$, let 
$\levelt{\rsym}$ and $\qcoeff(\rsym)$ satisfy $\tilde \ql_{\levelt{\rsym}}\leq \rsym\leq 
\tilde \ql_{\levelt{\rsym}+1}$ and $\rsym=\big(1-\qcoeff(\rsym)\big)\tilde 
\ql_{\levelt{\rsym}}+\qcoeff(\rsym)\tilde \ql_{\levelt{\rsym}+1}$, respectively. Let 
$\vbf\in\reals^d$ and $\rsym_i=v_i/\|\vbf\|$ for $i=1,\ldots,d$. 
\begin{definition}The symmetrical quantization of a vector $\vbf\in\reals^d$ is 
\begin{align}
Q_{\tilde\bql}(\vbf)\defeq [q_{\tilde\bql}(v_1),\ldots,q_{\tilde\bql}(v_d)]^\top,
\end{align} 
where $q_{\tilde\bql}(v_i)=\|\vbf\|\cdot h(\rsym_i)$ and $h(\rsym_i)$'s are independent random variables such that
$h(\rsym_i)= \tilde \ql_{\levelt{\rsym_i}}$ with probability $1-\qcoeff(\rsym_i)$ and 
$h(\rsym_i)= \tilde \ql_{\levelt{\rsym_i}+1}$ otherwise where $\qcoeff(\rsym)=(\rsym-\tilde \ql_{\levelt{\rsym}})/(\tilde \ql_{\levelt{\rsym}+1}-\tilde \ql_{\levelt{\rsym}})$.
\end{definition}
Let $r_i=|\rsym_i|$  for $i=1,\ldots,d$. We have the following propositions. 
\bprop\label{prop:Vsym} The variance of quantization with symmetric levels is given by
\begin{align}
\E[\|Q_{\tilde\bql}(\vbf)-\vbf\|^2]=\|\vbf\|^2 \sum_{i=1}^d\sigma^2(r_i),
\end{align} 
 where 
\begin{align}
    \sigma^2(r_i) &= \sum_{r_i\in[0,\ql_1]}
    (\ql_1^2 - r_i^2)
    + \sum_{j=1}^s \sum_{r_i\in[\ql_j,\ql_{j+1}]}
    (r_i - \ql_j)(\ql_{j+1} - r_i).
\end{align}

\eprop
\bpr
Note that for symmetrical levels, we have 
\begin{align}
\sum_{|\rsym_i|\in[\ql_j,\ql_{j+1}]}(|\rsym_i|-\ql_j)(\ql_{j+1}-|\rsym_i|)&=\sum_{\rsym_i\in[-\ql_{j+1},-\ql_j]}(\rsym_i+\ql_{j+1})(-\ql_j-\rsym_i)\nn\\
&\quad+\sum_{\rsym_i\in[\ql_j,\ql_{j+1}]}(\rsym_i-\ql_j)(\ql_{j+1}-\rsym_i).\nn
\end{align}
In addition, we have 
\begin{align}\nn
\sum_{|\rsym_i|\in[0,\ql_1]}(\ql_1^2-|\rsym_i|^2)=\sum_{\rsym_i\in[-\ql_1,\ql_1]}(\ql_1^2-\rsym_i^2).
\end{align}  
\epr

\bprop\label{prop:EVsym}
If PDF of normalized gradients is an even function,\ie $\pdf(-\rsym)=\pdf(\rsym)$ for 
$\rsym\in[-1,1]$, the expected normalized variance in \cref{exp_var_min_cond} 
can be rewritten as
\begin{align}
    \Psi(\bql)
    &=
    2\left(\int_{0}^{\ql_1}(\ql_1^2-r^2)\ud 
    \cdf(r)+\sum_{j=1}^s\int_{\ql_j}^{\ql_{j+1}}(\ql_{j+1}-r)(r-\ql_j)\ud 
    \cdf(r)\right).
\end{align}
\eprop

\subsubsection{GD}
For the case of symmetrical levels, the gradient of the expected normalized variance is given by 
\begin{align}
\begin{split}
    \frac{1}{2}\dxdy{\Psi(\bql(t))}{\ql_{1}}&=2\ql_{1}(t)\big(\cdf(\ql_{1}(t))-\cdf(0)\big)-\int_{\ql_1(t)}^{\ql_{2}(t)}(\ql_{2}(t)-r)\ud \cdf(r),\\
    \frac{1}{2}\dxdy{\Psi(\bql(t))}{\ql_{j}} 
    &=\int_{\ql_{j-1}(t)}^{\ql_j(t)}(r-\ql_{j-1}(t))\ud \cdf(r)
    -\int_{\ql_j(t)}^{\ql_{j+1}(t)}(\ql_{j+1}(t)-r)\ud \cdf(r)
\end{split}
\end{align} for $j=2,\ldots,s$.

For the special case of normal or truncated normal distribution, $\frac{1}{2}\dxdy{\Psi(\bql(t))}{\ql_{j}}$ is obtained by the R.H.S. of \cref{auto_tune_gd_normal} for $j=2,\ldots,s$. In addition, we have: 

\begin{align}
    \frac{1}{2}\dxdy{\Psi(\bql(t))}{\ql_{1}}
    =2\ql_{1}(t)\big(\cdf(\ql_{1}(t))-\cdf(0)\big)+(\mu-\ql_{2}(t))
    \left(\cdf(\ql_{2}(t)) - \cdf(\ql_{1}(t))\right)-\sigma^2\pdf(\ql_{2}(t)).\nn
\end{align}

\subsubsection{CD}
In the following lemma, we solve the problem of optimizing a single level, \ie  
$\min_b \tilde Q(b)$ where
\begin{align}\nn
    \tilde Q(b)= \int_0^b(b^2-r^2) \ud \cdf(r)+\int_b^c(c-r)(r-b) \ud \cdf(r). 
\end{align} 

\bprop\label{prop:amq_onelevel}
The optimal solution to minimize $\tilde Q$ satisfies
\begin{align}\label{tildeQeq}
    2b^*(\cdf(b^*)-\cdf(0))=\int_{b^*}^c(c-r)\ud \cdf(r),
\end{align} where $\cdf$ is the CDF of the normalized coordinate.  
\bpr

Using Leibniz's rule \citep{Calculus}, we have 
\begin{align}\nn
\frac{\ud \tilde Q}{\ud b}&= \int_0^b 2b\ud \cdf(r)-\int_b^c(c-r)\ud \cdf(r),\nn\\
\frac{\ud^2 \tilde Q}{\ud b^2}&=b\pdf(b)+c\pdf(b)+2(\cdf(b)-\cdf(0))>0\nn. 
\end{align}
We note that $\tilde Q$ is convex so we can find the unique optimal solution by satisfying the first order optimality condition. 
\epr 
\eprop

We can solve \cref{tildeQeq} efficiently through a bisection search on 
$[0,\ql_2(t)]$. In particular, for the special case of normal and truncated normal, 
starting with a random $\bql(0)$, the update rule at iteration $t+1$ is the 
same as \cref{update_coor_gen} for $j=2,\ldots,s$. For $\ql_1(t+1)$, we solve
\begin{align}\label{eq:onesol_sym}
    &(\ql_2(t)-\mu)\left(\cdf(\ql_2(t))-\cdf(\ql_1(t+1))\right)+2\ql_1(t+1)\left(\cdf(0)-\cdf(\ql_1(t+1))\right)\nn\\
&+\sigma^2\left(\pdf(\ql_2(t))-\pdf(\ql_1(t+1))\right)=0.\nn
\end{align}

\subsubsection{Exponentially spaced levels}

In this section, we focus on $\bql=[-1,-p,\ldots,-p^s,p^s,\ldots,p,1]^\top$, 
\ie exponentially spaced levels with symmetry. Following 
\cref{prop:EVsym}, the expected normalized variance is given by

\begin{align}
    \Psi(p)
    &=2\left(\int_{0}^{p^s}(p^{2s}-r^2)\ud 
    \cdf(r)+\sum_{j=0}^{s-1}\int_{p^{j+1}}^{p^j}(p^j-r)(r-p^{j+1})\ud 
    \cdf(r)\right).
\end{align}

Using Leibniz's rule, we can compute the first 
order derivative:
\begin{align}\nn
    \frac{1}{2}\frac{\ud \Psi(p)}{\ud p}
    =\int_{0}^{p^s}2sp^{2s-1}\ud 
    \cdf(r)+\sum_{j=0}^{s-1}\int_{p^{j+1}}^{p^j}\left((jp^{j-1}+(j+1)p^j)r-(2j+1)p^{2j}\right)\ud 
    \cdf(r).
\end{align}

In particular, in the special case of a normal or truncated normal 
distribution, we have
\begin{align*}
    \frac{1}{2}\frac{\ud \Psi(p)}{\ud p}
    &=2sp^{2s-1}\left(\cdf(p^s)-\cdf(0)\right)+\sigma^2\sum_{j=0}^{s-1}(jp^{j-1}+(j+1)p^j)
    \left(\pdf(p^{j+1})-\pdf(p^{j})\right)\\
    &\quad+\sum_{j=0}^{s-1}\left(\mu(jp^{j-1}+(j+1)p^j)-(2j+1)p^{2j}\right)
    \left(\cdf(p^j)-\cdf(p^{j+1})\right).    
\end{align*}

We can update $p$ efficiently by a gradient descent algorithm as we have 
a closed-form expression to find the gradient function.

\section{Expected variance minimization in \cref{body:norm}}\label{app:norm}

In the following, we provide the update rules and the analysis of computation complexity of ALQ, GD, and AMQ.  

\subsection{ALQ (CD update)} 
Starting with a random $\bql(0)$, for $t=0,1,\ldots$ and $j=1,\ldots,s$, we solve
\begin{align}
\overline\cdf(\ql_j(t+1))=\overline\cdf(\ql_{j+1}(t))-\int_{\ql_{j-1}(t)}^{\ql_{j+1}(t)}\frac{r-\ql_{j-1}(t)}{\ql_{j+1}(t)-\ql_{j-1}(t)}\ud \overline\cdf(r)
\end{align} by a bisection search on $[\ql_{j-1}(t),\ql_{j+1}(t)]$.

In the special case of (truncated) normal distribution, to obtain $\ql_j(t+1)$ for $j=1,\ldots,s$,  we solve
\begin{align}\label{tnALQNB}
\begin{split}
    &\sum_{n=1}^N\gamma_n
        \left((\ql_{j-1}(t)-\mu_n)
            \left(\cdf_n(\ql_{j+1}(t)) -\cdf_n(\ql_{j-1}(t))\right)
            +\sigma_n^2\left(\pdf_n(\ql_{j+1}(t)) - \pdf_n(\ql_{j-1}(t))\right)
        \right)\\
    &+\left(\ql_{j+1}(t)-\ql_{j-1}(t)\right)
        \left(\overline \cdf(\ql_{j+1}(t))
            - \overline \cdf(\ql_{j}(t+1))\right)=0
\end{split}        
\end{align}
by a bisection search on $[\ql_{j-1}(t),\ql_{j+1}(t)]$.

In the special case of symmetrical levels and (truncated) normal distribution, the update rule is the same as \cref{tnALQNB} for $j=2,\ldots,s$.  For $\ql_1(t+1)$, we solve
\begin{align}
\begin{split}
    &\sum_{n=1}^N\gamma_n
    \left((\ql_2(t)-\mu_n)
        \left(\cdf_n(\ql_{2}(t)) - \cdf_n(\ql_{1}(t+1))\right)
        +\sigma_n^2 \left(\pdf_n(\ql_{2}(t)) -\pdf_n(\ql_{1}(t+1))\right)
    \right)\\
    &+2\ql_1(t+1)\left(\overline \cdf(0)-\overline \cdf(\ql_{1}(t+1))\right)=0
\end{split}    
\end{align}
by a bisection search on $[0,\ql_2(t)]$. %

\subsection{GD update}
The GD algorithm to minimize \cref{expvarapprox} is based on the following 
update rule by starting from a random $\bql(0)$:
\begin{align}\label{auto_tune_gd_gen_NB}
\begin{split}
    \ql_j(t+1)&= \ql_j(t)-\sign(\hat g(t,j))\min\{\eta(t)|\hat g(t,j)|,\delta_j(t)/2\}\\
    \hat g(t,j)&=\int_{\ql_{j-1}(t)}^{\ql_j(t)}(r-\ql_{j-1}(t))\ud \overline\cdf(r)
    -\int_{\ql_j(t)}^{\ql_{j+1}(t)}(\ql_{j+1}(t)-r)\ud \overline\cdf(r)
\end{split}
\end{align} for $t=0,1,\ldots$ and $j=1,\ldots,s$.

In the special case of (truncated) normal distribution, we have
\begin{align}
\begin{split}
&\hat g(t,j)
    =\sum_{n=1}^N\gamma_n
    \Big((\mu_n-\ql_{j-1}(t))
        \left(\cdf_n(\ql_j(t)) - \cdf_n(\ql_{j-1}(t))\right)\\
        &+(\mu_n-\ql_{j+1}(t))
            \left(\cdf_n(\ql_{j+1}(t)) - \cdf_n(\ql_{j}(t))\right)
    +\sigma_n^2\left(\pdf_n(\ql_{j-1}(t)) -\pdf_n(\ql_{j+1}(t))\right)\Big).
\end{split}    
\end{align}

\subsection{AMQ (GD update with exponentially spaced levels)}
In this section, we focus on $\bql=[-1,-p,\ldots,-p^s,p^s,\ldots,p,1]^\top$, \ie 
exponentially spaced levels with symmetry. Following  
\cref{prop:EVsym}, the expected variance of quantization is given by
\begin{align}
    \tilde V(p)=2\left(\int_{0}^{p^s}(p^{2s}-r^2)\ud \overline 
    \cdf(r)+\sum_{j=0}^{s-1}\int_{p^{j+1}}^{p^j}(p^j-r)(r-p^{j+1})\ud \overline 
    \cdf(r)\right).
\end{align}

In the special case of (truncated) normal distribution, we have
\begin{align}
    \frac{\ud\tilde V(p)}{\ud p}
    &=+2sp^{2s-1} \left(\overline \cdf(p^s)-\overline \cdf(0)\right)\nn\\
    &+\sum_{n=1}^N\gamma_n\Big(\sum_{j=0}^{s-1}
        \left(\mu_n(jp^{j-1}+(j+1)p^j)-(2j+1)p^{2j}\right)
            \left(\cdf_n(p^j)-\cdf_n(p^{j+1})\right)\nn\\
        &\qquad +\sigma_n^2\sum_{j=0}^{s-1}(jp^{j-1}+(j+1)p^j)
            \left(\pdf_n(p^{j+1})-\pdf_n(p^{j})\right)\Big).\nn
\end{align}

We can update $p$ efficiently by a gradient descent algorithm as we have 
a closed-form expression to find the gradient function. 

\subsection{Computational complexity and scalability}\label{app:complexity}
The number of iterations for ALQ method to converge is in the order of 
$O(s\log(1/\epsilon))$ where $\epsilon$ is the suboptimality gap of bisection 
search. The number of iterations for AMQ method to achieve a local minimum 
with gap $\epsilon$ is $O(1/\epsilon)$. The total number of gradient computations for GD method to achieve a local minimum 
with gap $\epsilon$ is $O(s/\epsilon)$. Note that processors can run our 
methods in parallel. The time complexity of these methods is independent of the 
number of samples, the number of processors, and the number of parameters. The 
extra computational overhead is negligible compared to costs of computation of 
stochastic gradients and communication. Furthermore, we do not need to optimize 
levels at each iteration. Our experimental results suggest that it is 
sufficient to optimize levels at the lr\_scheduler iterations.

\section{Encoding}\label{app:coding}

A quantized gradient $Q_{\bql}(\vbf)$ can be uniquely determined by a tuple  
$(\|\vbf\|,\signvec,\hbf)$
where $\|\vbf\|$ is the Euclidean norm of the gradient,
$\signvec\defeq[\sign(v_1),\ldots,\sign(v_d)]^\top$ is the vector of signs of the coordinates $v_i$'s,
and $\hbf\defeq[h(r_1),\ldots,h(r_d)]^\top$ are the discrete values of the normalized coordinates after quantization.

We can describe the $\ENCODE$ function (for \cref{AQSGDalg}) 
in terms of the tuple $(\|\vbf\|,\signvec,\hbf)$ and
encoding/decoding scheme $\Gamma : \{\ql_0,\ql_1,\ldots,\ql_{s+1}\} \to \{0,1\}^*$ and $\Gamma^{-1} : \{0,1\}^* \to \{\ql_0,\ql_1,\ldots,\ql_{s+1}\}$. Any lossless prefix code can be used for encoding/decoding. In particular, we consider Huffman coding due to its efficient encoding/decoding and its optimality in terms of achieving the minimum expected code length among methods encoding symbols separately \citep{InfTheory}.

The encoding, $\ENCODE(\vbf)$, of a stochastic gradient is as follows:
We first encode the norm $\|\vbf\|$ using $b$ bits where, in practice, we use standard 32-bit floating point encoding. We then proceed in rounds, $t=0,1,\ldots,d$. Noting that we do not need to encode the sign bit for zero entries of $\hbf$, on round $t$, if $h(r_t)=0$, we transmit $\Gamma(0)$. If $h(r_t)\neq 0$, we transmit $\Gamma( h_{r_t})$, transmit one bit encoding the $\sign(v_t)$, and proceed to the next entry of $\hbf$.

The DECODE function (for \cref{AQSGDalg}) simply reads $b$ bits to reconstruct $\|\vbf\|$. 
Using $\Gamma^{-1}$, it decodes the index of the first coordinate, depending on whether the decoded entry is zero or nonzero, it may read one bit indicating the sign, and then proceeds to decode the next symbol. The process proceeds in rounds, mimicking the encoding process, finishing when all coordinates have been decoded. Note that we can improve coding efficiency by encoding blocks of symbols at the cost of increasing encoding/decoding complexity. In this paper, we focus on a simple lossless prefix coding scheme that encodes symbols separately.

In order to implement an efficient lossless prefix code, we need to know the probabilities associated with our symbols to be coded, \ie $\{\ql_0,\ql_1,\ldots,\ql_{s+1}\}$. Fortunately, we can compute those probabilities using the marginal PDF of normalized coordinates and quantization levels as shown in this proposition: 

\bprop\label{prop:codePMF} The probability of occurrence of $\ql_j$ (weight of symbol $\ql_j$) is given by 
\begin{align}\nn%
\Pr(\ql_j)=\int_{\ql_j-1}^{\ql_j}\frac{r-\ql_{j-1}}{\ql_{j}-\ql_{j-1}}\ud \cdf(r)+\int_{\ql_j}^{\ql_{j+1}}\frac{\ql_{j+1}-r}{\ql_{j+1}-\ql_{j}}\ud \cdf(r)
\end{align} for $j=1,\ldots,s$ where $F$ is the marginal CDF of normalized coordinates. In addition, we have 
\begin{align}
\Pr(\ql_0=0)=\int_{0}^{\ql_1}\frac{1-r}{\ql_1}\ud \cdf(r)~\and~\Pr(\ql_{s+1}=1)=\int_{\ql_s}^{1}\frac{r-\ql_{s}}{1-\ql_{s}}\ud \cdf(r).\nn
\end{align}
\eprop

In the special case of truncated normal distribution, we have the symbol probabilities in closed-form:  

\bcr Suppose normalized coordinates have truncated normal distribution with PDF $\pdf_{\Tc}$ and CDF $\cdf_{\Tc}$ defined in \cref{sec:Tnormal}. The probability of occurrence of $\ql_j$ (weight of symbol $\ql_j$) is given by 
\begin{align}
\Pr(\ql_j)&=\frac{\sigma^2(\pdf_{\Tc}(\ql_{j-1})-\pdf_{\Tc}(\ql_j))+(\mu-\ql_{j-1})(\cdf_{\Tc}(\ql_j)-\cdf_{\Tc}(\ql_{j-1}))}{\ql_j-\ql_{j-1}}\nn\\
&\quad+\frac{\sigma^2(\pdf_{\Tc}(\ql_{j+1})-\pdf_{\Tc}(\ql_j))+(\ql_{j+1}-\mu)(\cdf_{\Tc}(\ql_{j+1})-\cdf_{\Tc}(\ql_{j}))}{\ql_{j+1}-\ql_{j}}\nn
\end{align} for $j=1,\ldots,s$. In addition, we have
\begin{align}
\Pr(\ql_0=0)&=\frac{\sigma^2(\pdf_{\Tc}(\ql_{1})-\pdf_{\Tc}(0))+(\ql_{1}-\mu)(\cdf_{\Tc}(\ql_{1})-\cdf_{\Tc}(\ql_{0}))}{\ql_{1}},\nn\\
\Pr(\ql_{s+1}=1)&=\frac{\sigma^2(\pdf_{\Tc}(\ql_{s})-\pdf_{\Tc}(1))+(\mu-\ql_{s})(\cdf_{\Tc}(1)-\cdf_{\Tc}(\ql_{s}))}{1-\ql_{s}}.\nn
\end{align} 
\ecr

Note that each processor can construct the Huffman tree by knowing $\bql$ and estimating $\mu$ and $\sigma^2$. A Huffman tree of a source with $n$ symbols can be constructed  in time  $O(n)$ if the symbols are sorted by probability. Huffman codes are optimal in terms of expected code-length: 

\bth[{\citealt[Theorems 5.4.1 and 5.8.1]{InfTheory}}]\label{Huffman}
Let $X$ denote a random source. The expected code-length of an optimal prefix code, \eg Huffman code to compress $X$ is bounded by 
$H(X)\leq \E[L]\leq H(X)+1$ where $H(X)$ is the entropy of $X$ in bits. 
\eth

\section{Variance gap}\label{app:gap}
\bprop[Variance gap]\label{prop:vargap}
For any distribution where the gap between the expected variance of a normalized coordinate under an optimal quantization to minimize \cref{exp_var_min_cond} and the worst-case one is lower bounded by some constant, the total gap is lower bounded by $\Omega(d)$. We quantify this gap for the special case of one level with truncated normal density.

\eprop 
\bpr
Suppose we want to design a single level $b\in(0,1)$. As shown in \cref{cr:onelevel}, the optimal $b$ to minimize $Q(b)$ is given by $b^*=F^{-1}(1-\E[R])$. Let $R$ have truncated normal density with parameters $\mu,\sigma^2$ in the unit interval. Plugging PDF and CDF of $R$, the optimal level to minimize \cref{exp_var_min_cond} is given by 
\begin{align}\nn
b^*=\sigma \Phi^{-1}\Big(\Delta (1-\mu)+\sigma\delta+\Phi\Big(-\frac{\mu}{\sigma}\Big)\Big)+\mu,
\end{align} where $\Delta = \Phi((1-\mu)/\sigma)-\Phi(-\mu/\sigma)$, $\delta = \phi((1-\mu)/\sigma)-\phi(-\mu/\sigma)$, $\Phi(x)=\int_{-\infty}^x \exp(-u^2/2)\ud u/\sqrt{2\pi}$, and $\phi(x)=\exp(-x^2/2)/\sqrt{2\pi}$.

Note that $\hat b=1/2$ minimizes the worst-case variance upper bound in \cref{var} \citep{anonymous}. In general, $b^*\neq \hat b$ depending on $\mu$ and $\sigma^2$. Without loss of generality, assume $b^*> \hat b$.

As show in \cref{prop:alq_onelevel},  $Q(b)= \int_0^b(b-r)(r-a) \ud \cdf(r)+\int_b^1(1-r)(r-b) \ud \cdf(r)$ is convex and
\begin{align}\nn
\frac{\ud^2 Q}{\ud b^2}=\frac{\phi((b-\mu)/\sigma)}{\sigma\Delta}.
\end{align}

In the interval $[\hat b,b^*]$, we have 
\begin{align}\nn
\frac{\ud^2 Q}{\ud b^2}\geq \gamma\defeq\min\Big\{\frac{\phi((b^*-\mu)/\sigma)}{\sigma\Delta},\frac{\phi((\hat b-\mu)/\sigma)}{\sigma\Delta}\Big\}.
\end{align} In this interval, $Q$ is $\gamma$-strongly convex, \ie
\begin{align}\nn
Q(\hat b)\geq Q(b^*)+\frac{\gamma}{2}(b^*-\hat b)^2.
\end{align}
Hence, the gap in the expected normalized variance under $b^*$ and $\hat b$ is lower bounded by:
\begin{align}\nn
\frac{1}{2} d\gamma(b^*-\hat b)^2.
\end{align}
\epr

\section{Proof of \cref{thm:varbound} (variance bound)}\label{app:pr_var}
Let $\bql=[\ql_0,\ql_1,\ldots,\ql_s,\ql_{s+1}]^\top$ denote arbitrary quantization levels where $ \ql_0=0<\ql_1 < \dotsm < \ql_{s+1}=1$.
The variance of $Q_{\bql}(\vbf)$, \ie $V_{\bql}(\vbf)=\E[\|Q_{\bql}(\vbf)-\vbf\|_2^2]$, can be expressed as   
\begin{align}\label{var_reform}
V_{\bql}(\vbf)=\|\vbf\|_q^2\big(\sum_{r_i\in\Ic_0}(\ql_1-r_i)r_i+\sum_{j=1}^{s}\sum_{r_i\in\Ic_{j}}(\ql_{j+1}-r_i)(r_i-\ql_j)\big),
\end{align} where $r_i=|v_i|/\|\vbf\|_q$, $\Ic_0\defeq [0,\ql_1]$, and $\Ic_{j}\defeq[\ql_j,\ql_{j+1}]$ for $j=1,\ldots,s$.

We can find the minimum $k_j$ that satisfies $(\ql_{j+1}-r)(r-\ql_{j})\leq k_jr^2$ for $r\in \Ic_{j}$ and $j=1,\ldots,s$. Expressing $r=\ql_j\theta$, we can find $k$ through solving 
\begin{align}
\begin{split}\label{k_for_r2}
k_j &= \max_{1\leq\theta\leq \ql_{j+1}/\ql_j}\frac{(\ql_{j+1}/\ql_j-\theta)(\theta-1)}{\theta^2}\\
&=\frac{\big(\ql_{j+1}/\ql_{j}-1\big)^2}{4(\ql_{j+1}/\ql_{j})}.
\end{split} 
\end{align}

We note that $\ql_{j+1}/\ql_j>1$ and $(x-1)^2/(4x)$ is monotonically increasing function of $x$ for $x>1$.

Furthermore, note that 
\begin{align}\nn
\sum_{r_i\notin\Ic_{0}}r_i^2\leq \frac{\|\vbf\|_2^2}{\|\vbf\|_q^2}. 
\end{align}

Substituting \cref{k_for_r2} into \cref{var_reform}, an upper bound on $V_{\bql}(\vbf)$ is given by 
\begin{align}\nn
V_{\bql}(\vbf)\leq \|\vbf\|_q^2\Big(\frac{(\ql_{j^*+1}/\ql_{j^*}-1)^2}{4(\ql_{j^*+1}/\ql_{j^*})}\frac{\|\vbf\|_2^2}{\|\vbf\|_q^2}+\sum_{r_i\in\Ic_0}(\ql_1-r_i)r_i\Big),
\end{align} where $j^*=\arg\max_{1\leq j\leq s} \ql_{j+1}/\ql_j$.

In our proofs, we use the following known lemma. 
\blm\label{lm:normineq} 
Let $\vbf\in\reals^d$. Then, for all $0<p<q$, we have $\|\vbf\|_q\leq \|\vbf\|_p\leq d^{1/p-1/q}\|\vbf\|_q$.  
\elm 
Note that  \cref{lm:normineq} holds even when $q<1$ and $\|\cdot\|_q$ is merely a seminorm.

In the following, we derive a bound on $\sum_{r_i\in\Ic_0}(2^{-s}-r_i)r_i$, which completes the proof.

\blm\label{lm:K_p}
Let $p\in (0,1)$ and $r\in \Ic_0$. Then we have $r(\ql_1-r)\leq K_p{\ql_1}^{(2-p)} r^p$ where 
\begin{align}\label{K_p}
K_p=\Big(\frac{1/p}{2/p-1}\Big)\Big(\frac{1/p-1}{2/p-1}\Big)^{(1-p)}. 
\end{align}
\elm
\bpr
We can find the minimum $K_p$ through solving $K_p={\ql_1}^{(-2+p)}\max_{r\in \Ic_0}r(\ql_1-r)/r^p$. Expressing the optimization variable as $r=\ql_1\theta^{1/p}$, $K_p$ can be obtained by solving this problem: 
\begin{align}\label[prob]{opt_K_p}
K_p=\max_{0<\theta< 1} \theta^{1/p-1}-\theta^{2/p-1}. 
\end{align}
We can solve \cref{opt_K_p} and obtain the optimal solution $\theta^*=\big(\frac{1/p-1}{2/p-1}\big)^p $. Substituting $\theta^*$ into \cref{opt_K_p}, we obtain \cref{K_p}.  
\epr

Let ${\Sc_j}$ denote the coordinates of vector ${\vbf}$ whose elements fall 
into the ${(j+1)}$-th bin, \ie ${\Sc_j\defeq\{i:r_i\in[l_j,l_{j+1}]\}}$ for 
${j=0,\ldots,s}$. %

Then, for any $0<p<1$ and $q\geq 2$, we have 
\begin{align}
\|\vbf\|_q^{2}\sum_{r_i\in\Ic_0}r_i^p&= \|\vbf\|_q^{2-p}\sum_{i\in\Sc_0}|v_i|^p\nn\\
&\leq \|\vbf\|_q^{2-p}\|\vbf\|_p^p\nn\\
&\leq \|\vbf\|_q^{2-p}\|\vbf\|_2^pd^{1-p/2}\nn\\
&\leq \|\vbf\|_2^2d^{1-p/2}\nn,
\end{align} where the third inequality holds as $\|\vbf\|_p\leq \|\vbf\|_2d^{1/p-1/2}$ using \cref{lm:normineq} and the last inequality holds as $\|\vbf\|_q\leq \|\vbf\|_2$ for $q\geq 2$. 

This gives us an upper bound on $V_{\bql}(\vbf)$:  
\begin{align}\nn
V_{\bql}(\vbf)\leq \|\vbf\|_2^2\Big(\frac{(\ql_{j^*+1}/\ql_{j^*}-1)^2}{4(\ql_{j^*+1}/\ql_{j^*})}+K_p{\ql_1}^{(2-p)}d^{1-p/2}\Big). 
\end{align}

 For $q\geq 1$, we have $\|\vbf\|_q^{2-p}\leq \|\vbf\|_2^{2-p}d^{\frac{2-p}{\min\{q,2\}}-\frac{2-p}{2}}$, which gives 
\begin{align}\nn
V_{\bql}(\vbf)\leq \|\vbf\|_2^2\Big(\frac{(\ql_{j^*+1}/\ql_{j^*}-1)^2}{4(\ql_{j^*+1}/\ql_{j^*})}+K_p{\ql_1}^{(2-p)}d^{\frac{2-p}{\min\{q,2\}}}\Big). 
\end{align}

\section{Proof of \cref{thm:codebound} (code-length bound)}\label{app:pr_code}

Let $|\cdot|$ denote the length of a binary string.
In this section, we obtain an upper bound on $\E[|\ENCODE(\vbf)]$, \ie the expected number of communication bits per iteration. 
Recall from \cref{app:coding} that the 
quantized vectors $Q_{\bql}(\vbf)$ is uniquely determined by the tuple 
$(\|\vbf\|_q,\signvec,\hbf)$.

We first encode the norm $\|\vbf\|_q$ using $b$ bits where, in practice, we use standard 32-bit floating point encoding.

We send one bit for each nonzero entry of $\hbf$. Let ${\Sc_j\defeq\{i:r_i\in[l_j,l_{j+1}]\}}$ and ${d_j\defeq |\Sc_j|}$ for 
${j=0,\ldots,s}$. We have an upper bound on the expected number of nonzero entries as follows:

\blm\label{lm:sparsity}
Let $\vbf\in\reals^d$. The expected number of nonzeros in $Q_{\bql}(\vbf)$ is bounded above by
\begin{align}%
\E[\|Q_{\bql}(\vbf)\|_0]\leq {\ql_1}^{-q}+\frac{d^{1-1/q}}{\ql_1}.\nn 
\end{align}
\elm 
\bpr
Note that $d-d_0\leq {\ql_1}^{-q}$ since 
\begin{align}
(d-d_0){\ql_1}^{q}\leq \sum_{i\not\in \Sc_0}r_i^q\leq 1.
\end{align}
For each $i\in\Sc_0$, $Q_{\bql}(v_i)$ becomes zero with probability $1-r_i/\ql_1$, which results in 
\begin{align}
\begin{split}
\E[\|Q_{\bql}(\vbf)\|_0]&\leq d-d_0+\sum_{i\in\Sc_0}r_i/\ql_1\\
&\leq {\ql_1}^{-q}+\frac{d^{1-1/q}}{\ql_1},
\end{split}
\end{align} where the last inequality holds as $\|\vbf\|_1\leq \|\vbf\|_qd^{1-1/q}$ using \cref{lm:normineq}.
\epr

For each entry of $\hbf$, we send the associated codeword. The optimal expected code-length for transmitting one random symbol is within one bit of the entropy of the source. Hence, we need to transmit upto $d(H(L)+1)$ to transmit entries of $\hbf$ \citep{InfTheory}. Putting everything together, we have 
\begin{align}\nn
\E[|\ENCODE(\vbf)|]\leq b+n_{\ql_1,d}+d(H(L)+1).
\end{align}
Finally, note that the entropy of a source with $n$ outcomes is bounded above by $\log_2(n)$. %

\section{AQSGD for smooth nonconvex optimization}\label{app:nonconvex}
On nonconvex problems, we can establish  convergence guarantees in terms of convergence to a local minima for a smooth loss function along the lines of, e.g., \citep[Theorem 2.1]{Ghadimi}. 

\bth[AQSGD for smooth nonconvex optimization]\label{thm:nonconvbound}
Let $\fun:\Omega\rightarrow \reals$ denote a possibly nonconvex and $\beta$-smooth function. Let $\wbf_0\in\Omega$ denote an initial point, $\overline{\epsilon_Q}$ and $\overline{N_Q}$ be defined as in \cref{thm:convnonsmooth}, $T\in\integers^{>0}$, and $\fun^*=\inf_{\wbf\in\Omega}\fun(\wbf)$.
Suppose that \cref{AQSGDalg} is executed for $T$ iterations with a learning rate $\alpha<2/\beta$ on $M$ processors, 
each with access to independent stochastic gradients of $\fun$ with a second-moment bound $B$,  such that levels are updated $K$ times where $\bql_k$ with variance bound $\epsilon_{Q,k}$ and code-length bound $N_{Q,k}$ is used for $T_k$ iterations.
Then there exists a random stopping time $R\in\{0,\ldots,T\}$ such that AQSGD guarantees
\begin{align}\nn
\E[\|\nabla \fun(\wbf_R)\|^2]\leq \frac{\beta(\fun(\wbf_0)-\fun^*)}{T}+\frac{2(1+\overline{\epsilon_Q})B}{M}.
\end{align}

In addition, AQSGD requires at most $\overline{N_Q}$ communication bits per iteration in expectation.
\eth 

\section{AQSGD with momentum}\label{app:momentum}

The update rule for full-precision unified momentum SGD (UMSGD) is given by \citep{Unified}  
\begin{align}
\begin{split}\label{sgdm}
\ybf_{t+1}&=\wbf_t-\alpha g(\wbf_t)\\
\ybf^\ql_{t+1}&=\wbf_t-l\alpha g(\wbf_t)\\
\wbf_{t+1}&=\ybf_{t+1}+\mu\big(\ybf^\ql_{t+1}-\ybf^\ql_t\big),
\end{split}
\end{align} 
where $\wbf_t$ is the current parameter input and $\mu\in[0,1)$ is the momentum parameter. Note that the heavy-ball method \citep{Polyak} and Nesterov’s accelerated gradient method \citep{Nesterov} are the special cases of UMSGD obtained by substituting $l=0$ and $l=1$ into \cref{sgdm}, respectively. 

The steps for data-parallel version of UMSGD are those in \cref{AQSGDalg} by replacing \cref{step:agg} with an UMSGD update. We have convergence guarantees for adaptively quantized SGD with momentum (AQSGDM) along the lines of, e.g., \citep[Theorem 1]{Unified}.
We first establish the convergence guarantees for convex optimization in the following theorem. 

\bth[AQSGDM for convex optimization]\label{convQUM}
Let $\fun:\reals^d\rightarrow \reals$ denote a convex function with $\|\nabla \fun(\wbf)\|\leq V$ for all $\wbf$. 
Let $\wbf_0$ denote an initial point, $\wbf^*=\arg\min \fun(\wbf)$, $\hat\wbf_T=1/T\sum_{t=0}^T\wbf_t$, and $\overline{\epsilon_Q}$ and $\overline{N_Q}$ be defined as in \cref{thm:convnonsmooth}.

Suppose that AQSGDM %
is executed for $T$ iterations with  a learning rate $\alpha>0$ on $M$ processors, 
each with access to independent stochastic gradients of $\fun$ with a second-moment bound $B$,  such that levels are updated $K$ times where $\bql_k$ with variance bound $\epsilon_{Q,k}$ and code-length bound $N_{Q,k}$ is used for $T_k$ iterations.
Then AQSGDM satisfies 
\begin{align}\label{convUMbound}
\E[\fun(\hat\wbf_T)]-\min_{\wbf\in\Omega}\fun(\wbf)\leq \epsilon^\ql_{\mu},
\end{align} where $\epsilon^\ql_{\mu}=\mu (\fun(\wbf_0)-\fun(\wbf^*))/((1-\mu)(T+1))+(1-\mu)\|\wbf_0-\wbf^*\|^2/(2\alpha(T+1))+\alpha (1+2l\mu)(V^2+(1+\overline{\epsilon_Q}) B/M)/(2(1-\mu))$.

In addition, AQSGD requires at most $\overline{N_Q}$ communication bits per iteration in expectation.
\eth

On nonconvex problems, (weaker) convergence guarantees can be established for AQSGDM.  In particular, AQSGDM is guaranteed to converge to a local minima for smooth general loss functions.

\bth[AQSGDM for smooth nonconvex optimization]\label{nonconvQUM}
Let $\fun:\reals^d\rightarrow \reals$ denote a possibly nonconvex and $\beta$-smooth function with  $\|\nabla \fun(\wbf)\|\leq V$ for all $\wbf$. 
Let $\wbf_0$ denote an initial point, $\wbf^*=\arg\min \fun(\wbf)$, and $\overline{\epsilon_Q}$ and $\overline{N_Q}$ be defined as in \cref{thm:convnonsmooth}.

Suppose that AQSGDM %
is executed for $T$ iterations with $\alpha=\min\{(1-\mu)/(2\beta),C/\sqrt{T+1}\}$ for some $C>0$ on $M$ processors, 
each with access to independent stochastic gradients of $\fun$ with a second-moment bound $B$,  such that levels are updated $K$ times where $\bql_k$ with variance bound $\epsilon_{Q,k}$ and code-length bound $N_{Q,k}$ is used for $T_k$ iterations.
Then AQSGDM satisfies
\begin{align}\nn
\min_{t=0,\ldots,T}\E[\|\nabla \fun(\wbf_t)\|^2]\leq \frac{2(\fun(\wbf_0)-\fun(\wbf^*))(1-\mu)}{\alpha(T+1)}+\frac{C\tilde{V}}{(1-\mu)^3\sqrt{T+1}},
\end{align} where
\begin{align}\nn
\tilde{V}=\beta \big(\mu^2((1-\mu)l-1)^2+(1-\mu)^2\big)(V^2+(1+\overline{\epsilon_Q}) B/M).
\end{align}

In addition, AQSGD requires at most $\overline{N_Q}$ communication bits per iteration in expectation.
\eth

\section{Theoretical guarantees for levels with symmetry}\label{app:theorysym}
We first obtain variance upper bound for the symmetrical $\bql=[-\ql_{s+1},\ldots,-\ql_1,\ql_1,\ldots,\ql_{s+1}]^\top$. 

\bth[Variance bound]\label{thm:varboundSym}
Let $\vbf\in\reals^d$  and $q\geq 1$. The quantization of $\vbf$ under $L^q$ normalization satisfies $\E[Q_{\bql}(\vbf)]=\vbf$. Furthermore, we have
\begin{align}\label{varboundSym}
\E[\|Q_{\bql}(\vbf)-\vbf\|_2^2]\leq\epsilon_Q\|\vbf\|_2^2,
\end{align} where $\epsilon_Q=\ql_1^2d^{\frac{2}{\min\{q,2\}}}+\frac{(\ql_{j^*+1}/\ql_{j^*}-1)^2}{4(\ql_{j^*+1}/\ql_{j^*})}$ where $j^*=\arg\max_{1\leq j\leq s} \ql_{j+1}/\ql_j$.
\eth
\bpr
Following \cref{prop:Vsym}, the variance is given by
\begin{align}\nn
\E[\|Q_{\bql}(\vbf)-\vbf\|_2^2]=\|\vbf\|_q^2\Big(\sum_{r_i\in[0,\ql_1]}(\ql_1^2-r_i^2)+\sum_{j=1}^s\sum_{r_i\in[\ql_j,\ql_{j+1}]}(r_i-\ql_j)(\ql_{j+1}-r_i)\Big).
\end{align}
Note that $\ql_1^2-r^2\leq \ql_1^2$ for $r\in[0,\ql_1]$. The rest of the proof follows the proof of \cref{thm:varbound}.
\epr

In order to implement an efficient lossless prefix code, we need to know the probabilities associated with our symbols to be coded, \ie $\{-\ql_{s+1},\ldots,-\ql_1,\ql_1,\ldots,\ql_{s+1}\}$. We can obtain those probabilities using the marginal PDF of normalized coordinates: 

\bprop\label{prop:codePMFSym} Suppose $\pdf(-\rsym)=\pdf(\rsym)$. The probability of occurrence of $\ql_j$ (weight of symbol $\ql_j$) $\Pr(\ql_j)$ is equal to $\Pr(-\ql_j)$, given by 
\begin{align}\nn%
\Pr(\ql_j)=\int_{\ql_j-1}^{\ql_j}\frac{\rsym-\ql_{j-1}}{\ql_{j}-\ql_{j-1}}\ud \cdf(\rsym)+\int_{\ql_j}^{\ql_{j+1}}\frac{\ql_{j+1}-\rsym}{\ql_{j+1}-\ql_{j}}\ud \cdf(\rsym)
\end{align} for $j=2,\ldots,s$. In addition, we have

\begin{align}
\Pr(\ql_1)=\Pr(-\ql_1)&=\int_{-\ql_1}^{\ql_1}\frac{\rsym+\ql_1}{2\ql_1}\ud \cdf(\rsym)+\int_{\ql_1}^{\ql_2}\frac{\ql_2-\rsym}{\ql_2-\ql_1}\ud \cdf(\rsym),\nn\\
\Pr(\ql_{s+1}=1)=\Pr(-\ql_{s+1})&=\int_{\ql_s}^{1}\frac{\rsym-\ql_{s}}{1-\ql_{s}}\ud \cdf(\rsym).\nn
\end{align}
\eprop

In the special case of truncated normal distribution, we have the symbol probabilities in closed-form:  

\bcr Suppose normalized coordinates have truncated normal distribution with PDF $\pdf_{\Tc}$ and CDF $\cdf_{\Tc}$ defined in \cref{sec:Tnormal}. The probability of occurrence of $\ql_j$ (weight of symbol $\ql_j$) is given by 
\begin{align}
\Pr(\ql_j)=\Pr(-\ql_j)&=\frac{\sigma^2(\pdf_{\Tc}(\ql_{j-1})-\pdf_{\Tc}(\ql_j))+(\mu-\ql_{j-1})(\cdf_{\Tc}(\ql_j)-\cdf_{\Tc}(\ql_{j-1}))}{\ql_j-\ql_{j-1}}\nn\\ 
&\quad+\frac{\sigma^2(\pdf_{\Tc}(\ql_{j+1})-\pdf_{\Tc}(\ql_j))+(\ql_{j+1}-\mu)(\cdf_{\Tc}(\ql_{j+1})-\cdf_{\Tc}(\ql_{j}))}{\ql_{j+1}-\ql_{j}}\nn
\end{align} for $j=2,\ldots,s$. In addition, we have 
\begin{align}
\Pr(\ql_1)=\Pr(-\ql_1)&=\frac{\sigma^2(\pdf_{\Tc}(\ql_{2})-\pdf_{\Tc}(\ql_1))+(\ql_{2}-\mu)(\cdf_{\Tc}(\ql_{2})-\cdf_{\Tc}(\ql_{1}))}{\ql_{2}-\ql_{1}}\nn\\
&\quad+\frac{(\mu+\ql_{1})(\cdf_{\Tc}(\ql_1)-\cdf_{\Tc}(-\ql_{1}))}{2\ql_1},\nn\\
\Pr(\ql_{s+1}=1)=\Pr(-\ql_{s+1})&=\frac{\sigma^2(\pdf_{\Tc}(\ql_{s})-\pdf_{\Tc}(1))+(\mu-\ql_{s})(\cdf_{\Tc}(1)-\cdf_{\Tc}(\ql_{s}))}{1-\ql_{s}}.\nn
\end{align}
\ecr

Finally, we have the following bound on the expected number of communication bits per iteration for quantizing with symmetrical levels. 
\bth[Code-length bound]\label{thm:codeboundSym}
Let $\vbf\in\reals^d$ and $q\geq 1$. The expectation $\E[|\ENCODE(\vbf)|]$
 of the number of communication bits needed to transmit $Q_{\bql}(\vbf)$ under $L^q$ normalization is bounded by
\begin{align}\label{codeboundSym}\textstyle
\E[|\ENCODE(\vbf)|] \leq b+d(H(L)+1)\leq b+d(\log_2(2s+2)+1),  
\end{align} where $b$ is a constant and $L$ is a random variable with the probability mass function given by \cref{prop:codePMFSym}. 
\eth
\section{Experimental details and additional experiments}\label{app:exp}
In this section, we provide additional experiments for the methods evaluated in~\cref{sec:exp}.  In addition to baselines discussed in~\cref{sec:exp}, we
present results for \alqnless, which minimizes the expected normalized variance
using coordinate descent in~\cref{exp_var_min_cond}, \alqnbased with norm
adjustments in~\cref{body:norm}, ALQ adapted using gradient descent
in~\cref{body:pgd} (\alqgnbased, \alqgnless), \amqnless, and \amqnbased. This
section includes full ImageNet runs.
\cref{fig:cifar10_resnet110_/est_var,fig:cifar10_resnet110_-g/Vloss,fig:cifar10_resnet32_-g/Vloss,fig:cifar10_resnet32_/est_var}
have also been extended to include all variations of the proposed algorithms and
baselines.

\begin{table}
    \centering
    \caption{Training Hyper-parameters for CIFAR-10 and ImageNet}
    \begin{tabular}{C{3cm}|C{2.5cm}|C{2.5cm}|C{2.5cm}}
\toprule
       Hyperparameter    & ResNet-32 on CIFAR-10 & ResNet-110 on CIFAR-10 & ImageNet        \\
\midrule                                                                 
       Learning Rate     & 0.1                   & 0.1                    & 0.1             \\
       LR Decay Schedule & At 45K \& 60K         & At 45K \& 60K          & At 300K \& 450K \\
       Batch Size        & 128                   & 64                     & 64              \\
       Momentum          & 0.9                   & 0.9                    & 0.9             \\
       Total Iterations  & 80K                   & 80K                    & 600K            \\
       Weight Decay      & $10^{-4}$             & $10^{-4}$              & $10^{-4}$       \\
       Optimizer         & SGD                   & SGD                    & SGD             \\
\bottomrule
    \end{tabular}
    \label{tab:hp}
\end{table}

\begin{table}
    \centering
    \caption{Validation Accuracy on Full ImageNet Run}
    \begin{tabular}{C{3cm}|C{2.5cm}}
\toprule
       Quantization Method          & ResNet-18 on ImageNet       \\
\midrule
        Bucket Size                 & $8192$                      \\
\midrule
         \sgd                       & 64.67\% $\pm$ 0.13          \\
         \supersgd                  & \textbf{69.85\% $\pm$ 0.05} \\
\midrule
         \nuq~\citep{NUQSGD,Samuel} & 35.43\% $\pm$ 0.28          \\
         \qinf~\citep{QSGD}         & 67.48\% $\pm$ 0.08          \\
         \terngrad~\citep{TernGrad} & 63.97\% $\pm$ 0.11          \\
\midrule
         \alqnbased                 & \textbf{68.65\% $\pm$ 0.10} \\
         \alqnless                  & \textbf{68.50\% $\pm$ 0.10} \\
         \amqnbased                 & 67.76\% $\pm$ 0.09          \\
         \amqnless                  & 67.96\% $\pm$ 0.10          \\
\bottomrule
    \end{tabular}
    \label{tab:vacc_supp}
\end{table}

An implementation challenge is that the value of the statistics, especially the
variance, can become very small. This makes PDF and CDF calculations
challenging. The challenge is that the value of PDF is very close to zero when
it is far from the mean but not exactly zero. In order to overcome this
challenge, we use histograms to model the distribution of gradients as a
weighted sum of truncated normals. Another problem is the large number of
statistics that are calculated. As presented in \cref{sec:aqsgd}, we sample a
number of gradients and then normalize the gradients. Then we split the
gradients into buckets and calculate average, variance, and norm of each of the
buckets. The number of means, variances, and norms can become very large with
large networks and small bucket sizes. To reduce computational complexity of the
algorithm, we sample uniformly from these values. This number of samples is
equal to 20 for small networks such as ResNet-8 and networks trained on CIFAR-10;
however, in experiments on ImageNet, we used 350 samples to achieve the desired
accuracy.

One other understudied detail in quantizing is how bucketing is performed. In
~\citep{QSGD,NUQSGD}, gradient coordinates in each bucket do not exceed the
layer size. It means that the gradient coordinates in a bucket do not contain
gradient coordinates from the next layer even if the bucket size is not fully
utilized. This leads to creation of under-sized buckets that can be problematic
for quantization performance. Different tricks are employed to fix this problem.
These tricks include transmitting biases or under-sized buckets in
full-precision (not that typically biases are main sources of under-sized
buckets). In our implementation, we normalize the buckets network-wise and do
not consider the layer size as the bucket size boundary. We only transmit the
last bucket in full precision if it is smaller than the specified bucket size.

\textbf{Update Schedule.} In the ImageNet and CIFAR-10 runs, adaptive level
updates are scheduled at 100 and 2000 iterations only once and every 10K
iterations. The reason for this schedule is changes in the gradient statistics
over the course of training.  As shown in \cref{fig:variance-change}, the
average variance changes rapidly during the first iterations and then only
changes at every learning rate schedule. In practice, we noticed accuracy
degradation especially when the levels are not updated during the initial
iterations where the average variance is rapidly changing.

\begin{figure}
    \centering
    \begin{subfigure}[b]{\twocolfigwidth}
        \includegraphics[width=\textwidth]{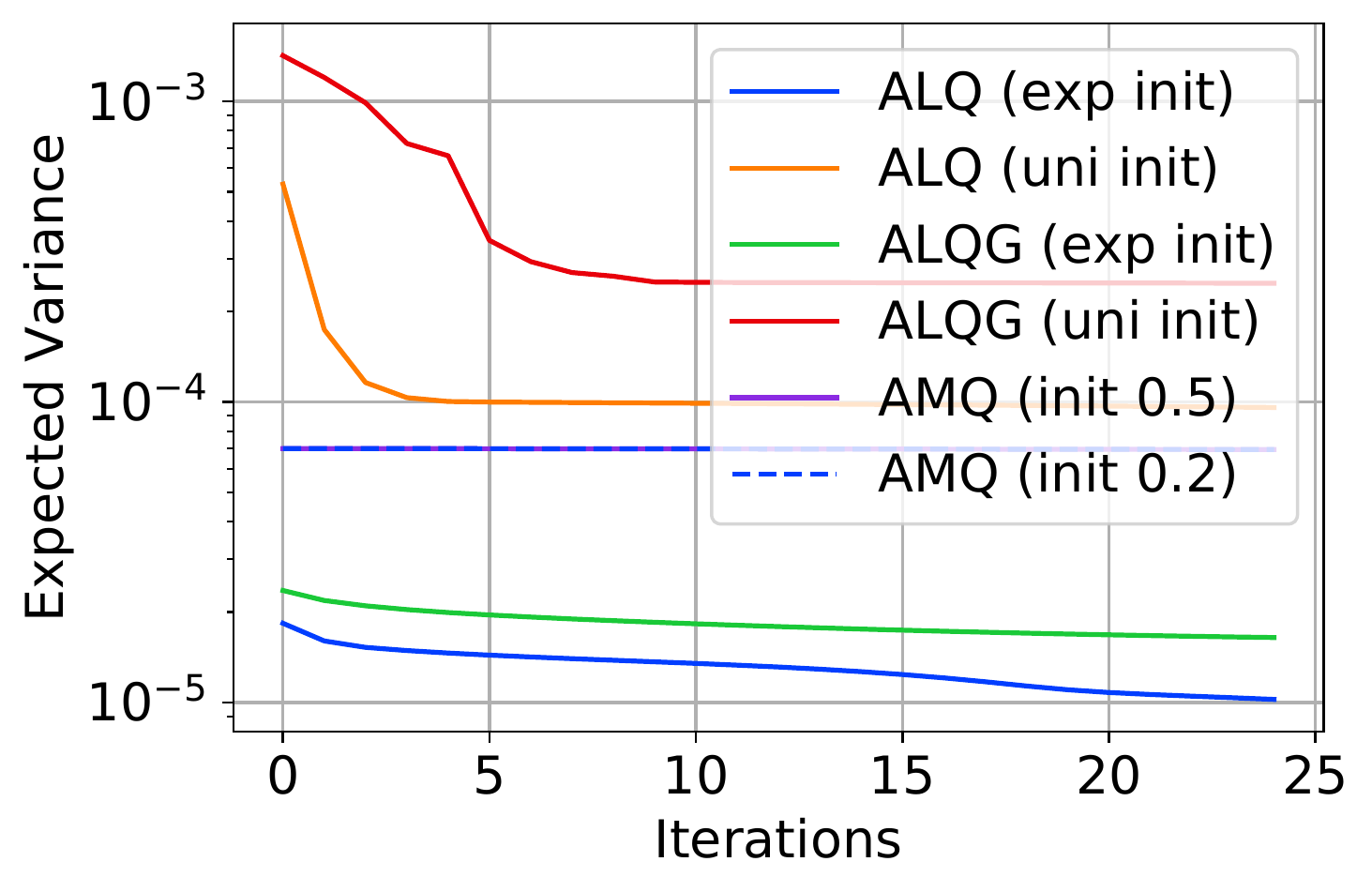}
        \vspace*{\capshift}
        \caption{Expected Normalized Variance}
        \label{fig:nb-level-update}
    \end{subfigure}
    \begin{subfigure}[b]{\twocolfigwidth}
        \includegraphics[width=\textwidth]{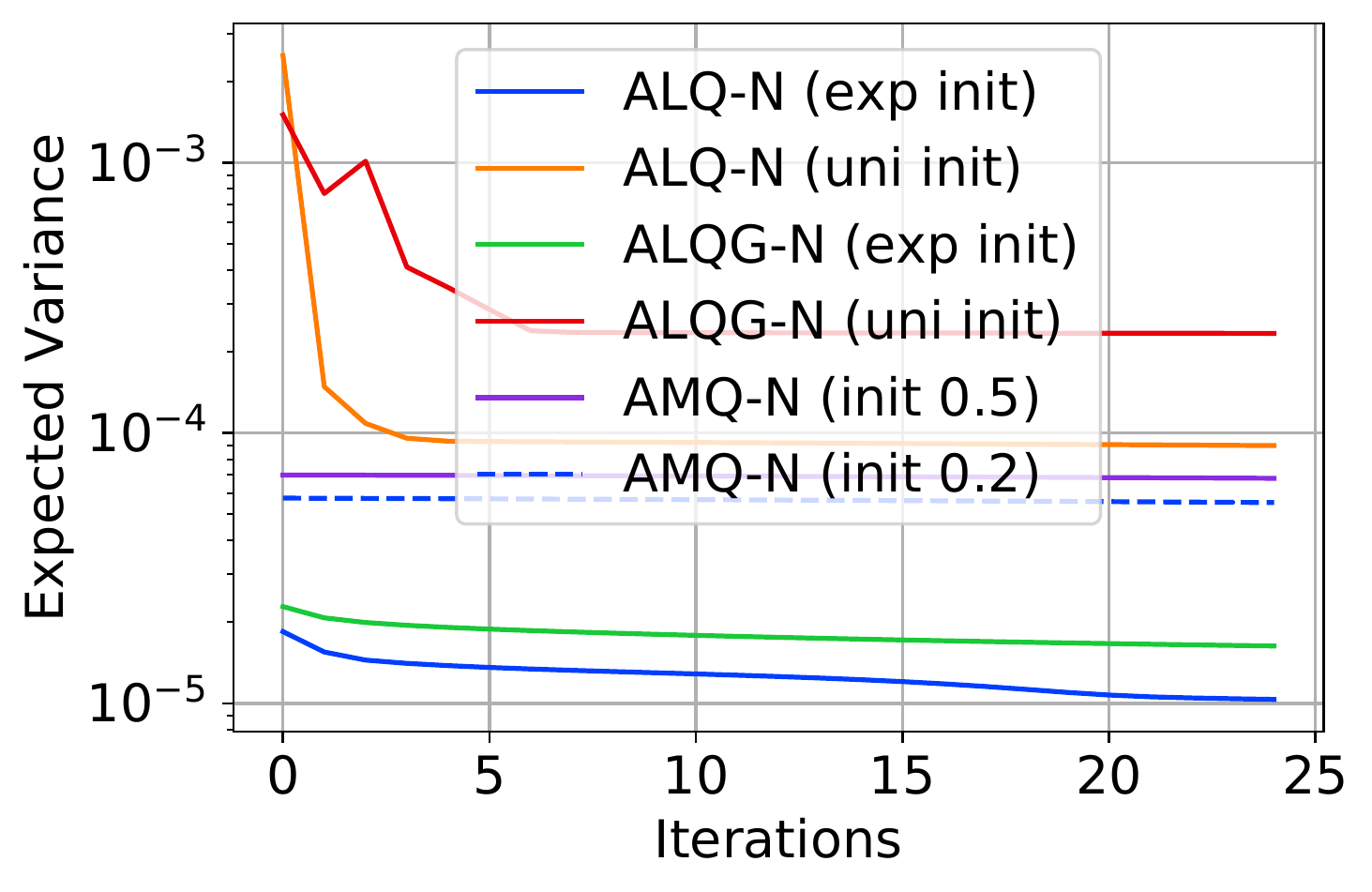}
        \vspace*{\capshift}
        \caption{Expected Variance}
        \label{fig:nl-level-update}
    \end{subfigure}
    \caption{\textbf{Convergence} of different level update methods}
    \label{fig:level-update}
\end{figure}

\textbf{Convergence of level updating.} \cref{fig:level-update} shows the
expected normalized variance (the objective in \cref{exp_var_min_cond}) and
expected variance (the objective in~\cref{exp_var_min_uncond}) during one step
of adapting levels. This figure shows that the objective function
in~\cref{exp_var_min_cond} is nonconvex and different initializations lead to
sub-optimal solutions.  \alqgnbased and \alqgnless refer to variations of ALQ
using gradient descent instead of coordinate descent.

\textbf{Hyperparameters used for training.} \cref{tab:hp} shows the
hyperparameters used for training CIFAR-10 and ImageNet. These are conventional
hyperparameters for training ResNet models. \supersgd is able to replicate the
accuracy reported in~\citep{Resnet} showing the correct setting for training.

\textbf{Validation accuracy on full ImageNet run} \cref{tab:vacc_supp} shows the
validation accuracy of full ImageNet runs on ResNet-18. Total number of
iterations required for a full ImageNet run is 600K. This table shows that
\alqnbased and \alqnless are able to outperform \qinf by 1\% on ImageNet.

Similar performance of normalized and unnormalized variations of \amqnbased and
\alqnbased methods suggests that for given datasets and deep models, the
distribution of normalized gradient coordinates can be represented by either of
the forumulations in~\cref{sec:aqsgd}. In \amqnless and \alqnless, $\mu$ and
$\sigma$ values for the truncated normal distribution is equal to the average of
$\mu$ and $\sigma$ for individual buckets.

\cref{fig:cifar10_resnet32_-g-e/Vloss,fig:cifar10_resnet110_-g-e/Vloss} show an
interesting observation for \nuq. Although \nuq has worse performance in terms
of the training loss and average variance compared to all other approaches, it
is able to achieve better validation accuracy. This suggests that \nuq is able
to generalize better in this specific setting. However, this pattern does not
repeat when it comes to the ImageNet dataset.

{\def\EXPNAME{-g-e/Tloss}\def\FIGTITLE{Extended Training loss}

\begin{figure}[t]
    \centering
    \begin{subfigure}[b]{\twocolfigwidth}
        \includegraphics[width=\textwidth]{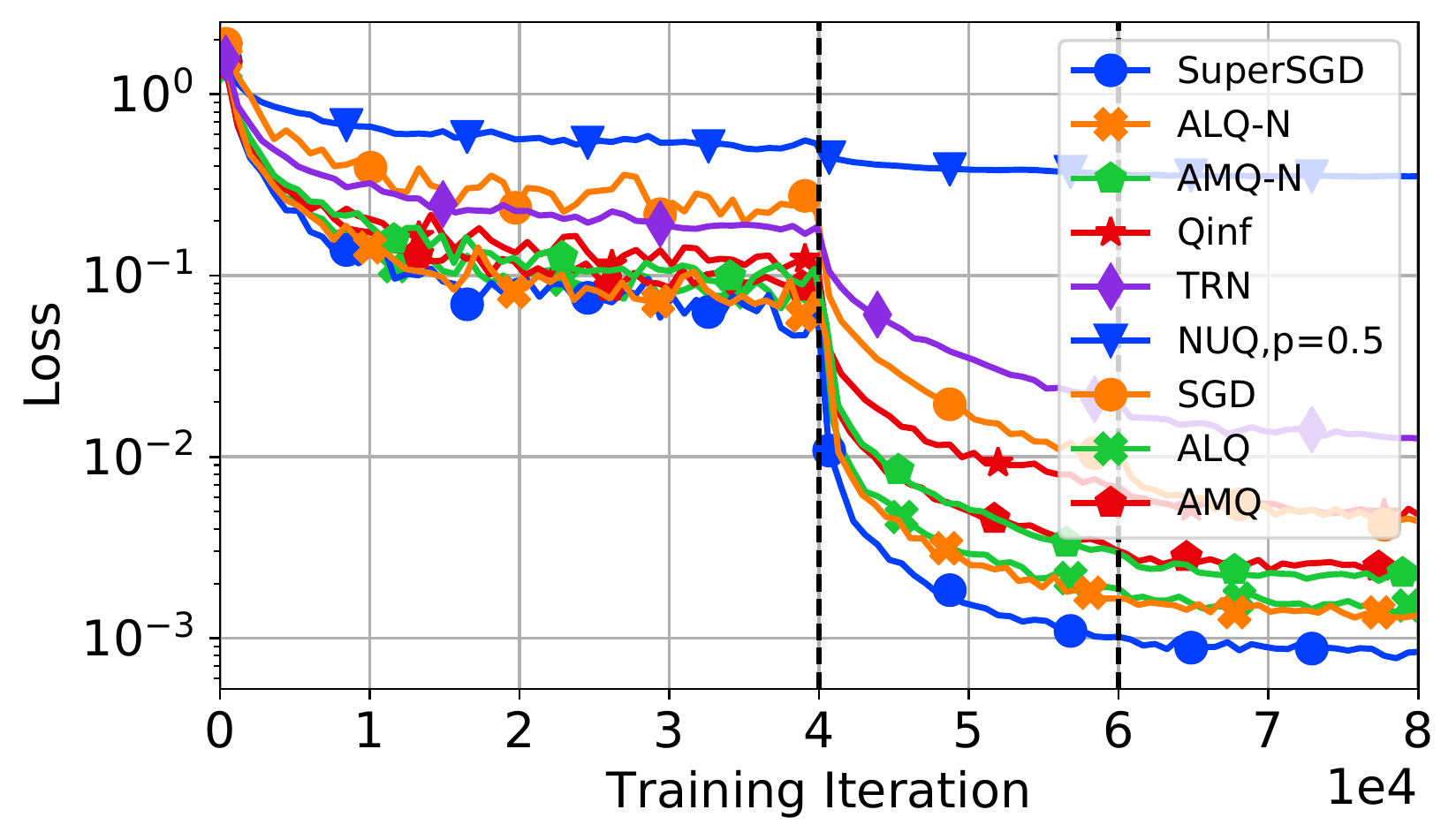}
        \vspace*{\capshift}
        \caption{ResNet-32 on CIFAR-10}
        \label{fig:cifar10_resnet32_\EXPNAME}
    \end{subfigure}
    \begin{subfigure}[b]{\twocolfigwidth}
        \includegraphics[width=\textwidth]{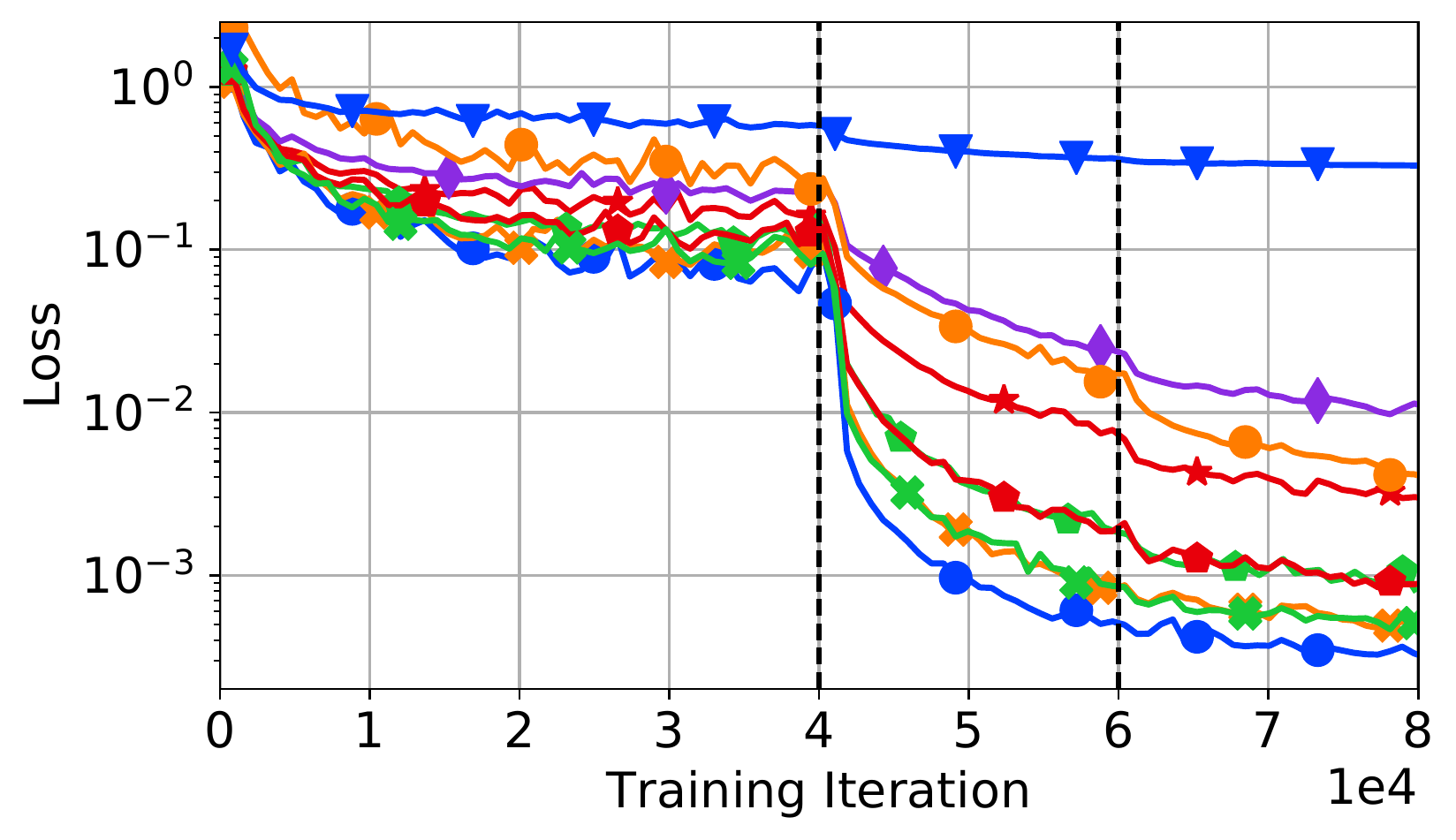}
        \vspace*{\capshift}
        \caption{ResNet-110 on CIFAR-10}
        \label{fig:cifar10_resnet110_\EXPNAME}
    \end{subfigure}
    \caption{\textbf{{\FIGTITLE}} on CIFAR-10. All methods use $3$ 
    quantization bits. Bucket size for ResNet-110 trained on CIFAR-10 is
    $16384$ and for ResNet-32 is $8192$.}
\label{fig:\EXPNAME}
\end{figure}
}
{\def\EXPNAME{-g-e/Vloss}\def\FIGTITLE{Extended Validation loss}

\begin{figure}[t]
    \centering
    \begin{subfigure}[b]{\threecolfigwidth}
        \includegraphics[width=\textwidth]{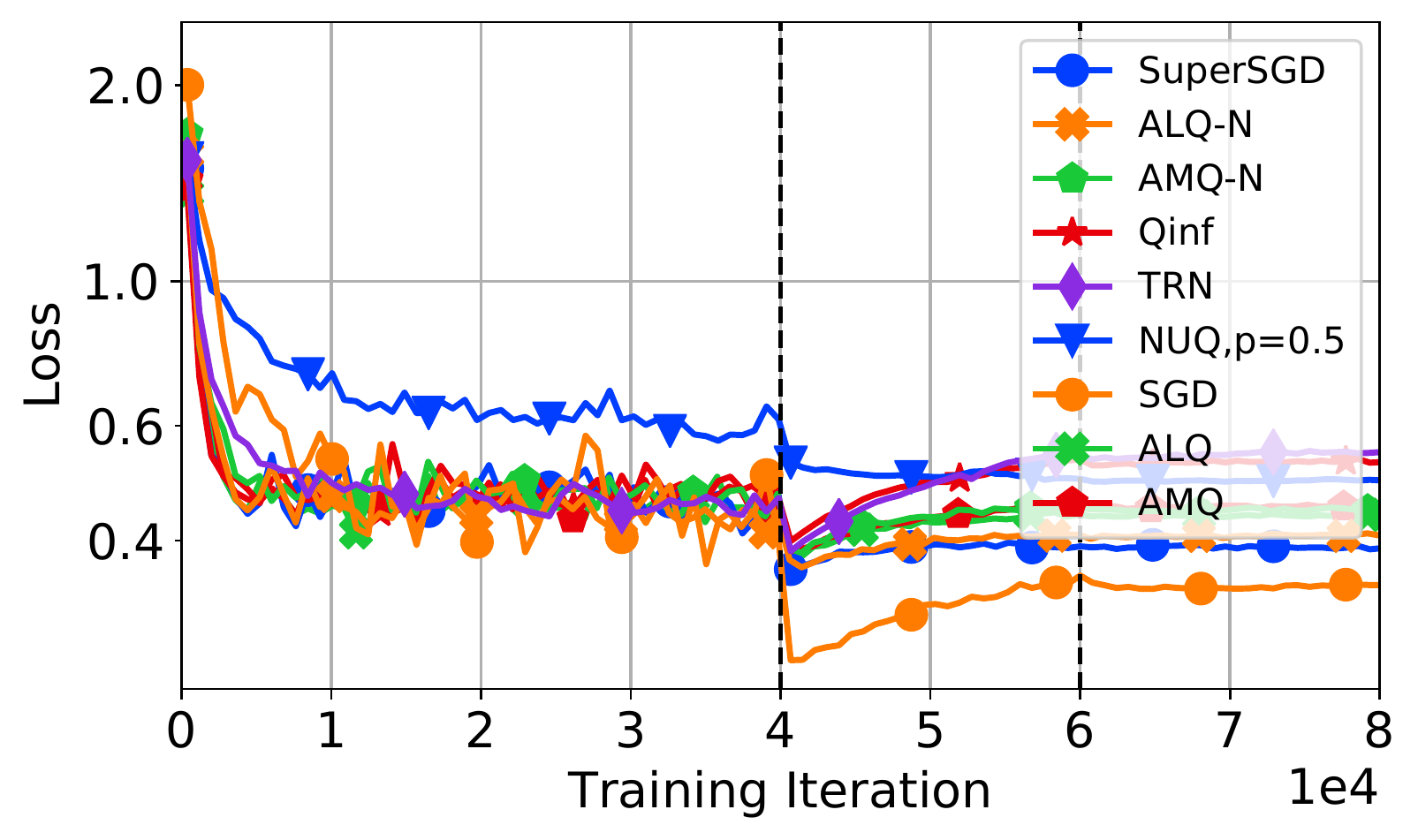}
        \vspace*{\capshift}
        \caption{ResNet-32 on CIFAR-10}
        \label{fig:cifar10_resnet32_\EXPNAME}
    \end{subfigure}
    \begin{subfigure}[b]{\threecolfigwidth}
        \includegraphics[width=\textwidth]{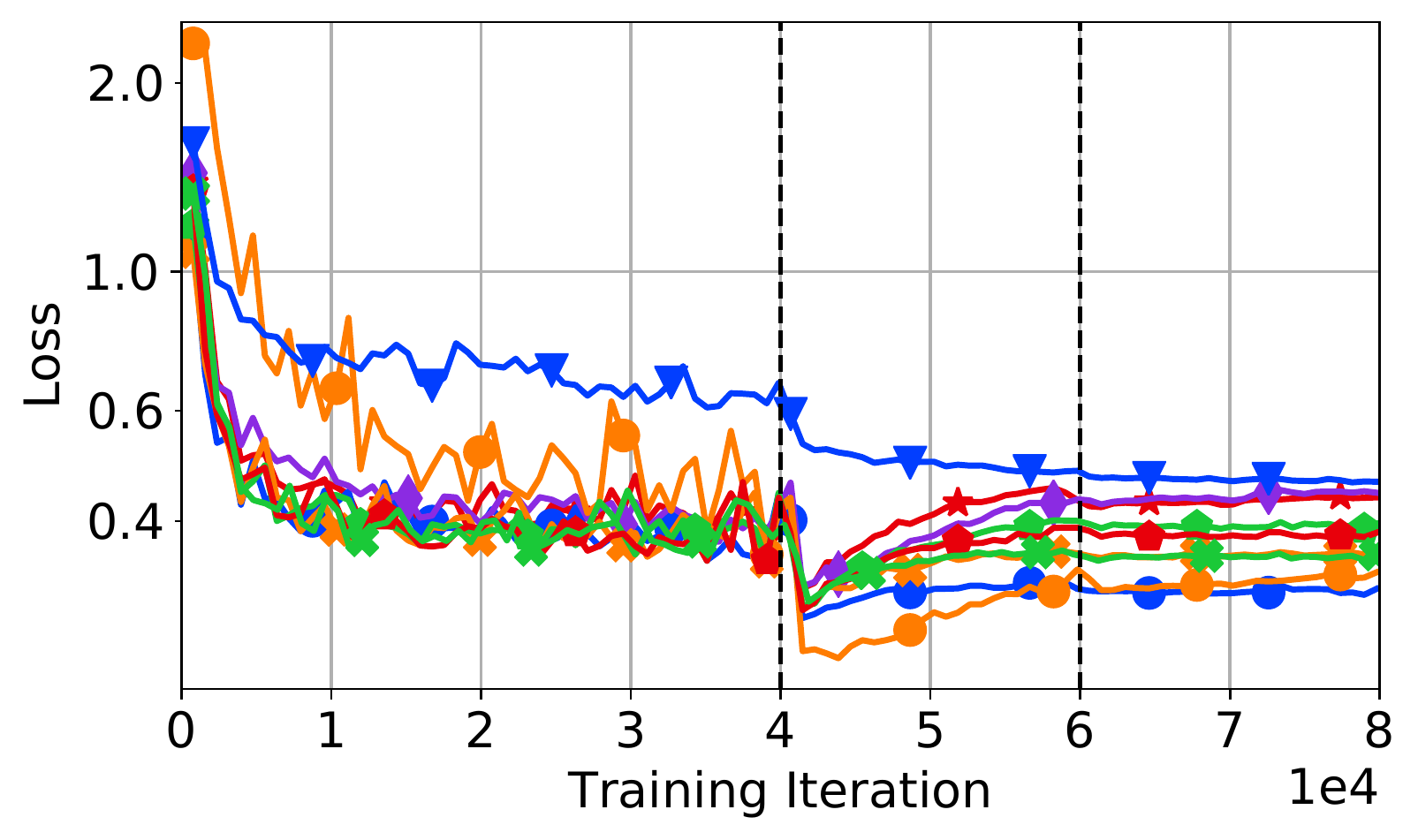}
        \vspace*{\capshift}
        \caption{ResNet-110 on CIFAR-10}
        \label{fig:cifar10_resnet110_\EXPNAME}
    \end{subfigure}
    \begin{subfigure}[b]{\threecolfigwidth}
        \includegraphics[width=\textwidth]{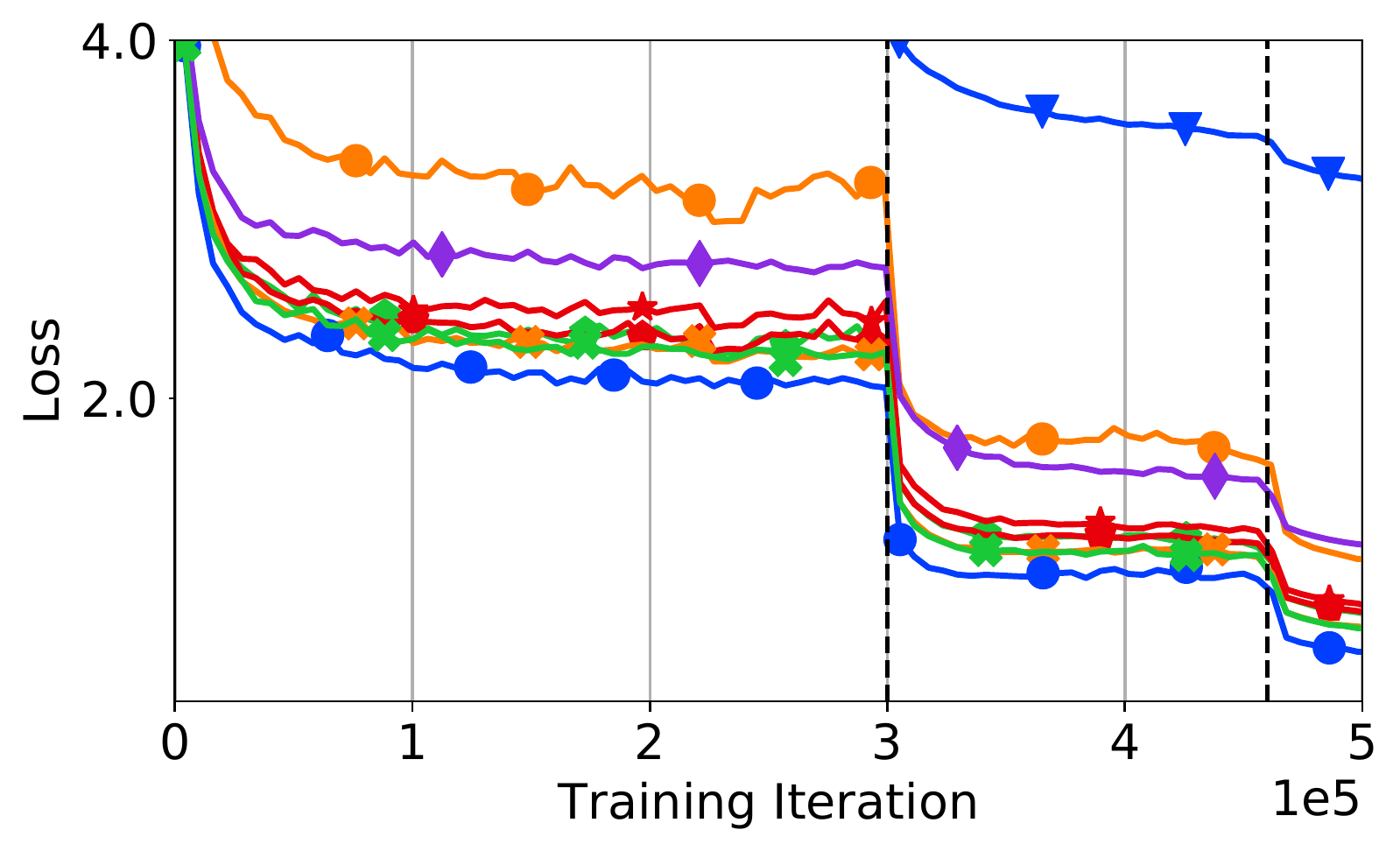}
        \vspace*{\capshift}
        \caption{ResNet-18 on ImageNet}
        \label{fig:imagenet_\EXPNAME}
    \end{subfigure}
    \caption{\textbf{{\FIGTITLE}} on CIFAR-10 and ImageNet. All methods use $3$ 
    quantization bits. Bucket size for ResNet-110 trained on CIFAR-10 is
    $16384$,  for ResNet-32 is $8192$, and for ResNet-18 on ImageNet is 
    $8192$.}
\label{fig:\EXPNAME}
\end{figure}
}
{\def\EXPNAME{-e/est_var}\def\FIGTITLE{Extended Variance (no train)}

\begin{figure}[t]
    \centering
    \begin{subfigure}[b]{\threecolfigwidth}
        \includegraphics[width=\textwidth]{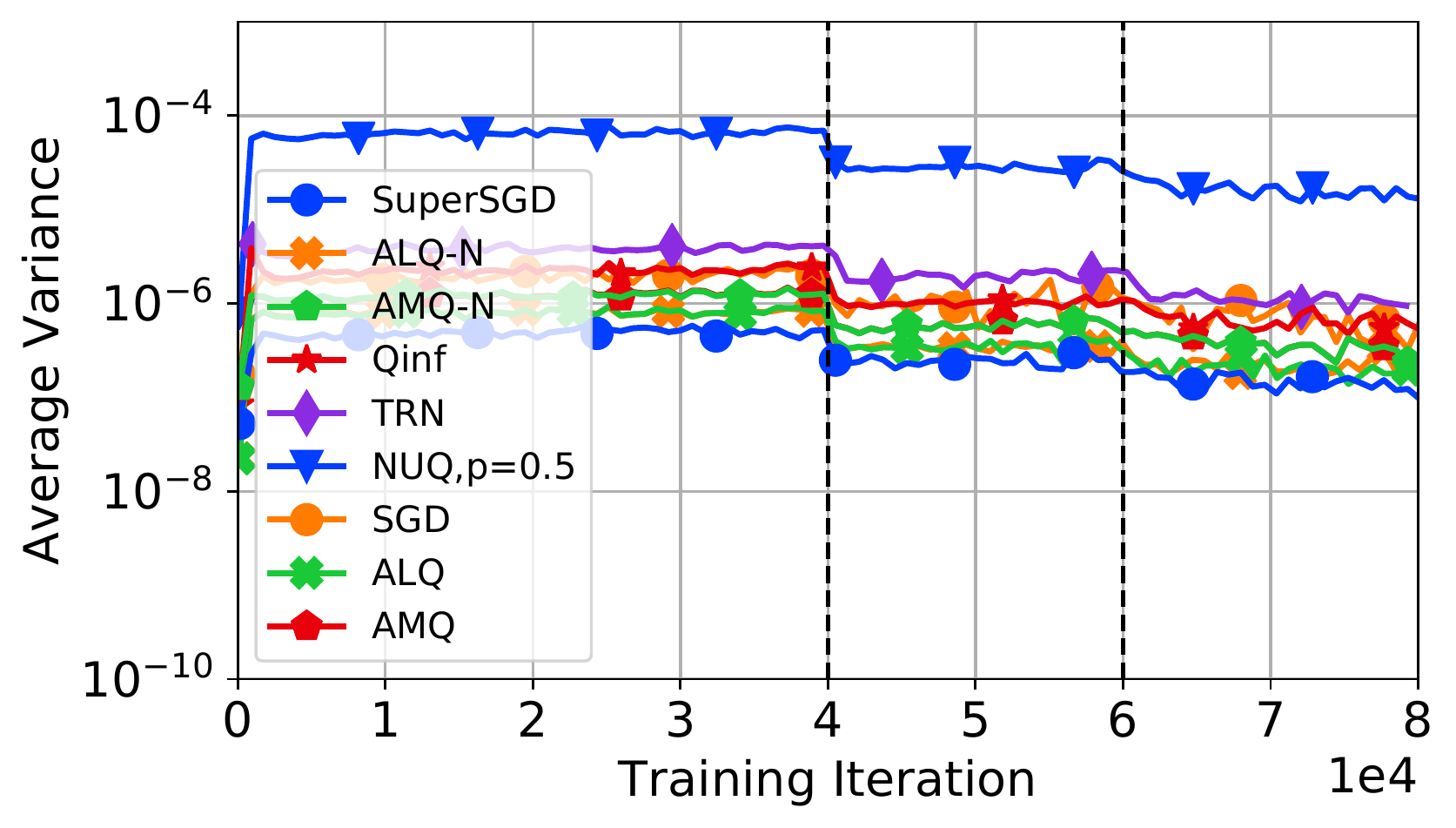}
        \vspace*{\capshift}
        \caption{ResNet-32 on CIFAR-10}
        \label{fig:cifar10_resnet32_\EXPNAME}
    \end{subfigure}
    \begin{subfigure}[b]{\threecolfigwidth}
        \includegraphics[width=\textwidth]{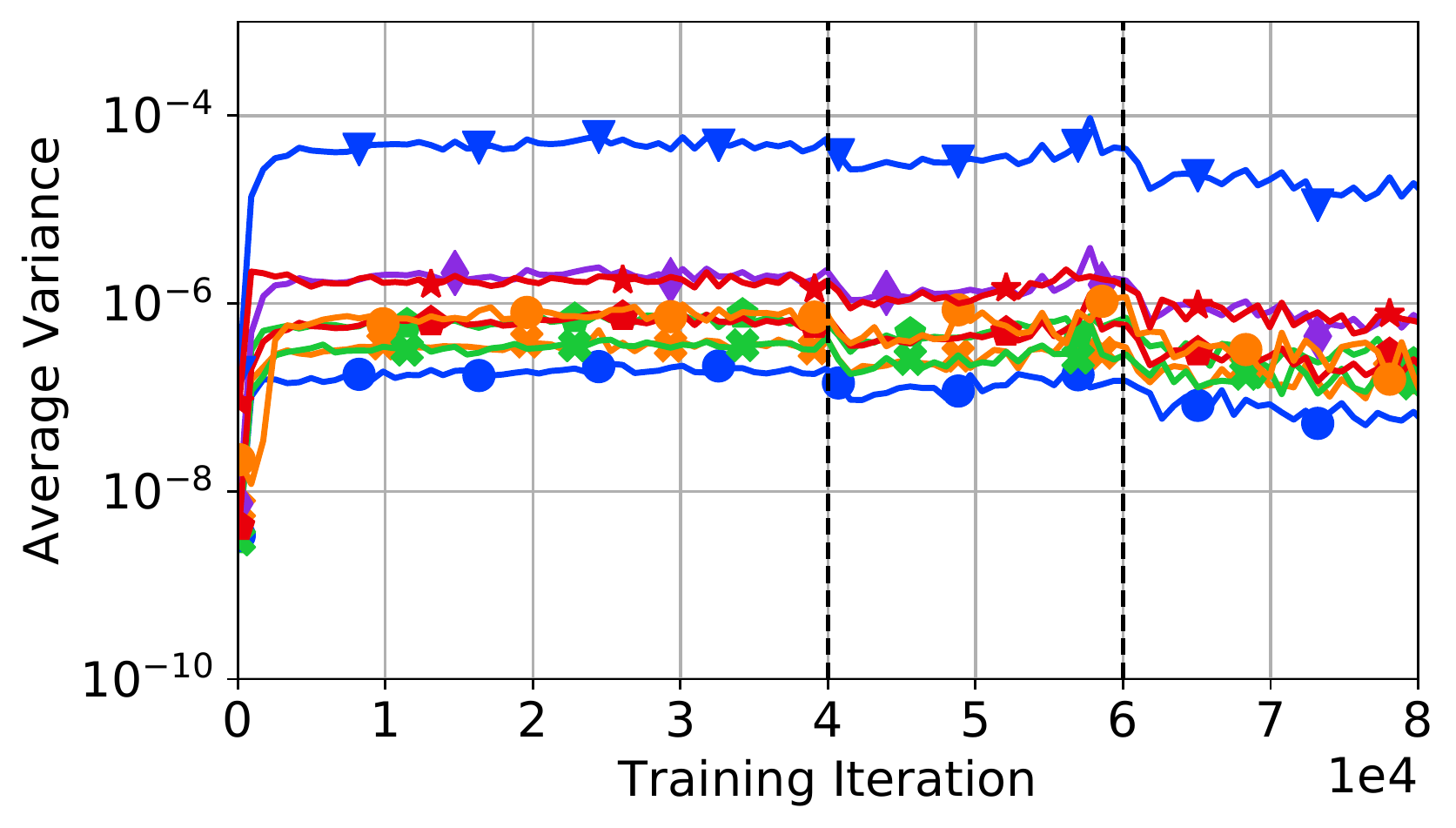}
        \vspace*{\capshift}
        \caption{ResNet-110 on CIFAR-10}
        \label{fig:cifar10_resnet110_\EXPNAME}
    \end{subfigure}
    \begin{subfigure}[b]{\threecolfigwidth}
        \includegraphics[width=\textwidth]{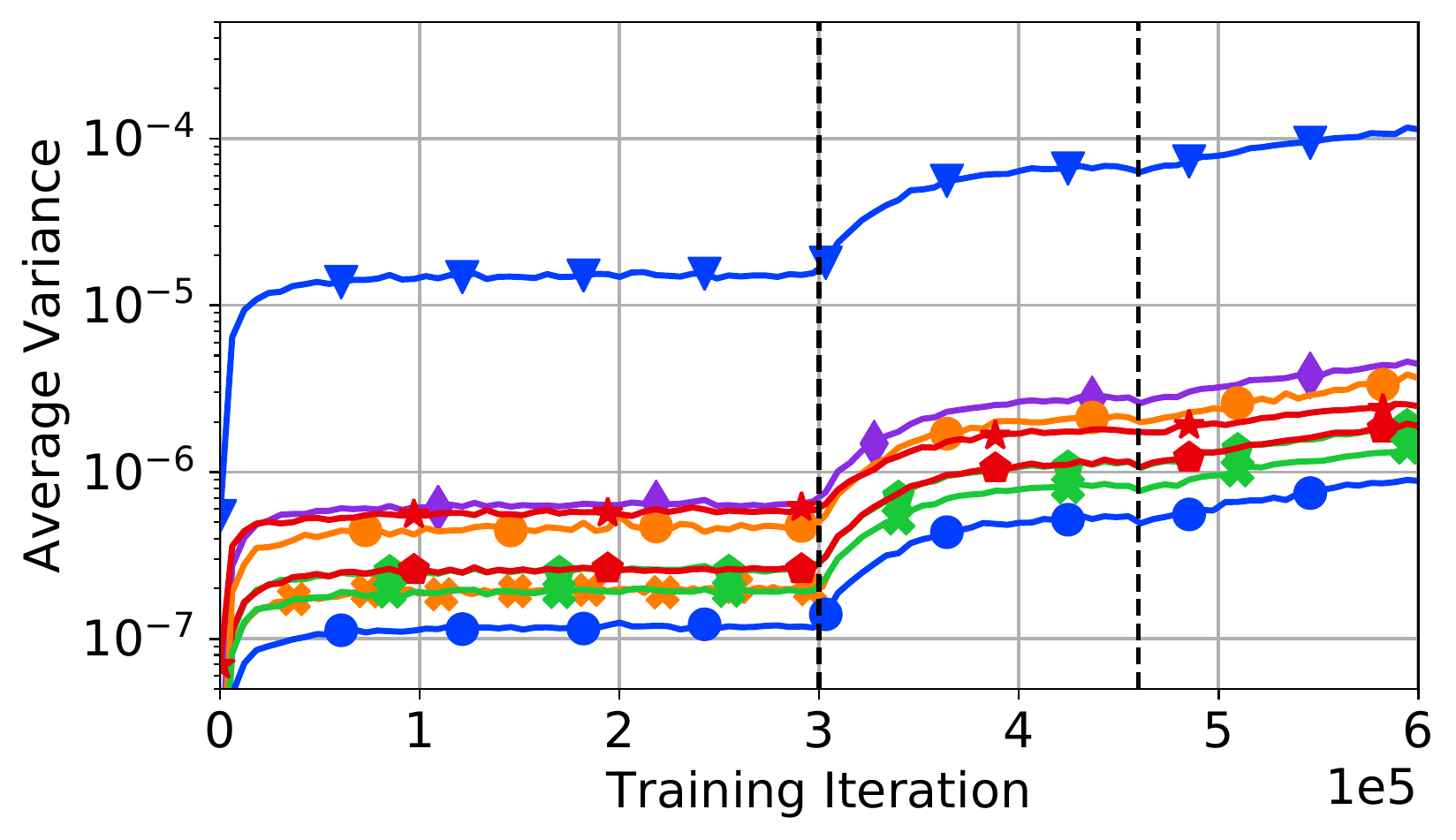}
        \vspace*{\capshift}
        \caption{ResNet-18 on ImageNet}
        \label{fig:imagenet_\EXPNAME}
    \end{subfigure}
    \caption{\textbf{{\FIGTITLE}} on CIFAR-10 and ImageNet. All methods use $3$ 
    quantization bits. Bucket size for ResNet-110 trained on CIFAR-10 is
    $16384$,  for ResNet-32 is $8192$, and for ResNet-18 on ImageNet is 
    $8192$.}
\label{fig:\EXPNAME}
\end{figure}
}
{\def\EXPNAME{-g-e/est_var}\def\FIGTITLE{Extended Variance}

\begin{figure}[t]
    \centering
    \begin{subfigure}[b]{\threecolfigwidth}
        \includegraphics[width=\textwidth]{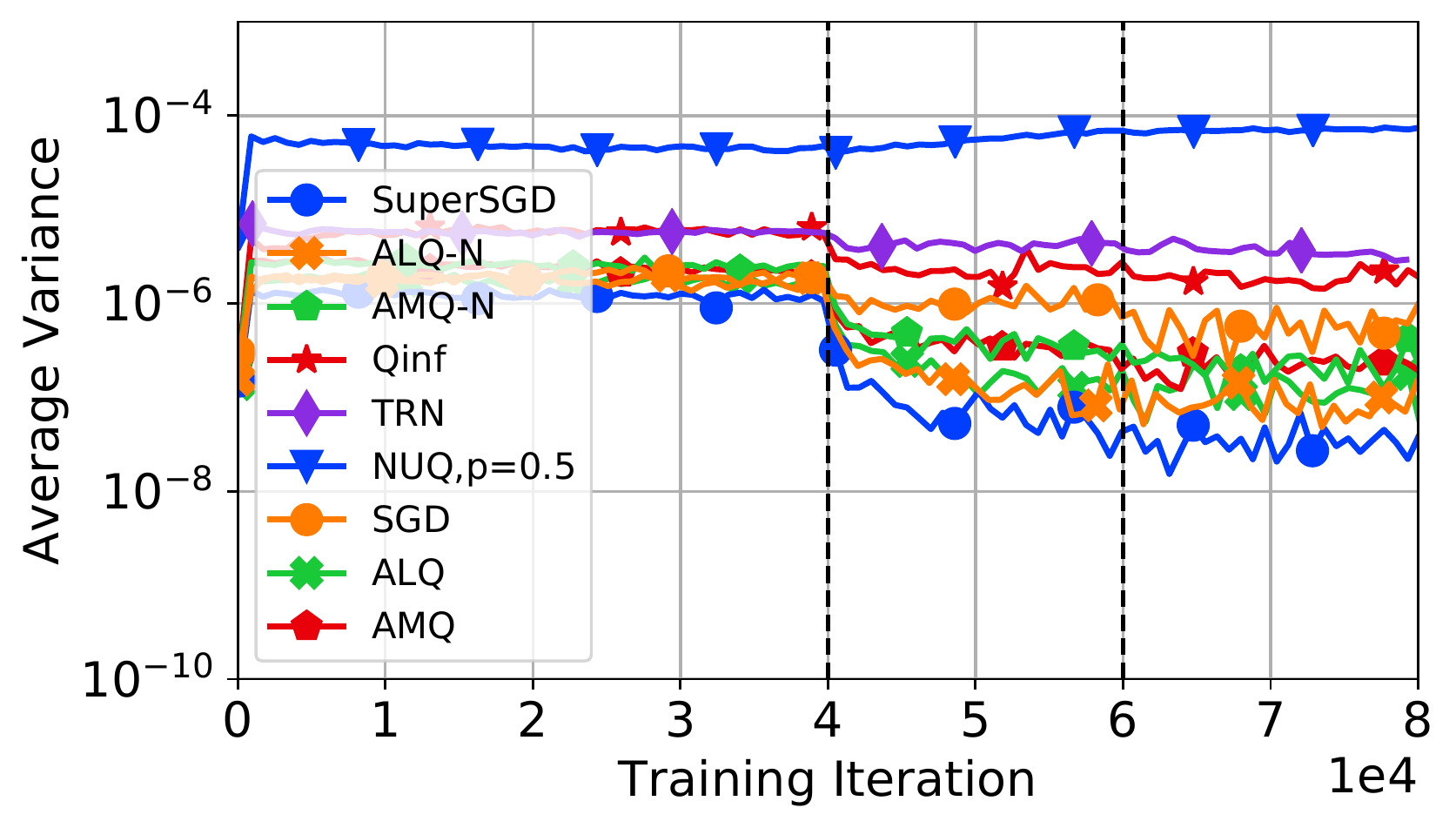}
        \vspace*{\capshift}
        \caption{ResNet-32 on CIFAR-10}
        \label{fig:cifar10_resnet32_\EXPNAME}
    \end{subfigure}
    \begin{subfigure}[b]{\threecolfigwidth}
        \includegraphics[width=\textwidth]{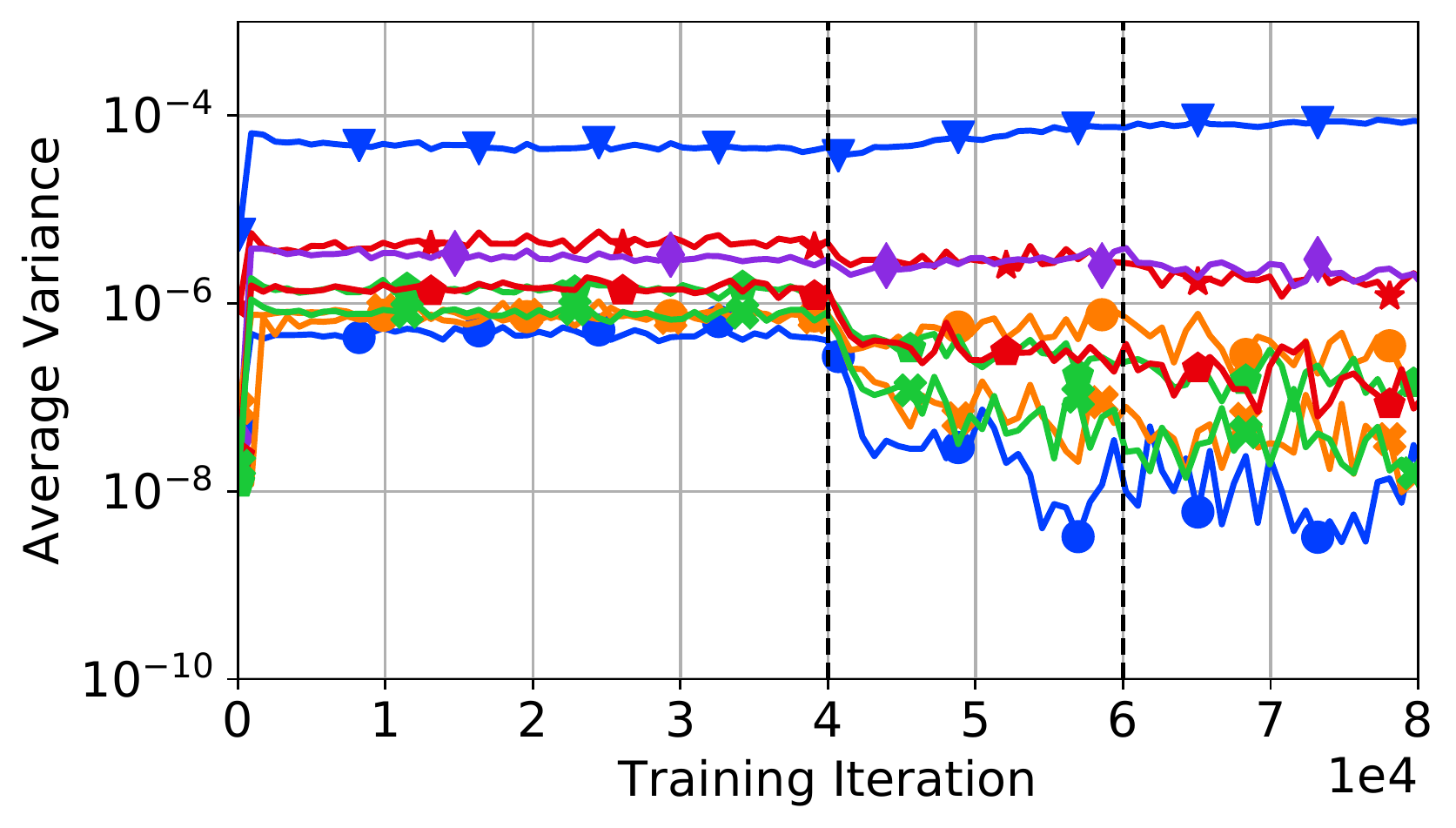}
        \vspace*{\capshift}
        \caption{ResNet-110 on CIFAR-10}
        \label{fig:cifar10_resnet110_\EXPNAME}
    \end{subfigure}
    \begin{subfigure}[b]{\threecolfigwidth}
        \includegraphics[width=\textwidth]{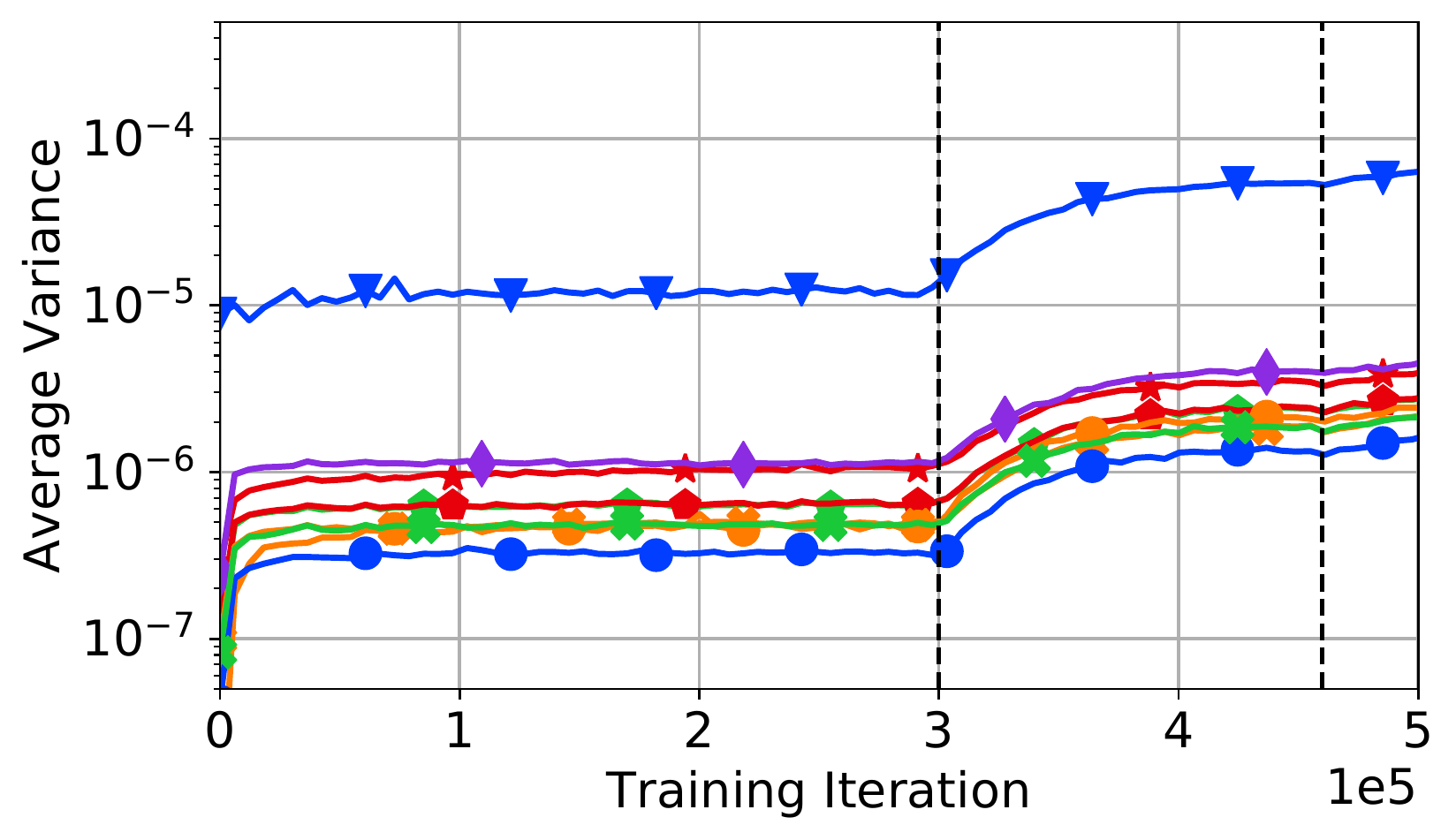}
        \vspace*{\capshift}
        \caption{ResNet-18 on ImageNet}
        \label{fig:imagenet_\EXPNAME}
    \end{subfigure}
    \caption{\textbf{{\FIGTITLE}} on CIFAR-10 and ImageNet. All methods use $3$ 
    quantization bits. Bucket size for ResNet-110 trained on CIFAR-10 is
    $16384$,  for ResNet-32 is $8192$, and for ResNet-18 on ImageNet is 
    $8192$.}
\label{fig:\EXPNAME}
\end{figure}
}

\subsection{Revised Experiments}
\cref{fig:-g-e/Tloss,fig:-g-e/Vloss,fig:-e/est_var,fig:-g-e/est_var} are
extended versions of
\cref{fig:-g/Vloss,fig:-e/est_var,fig:-g/est_var} figures in
the main body. The difference is that they contain more baselines compared to
the figures in the main body. \cref{fig:-g-e/Tloss}
contains the training loss for the experiments on CIFAR-10. It was not possible
to include the same figure for ImageNet, because calculating the full training
loss on ImageNet takes a very long time.

\begin{figure}[t]
    \centering
    \begin{subfigure}[b]{\threecolfigwidth}
        \includegraphics[width=\textwidth]{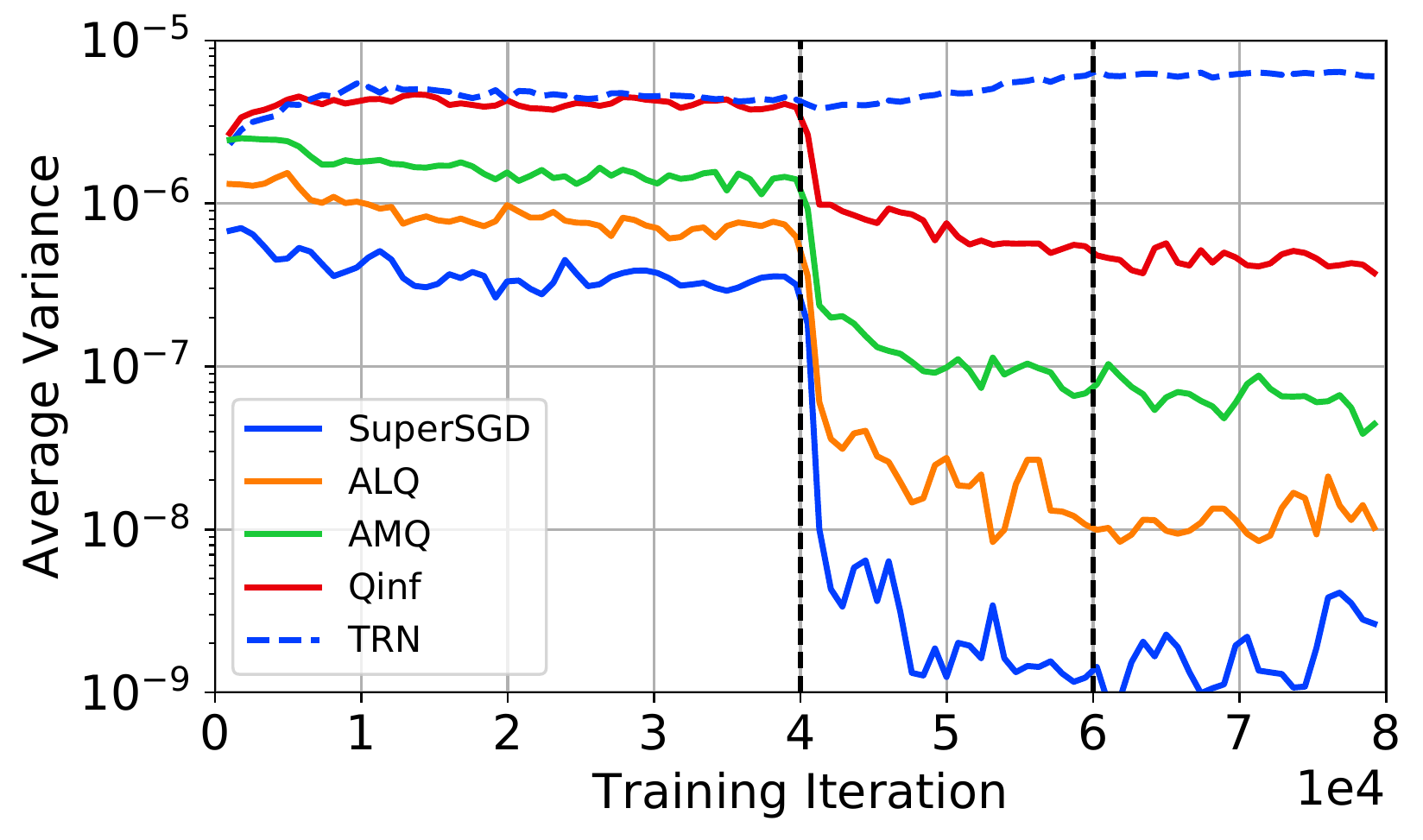}
        \vspace*{\capshift}
        \caption{Variance}
        \label{fig:gpu32_var}
    \end{subfigure}
    \begin{subfigure}[b]{\threecolfigwidth}
        \includegraphics[width=\textwidth]{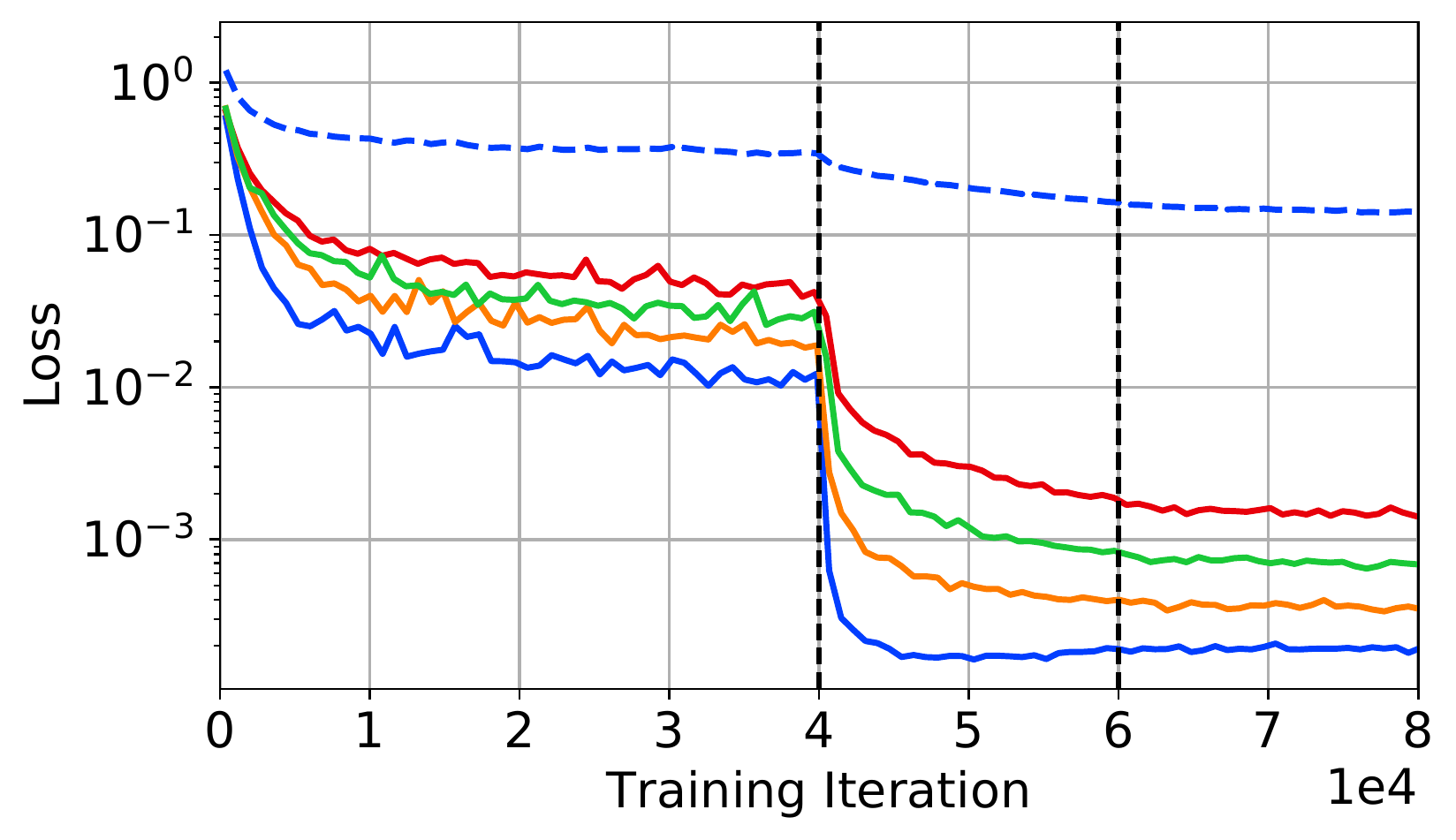}
        \vspace*{\capshift}
        \caption{Training Loss}
        \label{fig:gpu32_tloss}
    \end{subfigure}
    \begin{subfigure}[b]{\threecolfigwidth}
        \includegraphics[width=\textwidth]{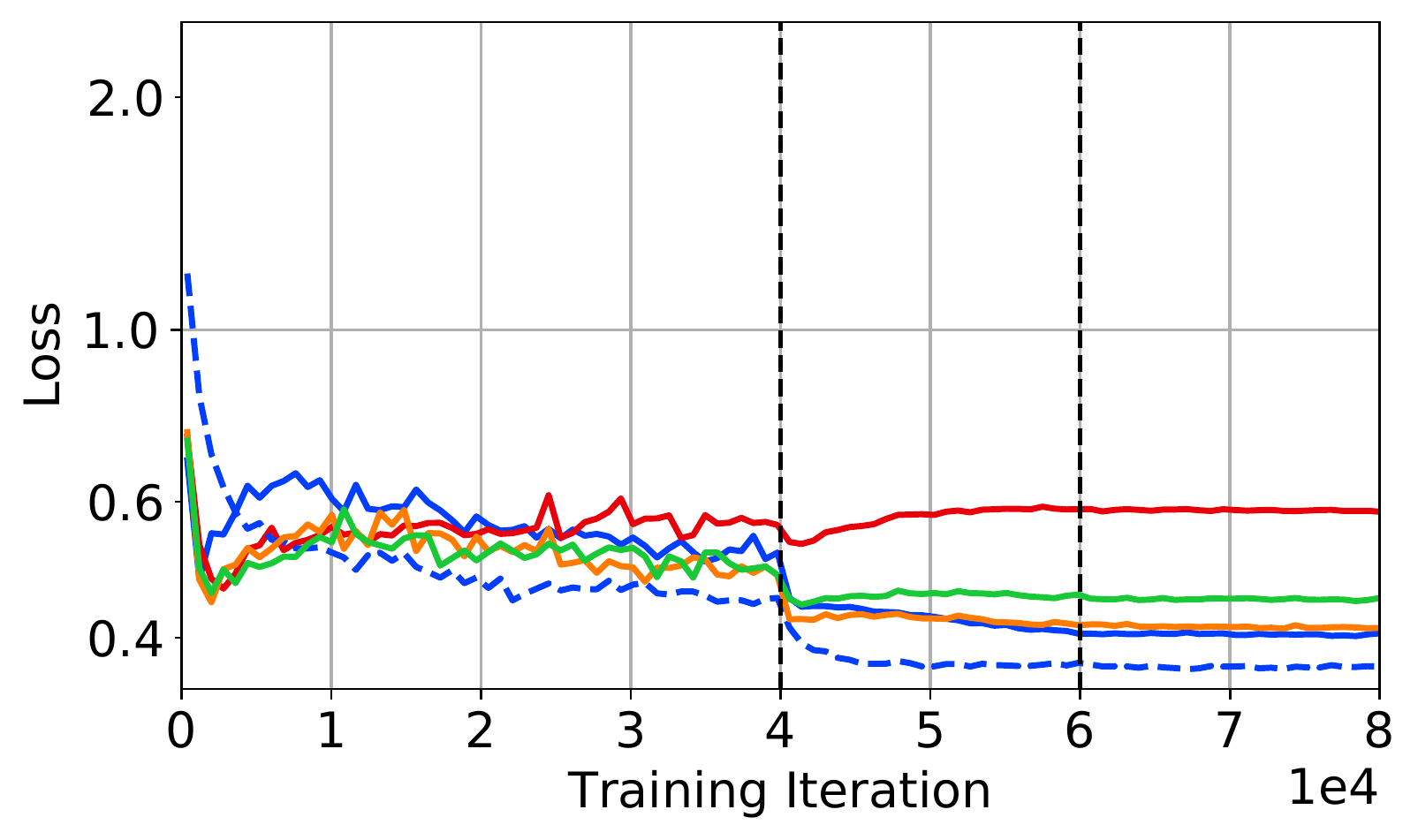}
        \vspace*{\capshift}
        \caption{Validation Loss}
        \label{fig:gpu32_val}
    \end{subfigure}
    \caption{Using 32-GPUs to train ResNet-32 on CIFAR-10.}
\label{fig:gpus}
\end{figure}

The expected variance, training loss, and the validation loss for the
results presented in \cref{tab:gpus} are shown in \cref{fig:gpus}. Although
\alqnbased performs better in expected variance and training loss, it seems to
have trouble when it comes to the validation loss for 32-GPUs. We suspect that
this is due to the large total batch size used for these experiment that results
in overfitting. The batch size for each GPU is 128.

\subsection{Effect of Using Gradient Clipping}
\terngrad~\citep{TernGrad} introduced the idea of gradient clipping before quantization.
Gradient clipping replaces the gradient coordinates that are far from the mean
to reduce the gradient variance. The gradient coordinates that are very far from
the mean can affect the normalization. In order to tackle this problem, they
clip the gradients before quantization. The clipping process can be described
using \cref{eq:clipping}:

\begin{align}
    f(g_i) = 
    \begin{cases}
        g_i               & |g_i| \leq c\sigma \\
        \sign(g_i). c\sigma & |g_i| > c\sigma 
    \end{cases}
    \label{eq:clipping}
\end{align}

\begin{wrapfigure}{R}{0.4\textwidth}
    \vspace*{-0.7cm}
    \includegraphics[width=0.39\textwidth]{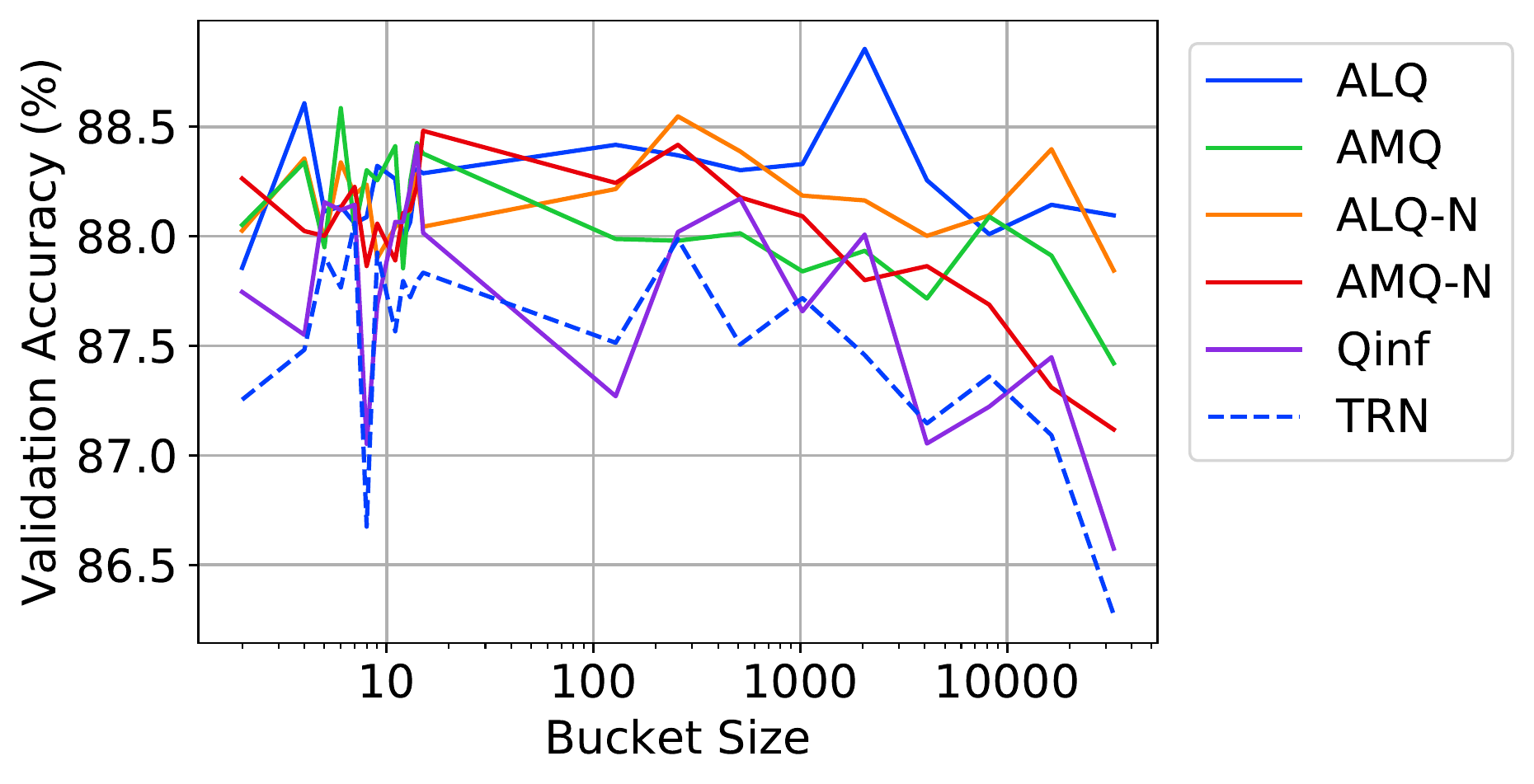}
    \caption{Validation Loss}
    \label{fig:clipping}
    \vspace*{-0.4cm}
\end{wrapfigure}
The constant $c$ used in \terngrad equals to 2.5. In order to investigate the
effect of gradient clipping in \alqnbased and \amqnbased, we train a ResNet-8 on
CIFAR-10 dataset for various bucket sizes. \cref{fig:clipping} shows the
validation accuracy of the baselines and the algorithms we proposed. \alqnbased
and \alqnless always maintain better or equal accuracy compared to the other
quantization schemes. It is also worth noting that the quantization is performed
by each layer instead of a performing the quantization across the network
without considering the layers.

\subsection{Timing Overhead}
\label{sec:timing}

\begin{table}
\centering
\caption{Training ResNet50 on ImageNet with min-batch size $512$. Time per step 
    for training with 32bits full-precision is $1.2$s and with $16$ bits full-precision is $0.61$s.}
\label{tab:timing_resnet50}
\begin{tabular}{ccccc}
\toprule
Bits&           Bucket size&            Time per step (s)&          Ratio to 
    FP32&          Ratio to FP16\\
    \midrule
$2$&            $64$&           $0.41$&         $0.34$&         $0.67$ \\
$2$&            $256$&          $0.39$&         $0.33$&         $0.64$ \\
$2$&            $1024$&         $0.38$&         $0.32$&         $0.62$ \\
$2$&            $8192$&         $0.38$&         $0.32$&         $0.62$ \\
$2$&            $16384$&                $0.38$&         $0.32$&         $0.62$ \\
$3$&            $64$&           $0.4$&          $0.33$&         $0.66$ \\
$3$&            $256$&          $0.4$&          $0.33$&         $0.66$ \\
$3$&            $1024$&         $0.4$&          $0.33$&         $0.66$ \\
$3$&            $8192$&         $0.4$&          $0.33$&         $0.66$ \\
$3$&            $16384$&                $0.38$&         $0.32$&         $0.62$ \\
$4$&            $64$&           $0.42$&         $0.35$&         $0.69$ \\
$4$&            $256$&          $0.41$&         $0.34$&         $0.67$ \\
$4$&            $1024$&         $0.4$&          $0.33$&         $0.66$ \\
$4$&            $8192$&         $0.4$&          $0.33$&         $0.66$ \\
$4$&            $16384$&                $0.4$&          $0.33$&         $0.66$ \\
$5$&            $64$&           $0.43$&         $0.36$&         $0.70$ \\
$5$&            $256$&          $0.42$&         $0.35$&         $0.69$ \\
$5$&            $1024$&         $0.41$&         $0.34$&         $0.67$ \\
$5$&            $8192$&         $0.4$&          $0.33$&         $0.66$ \\
$5$&            $16384$&                $0.4$&          $0.33$&         $0.66$ \\
$6$&            $64$&           $0.42$&         $0.35$&         $0.69$ \\
$6$&            $256$&          $0.41$&         $0.34$&         $0.67$ \\
$6$&            $1024$&         $0.41$&         $0.34$&         $0.67$ \\
$6$&            $8192$&         $0.41$&         $0.34$&         $0.67$ \\
$6$&            $16384$&                $0.41$&         $0.34$&         $0.67$ \\
$7$&            $64$&           $0.45$&         $0.38$&         $0.74$ \\
$7$&            $256$&          $0.43$&         $0.36$&         $0.70$ \\
$7$&            $1024$&         $0.42$&         $0.35$&         $0.69$ \\
$7$&            $8192$&         $0.42$&         $0.35$&         $0.69$ \\
$7$&            $16384$&                $0.43$&         $0.36$&         $0.70$ \\
$8$&            $64$&           $0.45$&         $0.38$&         $0.74$ \\
$8$&            $256$&          $0.44$&         $0.37$&         $0.72$ \\
$8$&            $1024$&         $0.43$&         $0.36$&         $0.70$ \\
$8$&            $8192$&         $0.43$&         $0.36$&         $0.70$ \\
$8$&            $16384$&                $0.43$&         $0.36$&         $0.70$ \\
\bottomrule
\end{tabular}
\end{table}

\begin{table}
\centering
\caption{Training ResNet18 on ImageNet with min-batch size $512$. Time per step 
    for training with 32bits full-precision is $0.57$s and with $16$ bits full-precision is $0.28$s.}
\label{tab:timing_resnet18}
\begin{tabular}{ccccc}
\toprule
Bits&           Bucket size&            Time per step&          Ratio to FP32&          Ratio to FP16\\
    \midrule
$2$&            $64$&           $0.13$&         $0.23$&         $0.46                                         $\\
$2$&            $256$&          $0.12$&         $0.21$&         $0.43                                         $\\
$2$&            $1024$&         $0.11$&         $0.19$&         $0.39                                         $\\
$2$&            $8192$&         $0.11$&         $0.19$&         $0.39                                         $\\
$2$&            $16384$&                $0.11$&         $0.19$&         $0.39                                     $\\
$3$&            $64$&           $0.13$&         $0.23$&         $0.46                                         $\\
$3$&            $256$&          $0.12$&         $0.21$&         $0.43                                         $\\
$3$&            $1024$&         $0.12$&         $0.21$&         $0.43                                         $\\
$3$&            $8192$&         $0.12$&         $0.21$&         $0.43                                         $\\
$3$&            $16384$&                $0.12$&         $0.21$&         $0.43                                     $\\
$4$&            $64$&           $0.13$&         $0.23$&         $0.46                                         $\\
$4$&            $256$&          $0.13$&         $0.23$&         $0.46                                         $\\
$4$&            $1024$&         $0.12$&         $0.21$&         $0.43                                         $\\
$4$&            $8192$&         $0.12$&         $0.21$&         $0.43                                         $\\
$4$&            $16384$&                $0.12$&         $0.21$&         $0.43                                     $\\
$5$&            $64$&           $0.13$&         $0.23$&         $0.46                                         $\\
$5$&            $256$&          $0.13$&         $0.23$&         $0.46                                         $\\
$5$&            $1024$&         $0.13$&         $0.23$&         $0.46                                         $\\
$5$&            $8192$&         $0.13$&         $0.23$&         $0.46                                         $\\
$5$&            $16384$&                $0.13$&         $0.23$&         $0.46                                     $\\
$6$&            $64$&           $0.14$&         $0.25$&         $0.50                                         $\\
$6$&            $256$&          $0.13$&         $0.23$&         $0.46                                         $\\
$6$&            $1024$&         $0.13$&         $0.23$&         $0.46                                         $\\
$6$&            $8192$&         $0.13$&         $0.23$&         $0.46                                         $\\
$6$&            $16384$&                $0.13$&         $0.23$&         $0.46                                     $\\
$7$&            $64$&           $0.14$&         $0.25$&         $0.50                                         $\\
$7$&            $256$&          $0.13$&         $0.23$&         $0.46                                         $\\
$7$&            $1024$&         $0.14$&         $0.25$&         $0.50                                         $\\
$7$&            $8192$&         $0.13$&         $0.23$&         $0.46                                         $\\
$7$&            $16384$&                $0.13$&         $0.23$&         $0.46                                     $\\
$8$&            $64$&           $0.15$&         $0.26$&         $0.54                                         $\\
$8$&            $256$&          $0.14$&         $0.25$&         $0.50                                         $\\
$8$&            $1024$&         $0.14$&         $0.25$&         $0.50                                         $\\
$8$&            $8192$&         $0.14$&         $0.25$&         $0.50                                         $\\
$8$&            $16384$&                $0.14$&         $0.25$&         $0.50                                     $\\
\bottomrule
\end{tabular}
\end{table}

\begin{table}
\centering
\caption{Additional overhead of proposed methods for training ResNet18 on 
    ImageNet (\cref{tab:timing_resnet18}). We also show the cost of performing 
    $3$ updates relative to the total cost of training for $60$ epochs. Full-precision
    training for $60$ epochs with 32 bits takes $95$ hours while with 
    16 bits takes $46$ hours.  }
\label{tab:timing_resnet50}
\begin{tabular}{cccccc}
\toprule
Bits&           Bucket size&            Quantization Method&            Time 
    per update&                Ratio to FP32&           Ratio to FP16            
    \\
    \midrule
$3$&            $64$&           \alqnless&            $1012$&         $0.89$&         $1.81$                                              \\
$3$&            $256$&          \alqnless&            $630$&          $0.55$&         $1.13$                                              \\
$3$&            $1024$&         \alqnless&            $533$&          $0.47$&         $0.95$                                              \\
$3$&            $8192$&         \alqnless&            $559$&          $0.49$&         $1.00$                                              \\
$3$&            $16384$&                \alqnless&            $591$&          $0.52$&         $1.06$                                          \\
$4$&            $64$&           \alqnless&            $1170$&         $1.03$&         $2.09$                                              \\
$4$&            $256$&          \alqnless&            $822$&          $0.72$&         $1.47$                                              \\
$4$&            $1024$&         \alqnless&            $733$&          $0.64$&         $1.31$                                              \\
$4$&            $8192$&         \alqnless&            $681$&          $0.60$&         $1.22$                                              \\
$4$&            $16384$&                \alqnless&            $684$&          $0.60$&         $1.22$                                          \\
$6$&            $64$&           \alqnless&            $2036$&         $1.79$&         $3.64$                                              \\
$6$&            $256$&          \alqnless&            $1710$&         $1.50$&         $3.05$                                              \\
$6$&            $1024$&         \alqnless&            $1556$&         $1.36$&         $2.78$                                              \\
$6$&            $8192$&         \alqnless&            $1574$&         $1.38$&         $2.81$                                              \\
$6$&            $16384$&                \alqnless&            $1671$&         $1.47$&         $2.98$                                          \\
$8$&            $64$&           \alqnless&            $5604$&         $4.92$&         $10.01$                                         \\
$8$&            $256$&          \alqnless&            $5253$&         $4.61$&         $9.38$                                              \\
$8$&            $1024$&         \alqnless&            $5478$&         $4.81$&         $9.78$                                              \\
$8$&            $8192$&         \alqnless&            $5180$&         $4.54$&         $9.25$                                              \\
$8$&            $16384$&                \alqnless&            $5576$&         $4.89$&         $9.96$                                          \\
$3$&            $64$&           \alqnbased&           $1032$&         $0.91$&         $1.84$                                          \\
$3$&            $256$&          \alqnbased&           $585$&          $0.51$&         $1.04$                                          \\
$3$&            $1024$&         \alqnbased&           $444$&          $0.39$&         $0.79$                                          \\
$3$&            $8192$&         \alqnbased&           $474$&          $0.42$&         $0.85$                                          \\
$3$&            $16384$&                \alqnbased&           $477$&          $0.42$&         $0.85$                                      \\
$4$&            $64$&           \alqnbased&           $930$&          $0.82$&         $1.66$                                          \\
$4$&            $256$&          \alqnbased&           $529$&          $0.46$&         $0.94$                                          \\
$4$&            $1024$&         \alqnbased&           $450$&          $0.39$&         $0.80$                                          \\
$4$&            $8192$&         \alqnbased&           $431$&          $0.38$&         $0.77$                                          \\
$4$&            $16384$&                \alqnbased&           $486$&          $0.43$&         $0.87$                                      \\
$6$&            $64$&           \alqnbased&           $974$&          $0.85$&         $1.74$                                          \\
$6$&            $256$&          \alqnbased&           $573$&          $0.50$&         $1.02$                                          \\
$6$&            $1024$&         \alqnbased&           $489$&          $0.43$&         $0.87$                                          \\
$6$&            $8192$&         \alqnbased&           $428$&          $0.38$&         $0.76$                                          \\
$6$&            $16384$&                \alqnbased&           $438$&          $0.38$&         $0.78$                                      \\
$8$&            $64$&           \alqnbased&           $1051$&         $0.92$&         $1.88$                                          \\
$8$&            $256$&          \alqnbased&           $637$&          $0.56$&         $1.14$                                          \\
$8$&            $1024$&         \alqnbased&           $516$&          $0.45$&         $0.92$                                          \\
$8$&            $8192$&         \alqnbased&           $508$&          $0.45$&         $0.91$                                          \\
$8$&            $16384$&                \alqnbased&           $516$&          $0.45$&         $0.92$                                      \\
\bottomrule
\end{tabular}
\end{table}

In this section, we provide the timing results per step for training ResNet-18 
(\cref{tab:timing_resnet18}) and ResNet-50 (\cref{tab:timing_resnet50}) on 
ImageNet with mini-batch size $512$. The training setup consists of $4$ AWS 
nodes with one V100 GPU on each.  Network bandwidth programmatically 
constrained to 1GBit/s.

\end{document}